%% file: iclr2026_conference.tex
\documentclass{article}
\usepackage{iclr2026_conference,times}

\usepackage{amsmath}
\usepackage{amssymb}
\usepackage{amsfonts}
\usepackage{mathtools}
\usepackage[ruled,vlined,linesnumbered]{algorithm2e}
\usepackage{amsmath}
\usepackage{graphicx}
\usepackage{subfig}
\usepackage{booktabs}
\usepackage{amsmath}
\usepackage{multirow}
\usepackage{mdframed}
\usepackage{quoting}
\usepackage{booktabs}
\usepackage{makecell}
\usepackage{tabularx}
\usepackage{longtable}
\usepackage{supertabular}
\usepackage{xltabular}
\usepackage{array}
\usepackage{colortbl}
\usepackage{tablefootnote}
\usepackage[T1]{fontenc}

\usepackage{graphicx}
\usepackage{float}
\usepackage{placeins}
\usepackage{adjustbox}
\usepackage{caption}

\usepackage{times}
\usepackage{inconsolata}
\usepackage{fontawesome5}
\usepackage{tikzsymbols}

\usepackage{microtype}
\usepackage{soul}
\usepackage{verbatim}
\usepackage{enumitem}
\usepackage{fancyvrb}
\usepackage{xfrac}
\usepackage{tabto}

\usepackage{listings}
\usepackage{caption}
\usepackage{xcolor}

\usepackage[most]{tcolorbox}

\definecolor{valbest}{HTML}{d9ead3}
\newcommand{\valbest}[1]{\colorbox{valbest}{#1}}
\definecolor{valgood}{HTML}{cfe6ec}
\newcommand{\valgood}[1]{\colorbox{valgood}{#1}}

\usepackage{etoolbox}
\patchcmd{\cite}{\@ifundefined}{\@ifpackageloaded{natbib}{}{\@ifundefined}}{}{}

\usepackage[pagebackref,breaklinks,colorlinks]{hyperref}
\usepackage{cleveref}
\usepackage{url}
\tcbuselibrary{listings,breakable}
\input{math_commands.tex}

\input{macros}

\newcommand{\somesh}[1]{}
\newcommand{\yaman}[1]{}
\newcommand{\yam}[1]{}

\newcommand{\someshCheck}[1]{}
\newcommand{\harini}[1]{}
\title{\centering Accelerating Social Science Research via Agentic Hypothesization and Experimentation}

\renewcommand{\headrulewidth}{0.4pt} 
\renewcommand{\footrulewidth}{0pt}
\author{%
\parbox{\textwidth}{\centering\vspace{4mm}

\setlength{\tabcolsep}{6pt}
\begin{tabular}{cccc}
  \auth{Jishu Sengupta\coauth}{\adobelogo} &
  \auth{Harini S I\coauth}{\adobelogo} &
  \auth{Somesh Singh\coauth}{\adobelogo~\ublogo~\iiitdlogo} &
  \auth{Syed Mohamad Tawseeq\coauth}{\adobelogo} \\
  \auth{Yaman K Singla}{\adobelogo} &
  \auth{David Doermann}{\ublogo} &
  \auth{Rajiv Ratn Shah}{\iiitdlogo} &
  \auth{Balaji Krishnamurthy}{\adobelogo} \\[3.5mm]  
\end{tabular}
{\adobelogo~Adobe Media and Data Science Research (MDSR)\\\quad
 \iiitdlogo~IIIT-Delhi,\quad
 \ublogo~SUNY at Buffalo\\}
 \faEnvelope\ \texttt{\href{mailto:behavior-in-the-wild@googlegroups.com}{behavior-in-the-wild@googlegroups.com}}
}%
}

\iclrfinalcopy
\begin{document}

\maketitle
\begin{abstract}
Data-driven social science research is inherently slow, relying on iterative cycles of observation, hypothesis formulation, and experimental validation. While recent data-driven methods promise to accelerate parts of this process, they largely fail to support end-to-end scientific discovery.
To bridge this gap, we introduce \textsc{ExperiGen}, an agentic framework that operationalizes end-to-end discovery through a Bayesian-optimization–inspired two-phase search. A \textbf{Gen}erator agent proposes candidate hypotheses, while an \textbf{Experi}menter agent empirically evaluates them. Across multiple domains, \textsc{ExperiGen} consistently discovers 2–4$\times$ more statistically significant hypotheses that are 7–17\% more predictive than prior approaches, and naturally extends to complex settings including multimodal and relational data. Beyond statistical performance, for hypotheses to drive real scientific progress, they must be novel, empirically grounded, and actionable. To assess these qualities, we conduct the first expert evaluation of machine-generated hypotheses, collecting feedback from senior faculty. Among 25 reviewed hypotheses, 88\% were rated moderately or strongly novel, 70\% were deemed impactful and worth pursuing, and most exhibited rigor comparable to senior graduate research. Finally, recognizing that ultimate validation requires real-world evidence, we perform the first A/B test of LLM-generated hypotheses, observing statistically significant effects $(p<10^{-6})$ with a large effect size of $+344\%$.

\end{abstract}

\begin{NoHyper}
    \blfootnote{\coauth \small Equal Contribution. }
\end{NoHyper}

\input{pages/intro}
\input{pages/method}

\input{pages/experiment}

\input{pages/results_discussion}

\vspace{-3mm}
\input{pages/background}

\bibliography{egbib}
\bibliographystyle{iclr2026_conference}

\newpage

\input{pages/appendix}

\clearpage
\newpage

\end{document}

%% file: math_commands.tex
\usepackage{amsmath,amsfonts,bm}

\def\eqref#1{equation~\ref{#1}}

\def\1{\bm{1}}

\DeclareMathAlphabet{\mathsfit}{\encodingdefault}{\sfdefault}{m}{sl}
\SetMathAlphabet{\mathsfit}{bold}{\encodingdefault}{\sfdefault}{bx}{n}

%% file: macros.tex
\usepackage{graphicx}

\newcommand{\coauth}{$^{*}$}
\DeclareRobustCommand{\coauth}{$^*$}

\DeclareRobustCommand{\iiitdlogo}{
  \raisebox{0.25\height}{\includegraphics[height=1.2ex]{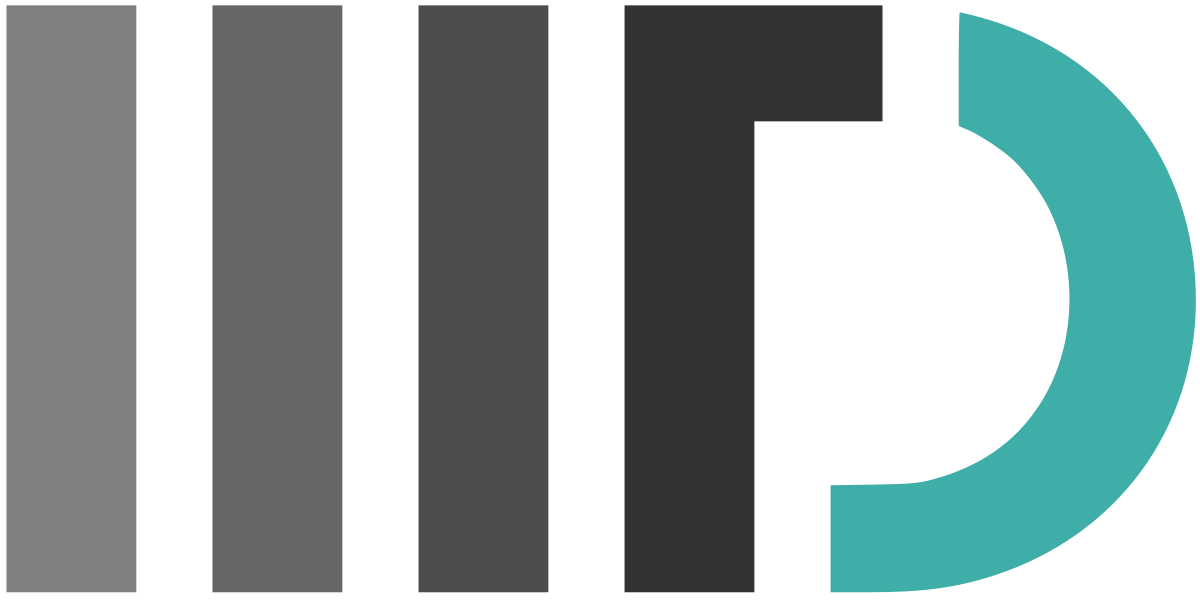}}}

\DeclareRobustCommand{\ublogo}{
  \raisebox{0.25\height}{\includegraphics[height=1.2ex]{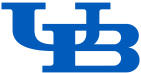}}}

\DeclareRobustCommand{\adobelogo}{
  \raisebox{0.15\height}{\includegraphics[height=2ex]{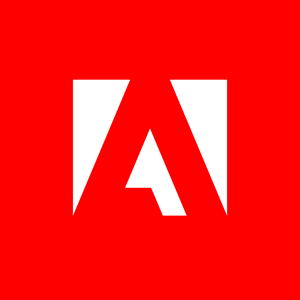}}}

\newcommand\blfootnote[1]{
  \begingroup
  \renewcommand\thefootnote{}\footnote{#1}
  \addtocounter{footnote}{-1}
  \endgroup
}
\newcommand{\auth}[2]{\textbf{#1}~#2}

%% file: pages/intro.tex
\section{Introduction}
For centuries, scientific discovery has advanced through a cycle of \emph{observation}, \emph{hypothesis generation}, and \emph{experimental validation}. From \citet{Kepler1609} inferring planetary motion laws from Brahe's charts to epidemiologists linking smoking with lung cancer \citep{wynder1950tobacco}, progress has always hinged on formulating and testing hypotheses from observational data. This cycle is not unique to theory driven sciences and underpins modern, data driven fields such as marketing, psychology, and behavioral sciences as well. However, in the latter, observations are drawn from large, noisy, and context-dependent corpora of human activity where controlled experimentation is expensive and rarely possible.
The combinatorial space of plausible explanations, combined with small effect sizes, selection biases, and unmeasured confounders, makes distinguishing genuine effects from spurious correlations fundamentally harder than in controlled physical sciences.

The growing abundance of digital data presents an unprecedented opportunity : vast  observations could yield richer and novel insights, but manual hypothesis exploration cannot scale to this high-dimensional, noisy corpora. This motivates our central challenge:
 \emph{How can we automatically generate hypotheses suitable for intervention in data-driven sciences?}

Consider the simple research question ``What makes a counterargument persuasive?'', which we will use as a running example throughout the discussion. A scientist studying online discourse might begin by observing successful debates and hypothesising that ``Longer counterarguments are more persuasive''. To test this, they construct features and run statistical tests (e.g., t-tests) to measure significance. Suppose the hypothesis proves significant but with a modest effect size. While examining the results, the scientist notices that longer arguments tend to include more citations and external evidence, raising the question: is length driving persuasion, or is it the supporting evidence? This observation prompts a follow-up experiment that controls for presence of citations as a covariate; the effect shrinks but remains significant, suggesting length contributes independently. The follow-up was only conceivable because of the first test. This iterative, adaptive process of proposing, testing, and refining hypotheses distinguishes robust findings from shallow correlations and builds the confidence needed for real-world interventions, such as A/B testing. However, each iteration requires considerable attention, time, and resources, making it essential to move forward only with hypotheses backed by strong statistical evidence.

Automating this cycle from observations to evidence and experiment backed intervention-ready hypotheses can reduce the process of discovery from months to hours in any domain where experiments can be simulated or deployed. This would fundamentally alter the scientific and economic landscape of multiple fields like human behaviour, social science, education, healthcare, marketing, media \& politics, and economics \citep{ludwig2024machine}; ranging from what makes an image memorable to what changes opinion in online discourse. As digital data like social media and behavioral logs continue to grow, so too does the space of hypotheses worth exploring, far exceeding what manual analysis can cover.

\textbf{Hypothesis Generation:} The iterative, adaptive structure of scientific inquiry stands in contrast to how hypothesis generation has evolved in machine learning. Recent LLM-based methods generate hypotheses directly in natural language, leveraging pretrained knowledge to propose interpretable explanations from unstructured data \citep{qiu2023phenomenal,zhou2024hypothesis}. This capability increases interpretability and accessibility to researchers across domains.
However, these methods verify hypotheses only through predictive metrics on held out examples, making them prone to spurious correlations as they do not control for covariates, test under varied conditions, or measure effect sizes with appropriate statistical tests. To emphasize this we show how a hypothesis from previous methods about partisan speeches from US Congress (``I ask unanimous consent" predicts Republican speakers) is actually confounded with the majority party during the session, and its effect reverses in different sessions. However due to the imbalance of majority party in the speeches, this hypothesis turns out to be significant with a high accuracy for both methods.
Moreover, because hypotheses are generated in a single pass, these methods cannot discover multi-variable or conditional hypothesis, we provide many such hypotheses discovered by us in \S\ref{subsec:Validation}.

\textbf{Hypothesis Validation:} Interestingly, another line of work explores automated hypothesis testing of natural language hypotheses through LLMs, where given a dataset, candidate hypotheses are decomposed into code-based experiments with formal significance tests sequentially \citep{huang2025automated,agarwal2025open}.

Unlike the hypotheses generation paradigm, where methods propose a novel feature and its relationship with the outcome, these methods aim to test whether a hypothesis is empirically supported for a predefined set of features and outcome. This requirement heavily narrows the scope of applicable problems: most real-world datasets like social media posts, news, discussions, and reviews do not come with predefined feature sets. More importantly, as our earlier example illustrates, which variables matter often emerges adaptively from prior experiments. Neither paradigm supports the iterative, adaptive, experimentally grounded process that characterizes how scientific knowledge actually accumulates. The juxtaposition of these methodologies, exposes a fundamental dichotomy and motivates our central question:
\vspace{-0.5em}
\begin{quoting}
\emph{Can we automate the iterative and adaptive cycle of scientific discovery by unifying hypothesis generation and experimental validation over unstructured data?}
\end{quoting}
\vspace{-0.5em}

We present \textsc{ExperiGen}, the first framework that closes the loop between hypothesis generation and experimental validation directly over unstructured data. ExperiGen orchestrates two specialized LLM agents: a \emph{Generator} that proposes testable hypotheses, and an \emph{Experimenter} that plans features, covariates, and statistical tests, producing a structured analysis. The agents communicate through multi-turn conversation and are employed in a two-phase search. In the \emph{outer loop}, the Generator proposes a seed hypothesis that is both plausible and novel relative to previously validated discoveries. In the \emph{inner loop}, the Experimenter evaluates the hypothesis and returns its analysis; the Generator uses this feedback to reformulate features, refine context, or test combined effects iterating until the hypothesis either attains statistical significance (subject to Bonferroni correction) or is rejected. Validated hypotheses accumulate and condition subsequent proposals, drawing on Bayesian Optimization to efficiently allocate expensive evaluations: yielding faster, broader, and more reliable discovery. The decoupled agent architecture, feature extraction, and code execution allows us to easily generalize hypothesis generation and testing across modalities (text, images), and complex datasets (non iid) like Reddit threads ( detailed discussion in \cref{sec:appendix:architectural_flexibility}).

Empirically, ExperiGen yields hypotheses that are both more predictive and individually reliable compared to prior methods. ExperiGen outperforms existing hypothesis generation methods by 7-17 percentage points on average across 10 diverse tasks, with consistent gains on out-of-distribution evaluation (\cref{tab:main_benchmarks}), and discovers 2-4$\times$ more statistically significant hypotheses across domains, including visual and complex relational datasets. Crucially, adaptive discovery enables hypotheses that are inconceivable in a single pass: for instance, the insight that ``Expanding the OP's perceived decision space is persuasive'' emerges only after observing that both ``concessions" and ``framing problem as degree as opposed to binary" are persuasive,(\cref{fig:architecture_outer_loop}). On a manual review by data scientists, ExperiGen exhibits a false discovery rate below 5\%, compared to 20-25\% for prior methods (\cref{subsec:Validation})

Scientific discovery demands hypotheses to be novel, empirically supported, and impactful to be considered for intervention. To test this, we conducted the first expert review, in which professors with publications in top journals, 5–20 years of experience, and experience mentoring multiple graduate students provided feedback. We collecting annotations over 25 hypotheses about persuasion in online discourse, 88\% were considered moderately or strongly novel, 70\% were considered impactful and worth pursuing as research, and most demonstrated the experimental rigor of a senior graduate student. This motivated us to proceed to the next stage of scientific discovery: we conducted the first A/B experiment evaluating LLM-generated hypotheses about what influences form signups, in collaboration with a Fortune 500 consumer brand, and observed statistically significant results $(p<10^{-6})$ with a large effect size of $+344\%$ (details in \cref{sec:ab}). Together these results demonstrate that ExperiGen's hypotheses translate into intervention-ready discoveries beyond offline benchmarks. Our core contributions are:

\begin{enumerate}[leftmargin=*,itemsep=1pt]
\item \textbf{ExperiGen} We introduce the first, and a novel framework to unify hypothesis generation and validation.

\item \textbf{Algorithm.} We propose a two-phase search inspired by Bayesian Optimization that balances exploration (novelty-driven seed generation) with exploitation (iterative validation under statistical control), being the most scalable method for hypothesis generation.

\item \textbf{Rigor and Coverage.} ExperiGen , discovers most number of (2--4$\times$) hypotheses that are also more predictive (7-14 points in out-of-distribution samples) and empirically validated (lowest false discovery rate) while generalizing across text, images, and complex datasets like relational datasets effortlessly.

\item \textbf{User Study} We provide the first expert review from senior professors and also conduct an A/B test of automatically generated hypotheses, showing that ExperiGen’s discoveries are novel, impactful, and translate to measurable outcomes beyond offline benchmarks.
\end{enumerate}

\begin{figure}
    \centering
    \includegraphics[width=1\linewidth]{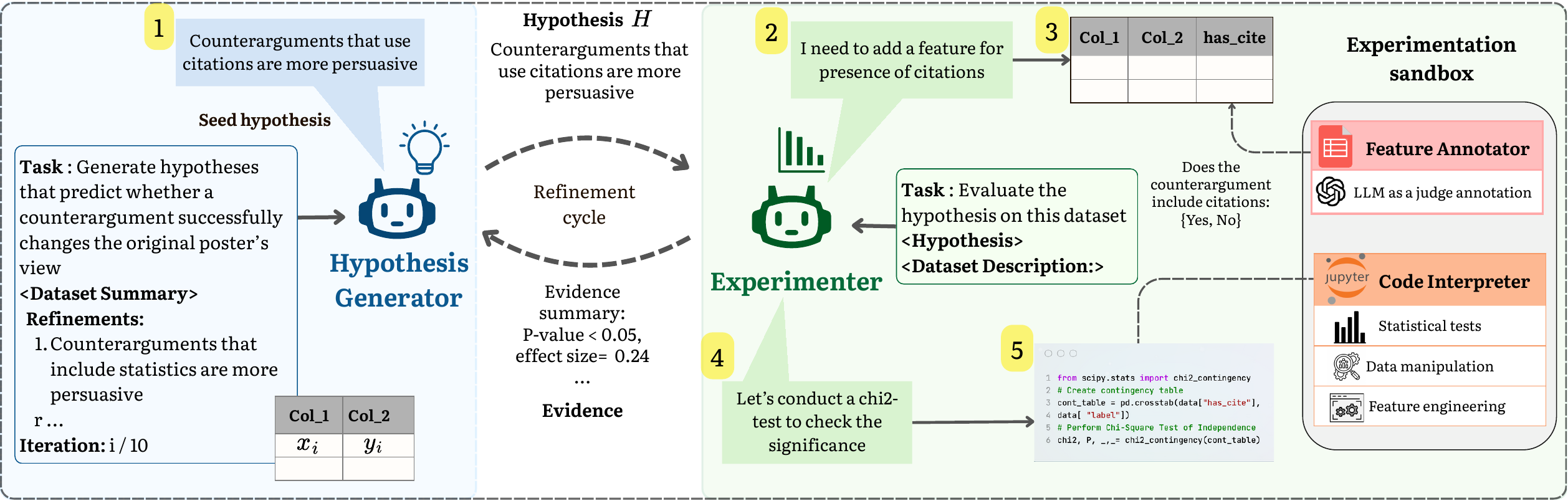}
    \caption{\textbf{Inner Loop: Iterative Refinement Cycle.} Given a dataset $\mathcal{D} = \{(x_i, y_i)\}$ and a seed hypothesis $H_{i,1}$, the system refines through $T$ steps. At refinement step $j$: (1)~The \emph{Generator} proposes hypothesis $H_{i,j}$ conditioned on short-term memory $\mathcal{M}_{i,j}$; (2)~The \emph{Experimenter} operationalizes the hypothesis by specifying required features; (3)~The \emph{Feature Annotator} augments $\mathcal{D}$ with the operationalized feature $f_H$ (e.g., \texttt{has\_cite}); (4)~The \emph{Code Interpreter} executes statistical tests on the augmented data; (5)~Evidence $E_H(\mathcal{D}) = (p, \delta)$ comprising p-value and effect size is returned. The memory $\mathcal{M}_{i,j} = \{(H_{i,k}, E_{H_{i,k}})\}_{k<j}$ accumulates hypothesis-evidence pairs from prior steps, enabling the Generator to propose refinements that address confounds or add contextual qualifiers. Hypotheses passing Bonferroni-corrected significance ($p < \alpha/T$) are candidates for the hypothesis bank $\mathcal{H}$. See \cref{fig:architecture_outer_loop} for the outer loop that orchestrates multiple refinement cycles. The exact prompt for the generator and experimenter are in \cref{subsec:Agent Prompts}}
    \label{fig:architecture}
\end{figure}

%% file: pages/method.tex
\vspace{-1.5em}
\section{Methodology}
\label{sec:methodology}
Scientific discovery proceeds by proposing hypotheses, testing them empirically, and refining them based on observed outcomes. Our goal is to automate this cycle for unstructured data. The core challenge is search: for a fixed dataset, the space of possible hypotheses is vast, yet only a small subset admit empirical support.
\textsc{ExperiGen} addresses this challenge through two specialized agents: a \textbf{Gen}erator that proposes plausible hypotheses and an \textbf{Experi}menter that evaluates them statistically.
The two agents employ a two-phase iterative search inspired by Bayesian Optimization, alternating between exploring novel regions of the hypothesis space and refining promising candidates, thereby accumulating an expanding set of validated hypotheses. In this section, we formalize the problem statement (\cref{sec:problem}), introduce the two agents (\cref{sec:generator,sec:experimenter}), and describe the two-phase search (\cref{sec:two_phase}). \cref{fig:architecture,fig:architecture_outer_loop} provide an overview of the inner refinement loop and outer acquisition-guided exploration, respectively.

\subsection{Problem Formulation}
\label{sec:problem}
Let $\mathcal{D} = \{(x_i, y_i)\}_{i=1}^n$ denote observation and outcome pairs, where $x_i$ is unstructured data (e.g. counterarguments, response time) and $y_i$ is the target variable (success in changing opinion).
A hypothesis $H\in\mathbb{H}$ (e.g. ``Using citations improves success rate") is a testable natural language statement that asserts a relationship between constructs derived from $x$ and the outcome $y$.

For a hypothesis to be empirically testable, there must exist an evaluation or experiment $E_H$ such that $E_H(\mathcal{D}) \in \texttt{\{supported, unsupported\}}$ indicates whether the hypothesis is supported or not. A rigorous evaluation requires (i) operationalization of constructs as features, (ii) identification and control of covariates, (iii) selecting and executing aligned tests, and (iv) robustness and sensitivity analyses with assumptions and specifications.

The space of such testable hypotheses $\mathbb{H}$ is combinatorially large: there exists multiple constructs which can be refined, combined, conditioned on contexts, and related in multiple ways. However, only a small subset of such hypotheses are empirically supported.
Data-driven hypothesis discovery seeks a maximal and diverse set $\mathcal{H} \subseteq \mathbb{H}$ such that $\forall H \in \mathcal{H}$, $E_H(\mathcal{D}) = \texttt{supported}$.

\subsection{The Generator Agent}
\label{sec:generator}

Exhaustive enumeration of the hypothesis space is infeasible. Large language models encode implicit beliefs about which scientific relationships are likely to hold, derived from pretraining on scientific literature, domain knowledge, and general reasoning patterns. We assume that, given only a problem description, an LLM induces an initial implicit distribution $q(H)$ over the hypothesis space $\mathbb{H}$, reflecting prior plausibility. Prompting an LLM to generate hypotheses can therefore be viewed as sampling from this induced distribution $H\sim q(H)$. Prior work has shown that such generations are strong estimates of plausibility in zero and few-shot settings \citep{liu2025_hypobenchsystematic_benchmarks}, we also observe this empirically (\cref{tab:main_benchmarks}). Therefore, we employ the Generator as an LLM-based agent that proposes candidate hypotheses conditioned on two inputs: (1) a structured summary of the dataset, including schema, feature distributions, and sampled observations; and (2) a description of the Experimenter’s evaluation capabilities. Conditioning on these inputs restricts generation to hypotheses that are both plausible and testable by the Experimenter on the given dataset. We detail this dataset summary design and Generator prompts in \S\ref{sec:appendix:dataset_description} and \cref{subsec:Agent Prompts}.

\subsection{The Experimenter Agent}
\label{sec:experimenter}

The \textbf{Experimenter} evaluates a natural-language hypothesis $H$ against a dataset $\mathcal{D}$ by constructing and executing an empirical evaluation procedure $E_H:\mathcal{D}$. We implement the Experimenter as a ReAct agent \citep{react2022} that plans features, covariates, and statistical tests conditioned on the hypothesis and dataset. It produces a structured analysis containing the chosen test, assumptions, significance levels, effect sizes, and robustness checks, applying a multiple-testing correction (e.g., FDR control) when evaluating families of related hypotheses. This makes $E_H$ reproducible given the same random seed, dataset, and tool outputs.

The Experimenter operates in a sandboxed environment with two tools: (i) a customized \textbf{Code Interpreter} tool for data processing, feature computation, and statistical testing; and (ii) an \textbf{LLM-based feature extractor} for semantic constructs not reliably derivable through programmatic rules e.g., ``presence of citations counterargument". The code tool handles computable features (length, counts, regex patterns); the LLM extractor handles higher-level judgments.

\begin{figure}[ht!]
    \centering
    \includegraphics[width=0.9\linewidth]{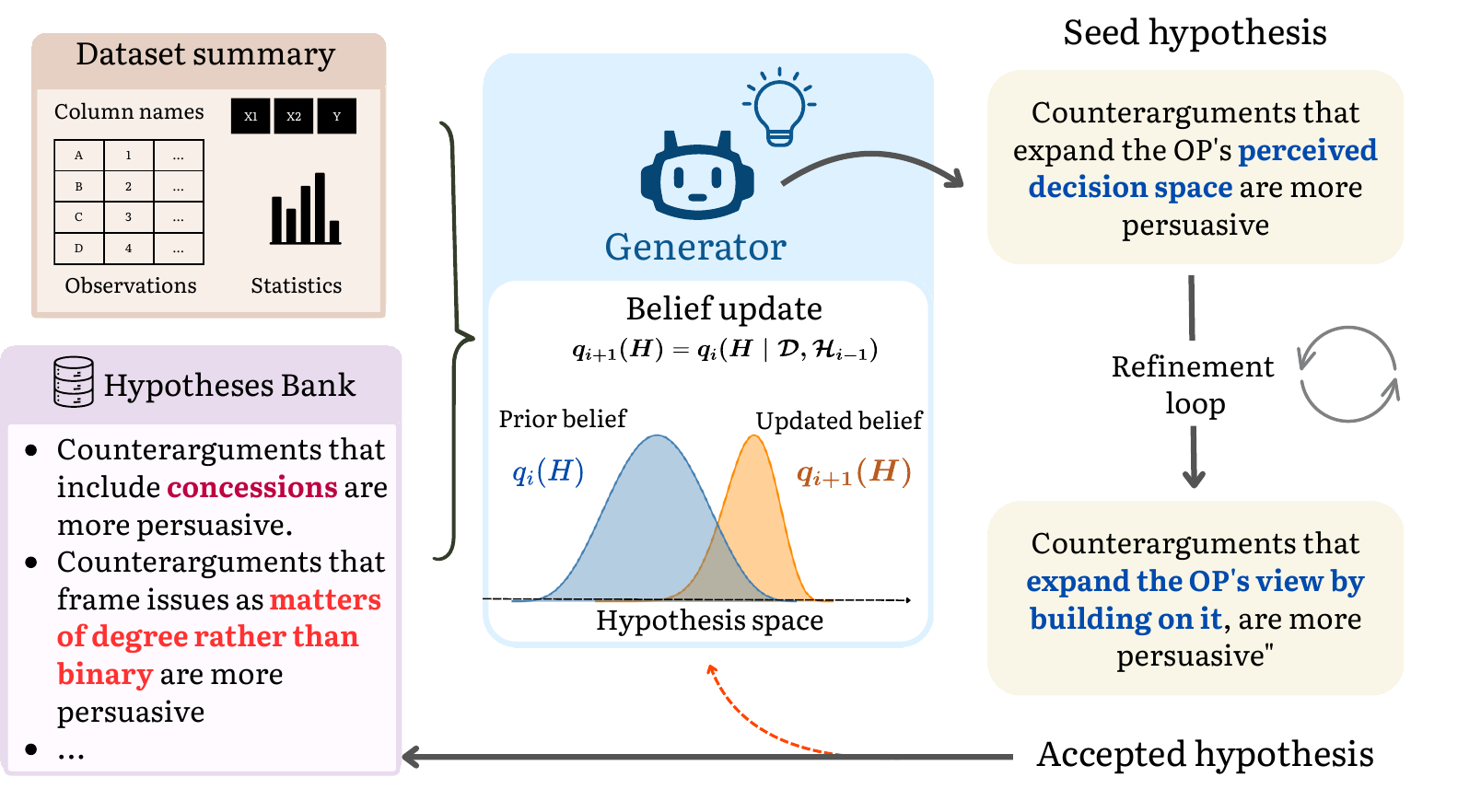}
    \caption{\textbf{Outer Loop: Acquisition-Guided Exploration.}
At each outer iteration $i \in \{1,\ldots,N\}$, the Generator conditions on a dataset summary $\mathcal{D}$ and the current hypothesis bank $\mathcal{H}_{i-1}$ to induce a proposal distribution \textcolor{blue}{\textbf{(shown in blue)}}
$q_i(H)=q(H \mid \mathcal{D}, \mathcal{H}_{i-1})$.
A seed hypothesis $H_{i,1}$ is sampled to implicitly maximize the acquisition objective
$\mathcal{A}(H)= s_i(H)+\mathcal{N}(H,\mathcal{H}_{i-1})$, which balances quality and novelty.
The seed enters the refinement loop (\cref{fig:architecture}), returning validated hypotheses that are added to $\mathcal{H}_i$.
The resulting belief update is visualized as a distribution shift \textcolor{orange}{\textbf{(shown in orange)}}, reflecting how accepted hypotheses reshape the Generator’s implicit prior in subsequent iterations.
For example, after observing hypotheses about \textbf{\textcolor{purple}{concessions}} and framing as \textbf{\textcolor{red}{degree vs.\ binary}}, the Generator begins to surface a new latent feature, \textbf{\textcolor{blue}{decision space}}, in later proposals.}
\vspace{-2mm}
    \label{fig:architecture_outer_loop}
\end{figure}

\subsection{Two-Phase Search}
\label{sec:two_phase}
The preceding sections establish two capabilities: a Generator that samples plausible, testable hypotheses from an implicit distribution $q(H \mid \mathcal{D})$, and an Experimenter that evaluates any hypothesis $H$ through a rigorous but computationally expensive procedure $E_H:D$.

Naively sampling from $q(H)$ and evaluating each hypothesis independently wastes expensive Experimenter calls on redundant or implausible candidates, while greedy exploitation of early discoveries misses distant regions of the space. We require a search algorithm that (i)~efficiently allocates expensive evaluations toward promising hypotheses, (ii)~maintains broad coverage to avoid local optima, and (iii)~exploits the compositional structure of hypotheses, where validated discoveries inform refinements and related candidates. These motivate our two-phase procedure: an outer loop that samples diverse seed hypotheses guided by an acquisition objective, and an inner loop that refines each seed through iterative validation to minimize Type-I error.
\subsubsection{Acquisition-Guided Seed Generation}
\label{sec:outer_loop}
The outer loop (\cref{fig:architecture_outer_loop}) draws inspiration from Bayesian Optimization, where a surrogate model approximates an expensive black-box function and an acquisition function balances exploration with exploitation. Here, the Generator's implicit beliefs serve as the surrogate, the inner refinement loop constitutes the expensive evaluation, and each accepted hypothesis updates the Generator's context for subsequent iterations. At outer iteration $i \in \{1, \ldots, N\}$, the Generator receives the dataset summary $\mathcal{D}$ and the hypothesis bank $\mathcal{H}_{i-1}$. Conditioning on these inputs induces a proposal distribution $q_i(H) \triangleq q(H \mid \mathcal{D}, \mathcal{H}_{i-1})$, which we shape via structured prompting toward an acquisition objective:

\begin{equation}
\label{eq:acquisition}
\mathcal{A}\!\left(H;\,\mathcal{H}_{i-1}\right)
=
\underbrace{s_i(H)}_{\text{plausibility}}
+
\underbrace{\mathcal{N}(H, \mathcal{H}_{i-1})}_{\text{novelty wrt }H_i-1}\,.
\end{equation}

where $s_i(H)$ and $\mathcal{N}(H, \mathcal{H}_{i-1})$ are implicit plausibility and novelty scores reflecting the Generator's beliefs given prior discoveries. The plausibility term shifts proposal mass toward $\mathcal{H}^*$ by leveraging patterns in validated hypotheses the novelty term provides an exploration bonus for candidates distant from existing discoveries.
The Generator outputs a seed hypothesis $H_{i,1}$ that implicitly maximizes $\mathcal{A}$.

\subsubsection{Local Refinement}
\label{sec:inner_loop}
The seed $H_{i,1}$ enters the refinement loop (\cref{fig:architecture}) for up to $T$ steps, indexed by $j = 1, \ldots, T$. At each step, the Experimenter evaluates $H_{i,j}$ and returns structured analysis $E_{H_{i,j}}(\mathcal{D})$ comprising p-value, effect size, and guidance for refinement. Accepted hypotheses are appended to the bank.

\paragraph{Inner loop:} For $j > 1$, the Generator's context shifts. Let $\mathcal{M}_{i,j} = \{(H_{i,k}, E_{H_{i,k}})\}_{k=1}^{j}$ denote the short-term memory containing session hypotheses and their analyses. The Generator now receives $(\mathcal{D}, \mathcal{M}_{i,j})$, and its objective reverses: rather than maximizing global novelty, it restricts search to the locality of $H_{i,1}$, proposing refinements that address identified confounds, add contextual qualifiers, or specify feature boundaries. The implicit goal of this local search is to subject the hypothesis family to rigorous falsification, thereby minimizing Type-I error. This contrasts with methods that optimize predictive accuracy across independently generated hypotheses, an objective that inflates false positives under multiple testing. We quantify this advantage through expert verification in \cref{subsec:Validation}.

\vspace{-3mm}
\paragraph{Hypothesis Bank Management.} The bank $\mathcal{H}_i$ accumulates validated discoveries across outer iterations. To maintain diversity and bound computation, we enforce a maximum capacity $K$. When $|\mathcal{H}_i| > K$, we retain only the top-$K$ hypotheses by pairwise novelty: we compute embeddings $\phi(H)$ (OpenAI \texttt{text-embedding-large}) for each hypothesis and greedily select the subset maximizing minimum pairwise semantic distance. This ensures the bank represents diverse and maximally inform future seed generation.
\vspace{-3mm}
\paragraph{Hypothesis Selection}
At the conclusion of each inner loop, we apply Bonferroni correction across the $T$ refinement steps, accepting only hypotheses with $p < \alpha_{\mathrm{sig}} / T$. Among accepted hypotheses, we select the one with largest support for inclusion in $\mathcal{H}$.

%% file: pages/experiment.tex
\section{Experiments}
\begin{table*}[ht!]
\centering
\begin{adjustbox}{max width=\textwidth}
\begin{tabular}{cl|cccccccccc|cccccc}
\toprule[1.2pt]
\multirow{2}{*}{\textbf{Inference}} & \multirow{2}{*}{\textbf{Method}}
& \multicolumn{2}{c}{\textbf{Decep.}}
& \multicolumn{2}{c}{\textbf{News}}
& \multicolumn{2}{c}{\textbf{Dread.}}
& \multicolumn{2}{c}{\textbf{GPTgc}}
& \multicolumn{2}{c|}{\textbf{Pers.}}
& \multicolumn{2}{c}{\textbf{Twitter}}
& \textbf{Design}
& \textbf{LaMem}
& \textbf{Cong.}
& \textbf{CMV} \\
\cmidrule(lr){3-4} \cmidrule(lr){5-6} \cmidrule(lr){7-8} \cmidrule(lr){9-10} \cmidrule(lr){11-12} \cmidrule(lr){13-14}
\textbf{Model} & & In & Out & In & Out & In & Out & In & Out & In & Out & In & Out & & & & \\
\midrule
\multirow{5}{*}{GPT-4o}
& ExperiGen & \valbest{78} & \valbest{85.2} & \valbest{70} & \valbest{66.5} & \valbest{74} & \valgood{80.8} & \valbest{88.8} & \valbest{85} & \valbest{94} & \valbest{91.2} & \valbest{67} & \valbest{66.1} & \valbest{88.0} & \valbest{60.1} & \valbest{79.4} & \valbest{77.5} \\
& HypoGenic & \valgood{76} & \valgood{77.7} & 63.2 & 62.5 & 64.2 & 67.2 & \valgood{87} & \valgood{84} & \valgood{93} & \valgood{89} & 60.3 & \valgood{61.9} & {84.25} & 50.8 & 72.6 & 61 \\
& HypotheSAEs & 62.9 & 60.6 & 61.9 & 61.4 & 60.9 & 62.3 & 76.9 & 79.8 & 87.1 & 84 & 54.1 & 56 & 84.25 & 49.2 & \valgood{73.9} & 59.2 \\
\cmidrule(lr){2-18}
& 0-shot CoT & 65.0 & 69.4 & \valgood{68.2} & \valgood{64.9} & 63.4 & 68.2 & 79.0 & 78.3 & 83.2 & 81.9 & \valgood{60.5} & 56.4 & \valgood{84.75} & 51.6 & 73.7 & 65 \\
& few-shot CoT & 64.8 & 75.8 & 67.2 & 62.9 & \valgood{73.4} & \valbest{81.6} & 77.0 & 78.3 & 81 & 80.5 & 58.6 & 59.2 & 84.60 & \valgood{51.9} & 72.6 & \valgood{69} \\
\midrule[1pt]
\multirow{5}{*}{Qwen3}
& ExperiGen & \valbest{69} & \valbest{75} & \valbest{59.1} & \valbest{61.5} & \valbest{70.1} & \valgood{70.6} & \valbest{73} & \valbest{77.4} & \valbest{88.1} & \valbest{89.2} & -- & -- & -- & -- & -- & -- \\
& HypoGenic & \valgood{65.1} & 66 & 57.2 & \valgood{60.1} & 58.9 & 63.1 & 71.1 & 73.9 & 80.6 & \valgood{82.7} & -- & -- & -- & -- & -- & -- \\
& HypotheSAEs & 52.4 & 61.17 & 55.4 & 56.3 & 61.4 & 64.4 & 68 & 71 & 78.9 & 80.1 & -- & -- & -- & -- & -- & -- \\
\cmidrule(lr){2-18}
(32B)& 0-shot CoT & 61.6 & 64.4 & \valgood{58.6} & 60.5 & 62.6 & 65.4 & 55.3 & 64.7 & \valgood{81.3} & 79.1 & 52.4 & 51.4 & -- & -- & -- & -- \\
& few-shot CoT & 62.0 & \valgood{70.8} & 55.2 & 58.3 & \valgood{65.8} & \valbest{75.8} & \valgood{72.3} & \valgood{76.3} & 77.9 & 78 & 54.4 & 53.2 & -- & -- & -- & -- \\
\bottomrule[1.2pt]
\end{tabular}
\end{adjustbox}
\caption{Predictive accuracy on HypoBench (5 tasks) and Cross-Domain benchmarks (5 tasks). We report in-domain (In) and out-of-domain (Out) accuracies using hypotheses generated by each method. ExperiGen achieves the highest accuracy on 9 of 10 task-split combinations. Cross-domain tasks include Twitter (social text), Design preference, Image Memorability (LaMem), Congress (large-scale, 40k examples), and CMV (persuasion). For reference, task-specific expert models achieve: Twitter 80.9/77.3 (In/Out) using a fine-tuned Vicuna-1.5 (13B)~\citep{singh2024measuring}, Design 87.2 following~\citet{patnaik2025aesthetiq}, and LaMem 75.0 using a fine-tuned 13B model~\citep{harini2025long}.}
\label{tab:main_benchmarks}
\end{table*}

We evaluate ExperiGen on a diverse suite of \emph{10 tasks} spanning text-only, multimodal, and metadata-rich settings. Our evaluation begins with an existing benchmark and extends to more challenging scenarios which we call hard set (HS) that probe three key dimensions: (i) \emph{scalability} to large datasets (Congress, 40k examples), (ii) \emph{complex metadata} where hypotheses must exploit rich covariates (Twitter, CMV), and (iii) \emph{multimodality} requiring reasoning over visual inputs (Design, Memorability).

\subsection{Tasks and Datasets}

\textbf{HypoBench :}
We first evaluate ExperiGen on \emph{HypoBench} \citep{liu2025hypobenchsystematicprincipledbenchmarking}, the standard benchmark for principled hypothesis generation. We use five of its real-world tasks: (i) Deception Detection, (ii) AI-Generated Content Detection, (iii) Persuasive Argument Prediction, (iv) Mental Stress Detection, and (v) News Headline Engagement. Each task provides in-distribution (ID) test sets, and out-of-distribution (OOD) test sets to evaluate robustness against domain shifts. Further details on these tasks are in Appendix~\ref{sec:Datasets}.

While HypoBench provides a standardized evaluation, its tasks are relatively small ($<$1k examples), contain only raw text without structured metadata or covariates, and are solvable with moderate accuracy by prompting alone. To test in more realistic settings, we introduce five additional tasks that probe scalability, use of metadata, and multimodality.

\textbf{Congress:} U.S. congressional speeches from 2005--2007 \citep{gentzkow2010drives}, where the task is to predict party affiliation (Republican or Democrat) from speech text. We use 40k training, 5k validation, and 5k test examples.

\textbf{Twitter:} Pairs of topic-controlled tweets with rich covariates (usernames, tags, metadata) \citep{singh2024measuring}, where the task is to predict which tweet is more persuasive. We use 5k training, 1k validation, and 2k test samples.

\textbf{CMV:} Reddit r/ChangeMyView posts \citep{tan2016winning}, where the task is to predict whether a counterargument successfully changes the original poster's view. Metadata include: time of posting, time of response, author flairs, etc. We use 4k training and 500 test examples.

\textbf{Design:} Paired graphic layouts from AesthetiQ \citep{patnaik2025aesthetiq}, where the task is to identify which layout was AI-generated. We use 1.6k training, 200 validation, and 200 test examples.

\textbf{Memorability :}We use the LaMem dataset \citep{khosla2015} to form pairs of visually similar images with significantly different memorability scores. The task is to predict which image is more memorable. We use 8k training pairs and 1k each for validation and testing.

\subsection{Evaluation Setup}

We evaluate hypothesis generation methods along three axes:
\vspace{-2mm}
\paragraph{Predictive Performance.} Following \cite{zhou2024hypothesis}, our primary metric is \emph{classification accuracy} on held-out test sets. We report both in-distribution (ID) and out-of-distribution (OOD) accuracy to assess generalization.

\paragraph{Hypothesis Quantity and Statistical Validity.} We report the number of hypotheses each method generates and the subset that pass statistical validation. A hypothesis is considered \emph{statistically valid} if its associated features show significant predictive power ($p < 0.05$, Bonferroni-corrected) in a multivariate regression on held-out data. Methods that generate many hypotheses but few valid ones suffer from high false discovery rates. We report the false discovery rate (FDR) as the fraction of accepted hypotheses rejected by experts.

\paragraph{Expert Review.}  To assess whether generated hypotheses meet publication-level standards, we conduct an expert study where domain experts rate hypotheses on four dimensions: (1)~\emph{research-worthiness}, whether they would recommend pursuing the hypothesis as a serious research project, (2)~\emph{novelty} relative to existing literature, (3)~\emph{quality} in terms of clarity, specificity, and testability, and (4)~\emph{research expertise} rigor reflected in the experimental design. Full study details are in Appendix~\ref{sec:expert_study}.

\subsubsection{Inference Procedure}
We evaluate two inference methods and report the best-performing one for each baseline:
\vspace{-2mm}
\begin{enumerate}[leftmargin=*,itemsep=1pt]
  \item \textbf{Two-Step LLM Inference:} Following \citep{zhou2024hypothesis}, we decompose inference into two steps: (i)~\textbf{Selection}: given an input, the LLM retrieves the top-\(k\) most relevant hypotheses from the bank (\(k{=}3\));
  (ii)~\textbf{Prediction}: the LLM produces the final output conditioned only on the selected hypotheses and the observation.

  \item \textbf{Logistic Regression:} Following \citet{movva2025sparse}, we fit a multivariate logistic regression using all hypothesis-derived features and report test accuracy.

\end{enumerate}
\vspace{-2mm}
\subsection{Baselines}

We compare \textsc{ExperiGen} against two state-of-the-art hypothesis generation methods,
HypoGenic \citep{zhou2024hypothesis} and HypotheSAEs \citep{movva2025sparse},
and against Chain-of-Thought (CoT) prompting to benchmark the performance of strong
LLM-only baselines in zero-shot and few-shot ($k=3$) settings, following prior work
\citep{zhou2024hypothesis}. For hyperparam details on the baselines, see Appendix~\ref{sec:hyp_baseline}.

We use Qwen3-32B as the default Generator and Experimenter for \textsc{ExperiGen}, Qwen3-32B as the LLM for HypoGenic, and OpenAI's text-embedding-3-large for embeddings with Qwen3-32B for hypothesis proposal and annotation in HypotheSAEs. We report inference results using both GPT-4o and Qwen3-32B to evaluate cross-model transfer.

%% file: pages/results_discussion.tex
\section{Results and Discussion}

We evaluate ExperiGen across four dimensions: \emph{predictive performance}, \emph{statistical validity and novelty}, \emph{real-world A/B test impact}, and \emph{ablations}. Ablations for all architectural choices are reported in Appendix~\ref{sec:addl_tables}.

\subsection{Predictive Performance}

\paragraph{ExperiGen achieves best performance on 9/10 benchmark settings.}
Table~\ref{tab:main_benchmarks} reports accuracy on the HypoBench suite across both \emph{in-domain} (ID) test sets and \emph{out-of-distribution} (OOD) test sets that differ in source, time period, or domain. ExperiGen achieves the highest accuracy on 9 of 10 task-split combinations, with 6--17\% relative improvement over hypothesis generation baselines. The consistent OOD advantage suggests that iterative refinement filters out spurious correlations. ExperiGen also outperforms zero-shot and few-shot prompting by 10--12\%, indicating that the discovered hypotheses provide predictive value beyond the LLM's prior knowledge.
\vspace{-3mm}
\paragraph{ExperiGen generalizes to hard tasks.}
On the hard set (Table~\ref{tab:main_benchmarks}), ExperiGen demonstrates consistent gains across scale, modality, and metadata complexity. ExperiGen outperforms baselines on all 5 tasks and 6 test sets by 6-8\% and 0-shot/few-shot settings by 5\%.  For example, on CMV, ExperiGen discovered that being among the first responders to a post doubles persuasion success (19.9\% vs.\ 9.5\%). Testing this hypothesis requires computing \emph{within-thread rankings}, features that depend on relationships across samples rather than properties of individual comments---infeasible without the code interpreter. On Image memorability (LaMem) ExperiGen is the \emph{only} method to surface any significant memorability hypotheses, discovering patterns such as ``images with human faces in unexpected contexts are more memorable"

\subsection{Statistical Validity and Novelty}
\label{subsec:Validation}

\paragraph{ExperiGen discovers more statistically significant hypotheses.}
Statistical validity is important because unvalidated hypotheses can drive costly downstream decisions, including failed clinical trials, A/B tests, and policy interventions. A hypothesis may not improve multivariate predictive performance (particularly under finite samples or correlated features) yet still be informative if it captures a distinct, validated association. To quantify this, we fit a multivariate regression over all hypothesis-derived features and count the number with statistically significant coefficients (\(p < 0.05\), Bonferroni-corrected). This evaluation penalizes redundant hypotheses while favoring those with independent explanatory effects.

Table~\ref{tab:sig_hypotheses} summarizes the count of significant hypotheses discovered across all datasets. ExperiGen discovers 2--4$\times$ more significant hypotheses than baselines. This advantage stems from two factors. First, \emph{adaptive exploration}: iterative refinement learns from rejected hypotheses and steers toward promising directions, whereas baselines generate hypotheses independently and cannot correct course. Second, \emph{broader feature scope}: ExperiGen spans three feature types (syntactic, semantic, and relational), while baselines are limited to per-sample semantic features and cannot express relational hypotheses that require computation across samples (Appendix~\ref{sec:experimentation_sandbox}).

Beyond discovering more hypotheses, Experigen also \emph{rejects} spurious ones that baselines accept, directly reducing the false discovery rate (FDR). On Congress, HypotheSAEs reports ``I ask unanimous consent'' as one of its most predictive features for Republican speakers ($p{<}0.001$). However this is a dataset artifact because the term is primarily said by ``majority party" as opposed to Republican or Democrat and in the dataset 95\% of speeches belong to sessions with Republican as the majority. If deployed in practice (e.g., to flag partisan rhetoric), this hypothesis would fail because. Since 95\% of speeches come from Republican-majority sessions. ExperiGen discovers the same initial correlation, but through iterative refinement the Generator identifies session as a potential confounder and asks the Experimenter to control for it. Once session is included, the predictive power vanishes; testing on Democrat-majority sessions reverses the coefficient entirely. The hypothesis is rejected, for a complete walkthrough refer to (\cref{sec:case_congress}).

\vspace{-1em}
\paragraph{Domain experts validate that ExperiGen's hypotheses are novel, high-quality, and research-worthy.}
Statistical significance alone does not guarantee scientific validity a hypothesis may exploit dataset artifacts, restate well-known results, or lack the clarity required for actionable research. To assess whether ExperiGen's hypotheses meet the standards domain experts apply when evaluating research contributions, we conducted an expert study on the Change My View (CMV) persuasion task.

We recruited 5 senior faculty members with professorial appointments in computational social science, NLP, and persuasion research, with 5--20 years of experience publishing and reviewing at top venues in their fields. Each expert has mentored over 20 graduate researchers. Experts reviewed 25 hypotheses generated by ExperiGen on CMV, presented alongside their full experimental designs. Experts rated each hypothesis on four dimensions: (1)~\emph{research worthiness}: whether they would recommend pursuing the hypothesis as a serious research project; (2)~\emph{novelty}: whether the finding advances beyond existing literature; (3)~\emph{methodological rigor}: the level of sophistication reflected in the experimental design; and (4)~\emph{quality}: clarity, specificity, and testability of the hypothesis.

\begin{figure}[t]
    \centering
    \includegraphics[width=0.6\linewidth]{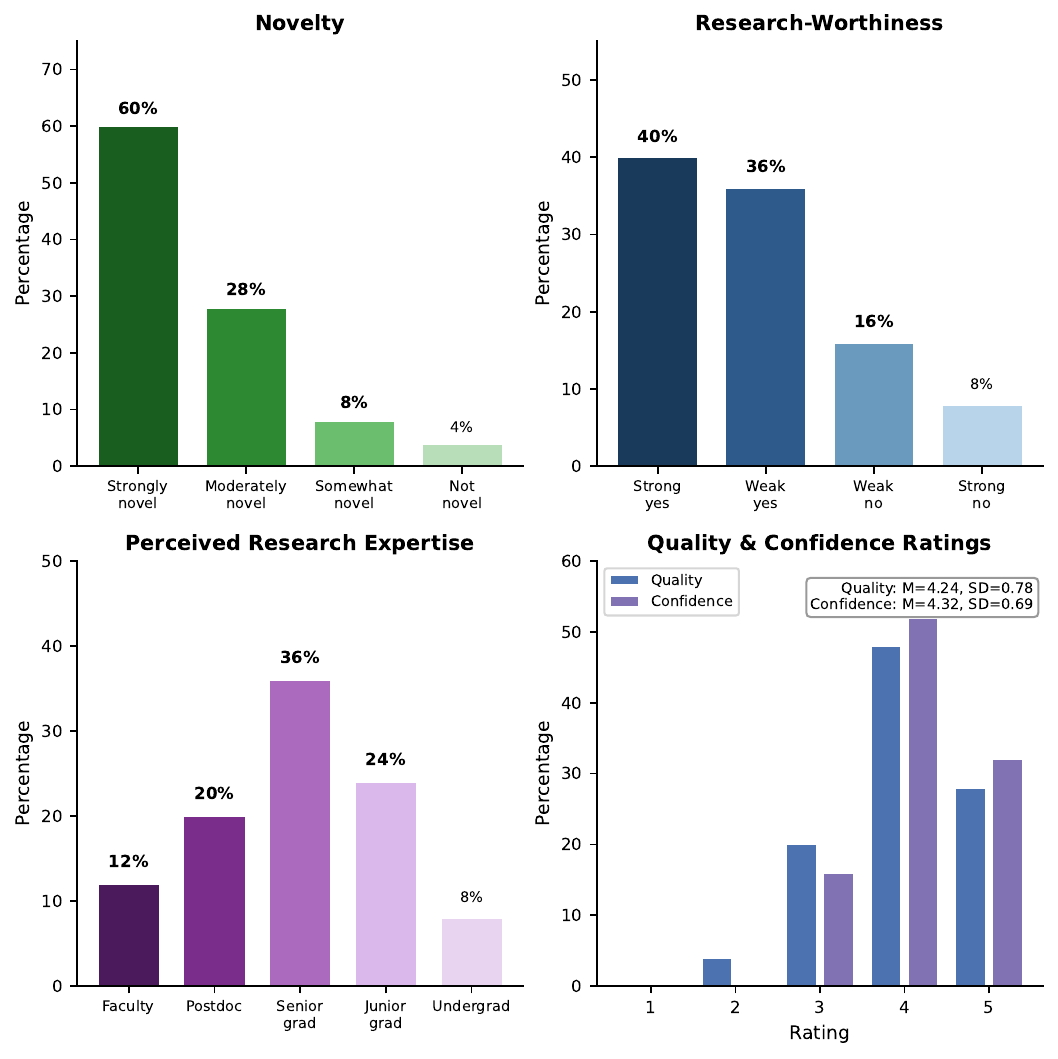}
    \caption{Expert evaluation results across four dimensions (N=25 annotations from 5 domain experts). Experts rated 88\% of hypotheses as novel, 76\% as research-worthy, and perceived the experimental designs as reflecting senior graduate student level expertise. }
    \label{fig:expert-evaluation}
    \vspace{-2em}
\end{figure}

As shown in Figure~\ref{fig:expert-evaluation}, experts perceived ExperiGen's hypotheses favorably across all dimensions. 88\% of hypotheses were rated as novel, indicating that the iterative refinement process surfaces findings beyond obvious or well-established results. 76\% were deemed research-worthy, with reviewers who declined citing concerns about intuitive effects rather than methodological flaws. The experimental designs were judged to reflect senior graduate student level expertise, with a mean quality rating of 4.24/5. One reviewer noted: ``The interaction effect between accuracy-motivating elements and identity-relevant claims is genuinely surprising---I would not have predicted this pattern from prior literature.'' Full study details are provided in Appendix~\ref{sec:expert_study}.

We extended this novelty assessment beyond CMV by conducting a literature search across all datasets. Approximately 40\% of ExperiGen's significant hypotheses were not documented in prior peer-reviewed work on the respective tasks.

\vspace{-2mm}

\subsection{Real-World A/B Test}
\label{sec:ab}

\paragraph{ExperiGen's hypotheses produce measurable impact in deployed A/B experiments.}
We partnered with a Fortune 500 consumer brand to validate a hypothesis generated by ExperiGen: \emph{``Forms that are horizontally centered and have a soft shadow effect achieve 3.8$\times$ higher conversion rates"}.

\textbf{Discovery cycle.} The Generator proposed an initial seed on Hypotheses \#12 (``forms in the footer of the webpage have low conversions") and refined it after EDA to include \textbf{layout} = $\{\text{horizontal centered, vertical centered, non centered}\}$ and \textbf{background} $= \{\text{sharp, soft-shadow}\}$ as key design features. The Experimenter extracted visual/structural features, ran EDA comparing form view rates by presentation and position, and fit a controlled logistic regression including position, webpage type(homepage or blog), and interaction effects. Coefficients for the hypothesized features remained positive and significant after controls leading to the aforementioned discovery.

\textbf{A/B experiment.} The online experiment randomized page traffic 50/50 between business-as-usual control and a challenger implementing the hypothesized design. The challenger produced 15.1k sign-ups versus 3.4k for control---a \textbf{+344\% uplift} in sign-ups. A two-proportion test indicated a highly significant effect ($p < 10^{-6}$). At a conservative cost-per-click of \$0.22, achieving equivalent sign-ups via paid traffic would cost an estimated \$29,570 over week. To our knowledge, this is the first deployment of automatically generated hypotheses with statistically significant impact.

%% file: pages/background.tex
\section{Additional Related Work}
\textbf{Hypothesis Generation} aims to discover insights from data. Classical methods like topic modeling or n-gram analysis produce statistical patterns that are not directly interpretable as human-readable hypotheses \citep{monroe2008fightin, blei2003latent}. A parallel effort uses LLMs to propose interpretable concepts for bottleneck models, guided either by a task description \citep{ludan2024concept} or by the activations of a trained neural network. Most relevant to our work is the direct use of LLMs for end-to-end hypothesis generation from unstructured data. This line of research, motivated by the capacity of LLMs for human-like inductive reasoning \citep{qiu2023phenomenal, tenenbaum2011grow}, aims to produce natural language statements (e.g., ``shorter tweets receive more likes''). However, leading approaches like \textbf{HypoGeniC} \citep{zhou2024hypothesis} and \textbf{HypotheSAEs} \citep{movva2025sparse} are fundamentally limited; they validate hypotheses using only predictive performance, which can be spurious, and struggle with complex, multimodal datasets. \textbf{Hypothesis Validation}, conversely, focuses on rigorously testing pre-existing ideas. Frameworks in this area use LLMs as coding assistants to perform statistical tests, but they cannot generate novel hypotheses from raw, unstructured data, as they require pre-specified, structured features to operate \citep{huang2025automated, agarwal2025open}. \textsc{ExperiGen} is the first framework to bridge this critical gap. Unlike generation-only methods, it moves beyond weak predictive metrics to ground hypotheses in formal statistical evidence. Unlike validation-only systems, it discovers these hypotheses directly from raw, unstructured data without needing pre-defined features. Our core contribution is the closed-loop refinement process, where statistical evidence directly guides the search for better hypotheses. This allows \textsc{ExperiGen} to unify discovery and validation, producing rigorously tested, interpretable insights from complex datasets where prior methods could only do one or the other.

%% file: pages/appendix.tex
\section*{Appendix}
\section{Additional Experiments and Ablations}
\label{sec:addl_tables}

We conduct systematic ablations to isolate the contribution of each component in \textsc{ExperiGen}. All experiments use $T=4$ refinement steps in the inner loop. We evaluate on two datasets selected to span the range of LLM prior knowledge: \textbf{News Headlines}, where zero-shot and few-shot baselines achieve similar performance (minimal gap), indicating the task is well-represented in pretraining data; and \textbf{Deceptive Reviews}, where few-shot substantially outperforms zero-shot (maximal gap), indicating the task requires learning from observations rather than relying on prior knowledge. This selection allows us to understand how \textsc{ExperiGen}'s components behave in both regimes---where pretrained knowledge suffices and where data-driven discovery is essential. We measure out-of-domain accuracy (\textbf{OOD Acc.}) and number of statistically significant hypotheses (\textbf{\#Sig.}).

\subsection{Ablation Configurations}

\subsubsection{Acquisition Objective.}
We ablate the two-phase search structure by modifying how seed hypotheses are generated. \textbf{No Acquisition} removes conditioning on the hypothesis bank $\mathcal{H}_{i-1}$; the Generator receives only the dataset summary $\mathcal{D}$, preventing it from building on prior discoveries. \textbf{Seed Only} further removes the inner refinement loop ($T=1$), testing whether iterative refinement adds value beyond seed diversity. Without acquisition-guided exploration, the Generator lacks awareness of previously validated hypotheses, leading to redundant proposals and reduced coverage. Seed Only compounds this by eliminating the refinement phase that allows hypotheses to be corrected based on experimental feedback. Comparing \textsc{ExperiGen} to these ablations reveals that the inner loop contributes $+4$--$13$ percentage points in OOD accuracy and approximately $2\times$ more significant hypotheses, confirming that refinement based on experimental feedback yields more robust discoveries.

\subsubsection{Novelty Constraint.}
We isolate the novelty term $\mathcal{N}(H, \mathcal{H}_{i-1})$ in the acquisition objective (Eq.~\ref{eq:acquisition}). \textbf{No Novelty} removes the novelty instruction from the Generator prompt, setting $\lambda = 0$. Without explicit novelty pressure, the Generator tends to fixate on variations of the top 2--3 most salient patterns in the data, repeatedly proposing minor reformulations rather than exploring distinct hypothesis families. This results in 40--45\% fewer significant hypotheses than the full model. However, No Novelty still outperforms No Acquisition because it retains conditioning on the hypothesis bank, allowing the Generator to avoid exact duplicates even if it cannot escape local optima.

\subsubsection{Experimentation.}
We ablate the Experimenter agent's role and feedback mechanism. \textbf{No Feedback} returns only accept/reject decisions without structured analysis $(p, \delta, \text{guidance})$, preventing the Generator from understanding why a hypothesis failed. \textbf{Single Pass} fixes $T=1$ with no iterative refinement, equivalent to independent hypothesis evaluation. \textbf{Accuracy Based} replaces statistical testing with held-out accuracy, accepting hypotheses if classifier accuracy exceeds 60\% on a validation split, mirroring the evaluation protocol of prior work~\citep{qiu2023phenomenal,zhou2024hypothesis}. The Accuracy Based variant achieves competitive OOD accuracy but discovers $3\times$ fewer valid hypotheses, consistent with our claim that predictive accuracy alone inflates false positives under multiple testing.

\subsubsection{Feature Extractor.}
We evaluate robustness to feature extraction noise by injecting label errors into LLM extractor outputs. \textbf{Noisy @ 10\%} flips 10\% of extracted binary labels uniformly at random; \textbf{Noisy @ 20\%} increases this to 20\%. These ablations simulate realistic annotation errors and reveal how hypothesis validation degrades under imperfect feature extraction. Performance degrades gracefully: 10\% noise reduces OOD accuracy by 2--3 points, while 20\% noise incurs a 4--5 point drop. The number of significant hypotheses decreases more sharply, as noisy features introduce measurement error that inflates p-values and obscures true effects.

\subsubsection{Search Strategy.}
We compare iterative refinement against parallel generation under matched compute. \textbf{Parallel} generates $N \times T$ hypotheses simultaneously without sequential feedback, evaluates all independently, and selects the top-$K$ by effect size. Total LLM calls are matched to \textsc{ExperiGen}. Under matched compute, iterative refinement outperforms parallel generation by $+5$--$8$ points in OOD accuracy, validating that adaptive search exploits experimental feedback more effectively than independent sampling.

\subsubsection{Observation Sampling.}
We ablate how observations are sampled for the Generator's dataset summary; see \cref{fig:observation} for iteration-wise results. \textsc{ExperiGen} uses uniform random sampling of 5 observations per iteration. \textbf{Boosting} identifies samples with the highest prediction error under the current hypothesis bank and samples in proportion to their errors, prioritizing difficult instances that existing hypotheses fail to explain. \textbf{Clustering} embeds all observations using \texttt{text-embedding-large}, partitions them via $k$-means clustering, and samples from different clusters to maximize diversity. \textbf{No Data} provides only the dataset schema without concrete examples.

\subsection{Ablation Results}

\begin{table}[t]
\centering
\small
\begin{tabular}{llcccc}
\toprule
& & \multicolumn{2}{c}{\textbf{News Headlines}} & \multicolumn{2}{c}{\textbf{Deceptive Reviews}} \\
\cmidrule(lr){3-4} \cmidrule(lr){5-6}
\textbf{Category} & \textbf{Condition} & \textbf{OOD Acc.} & \textbf{\#Sig.} & \textbf{OOD Acc.} & \textbf{\#Sig.} \\
\midrule
\multicolumn{2}{l}{\textit{Full Model}} \\
& \textsc{ExperiGen} & \textbf{65.2} & \textbf{18} & \textbf{85.1} & \textbf{16} \\
\midrule
\multicolumn{2}{l}{\textit{Acquisition Objective}} \\
& \makecell[l]{No Acquisition\\{\scriptsize(no conditioning on $\mathcal{H}_{i-1}$)}} & 61.5 & 11 & 78.2 & 10 \\[4pt]
& \makecell[l]{Seed Only\\{\scriptsize(no refinement, $T$=1)}} & 58.3 & 7 & 72.4 & 6 \\
\midrule
\multicolumn{2}{l}{\textit{Novelty Constraint}} \\
& \makecell[l]{No Novelty\\{\scriptsize(fixates on top 2--3 patterns)}} & 62.1 & 10 & 79.6 & 9 \\
\midrule
\multicolumn{2}{l}{\textit{Experimentation}} \\
& \makecell[l]{No Feedback\\{\scriptsize(accept/reject only)}} & 60.4 & 9 & 76.8 & 8 \\[4pt]
& \makecell[l]{Single Pass\\{\scriptsize($T$=1, no iteration)}} & 59.1 & 8 & 74.5 & 7 \\[4pt]
& \makecell[l]{Accuracy Based\\{\scriptsize(no statistical testing)}} & 62.7 & 5 & 80.3 & 5 \\
\midrule
\multicolumn{2}{l}{\textit{Feature Extractor}} \\
& \makecell[l]{Noisy @ 10\%\\{\scriptsize(10\% label noise)}} & \underline{63.8} & \underline{15} & \underline{82.4} & \underline{13} \\[4pt]
& \makecell[l]{Noisy @ 20\%\\{\scriptsize(20\% label noise)}} & 61.2 & 12 & 79.6 & 10 \\
\midrule
\multicolumn{2}{l}{\textit{Search Strategy}} \\
& \makecell[l]{Parallel\\{\scriptsize(no sequential feedback)}} & 60.2 & 10 & 77.1 & 9 \\
\midrule
\multicolumn{2}{l}{\textit{Observation Sampling}} \\
& \makecell[l]{Boosting\\{\scriptsize(error-weighted sampling)}} & 63.4 & 14 & 78.2 & 12 \\[4pt]
& \makecell[l]{Clustering\\{\scriptsize(max diversity via embeddings)}} & 62.8 & 15 & \underline{82.6} & 14 \\[4pt]
& \makecell[l]{No Data\\{\scriptsize(schema only)}} & 58.6 & 5 & 65.3 & 4 \\
\bottomrule

\end{tabular}
\caption{\textbf{Ablation Study Results.} OOD Acc. = Out-of-domain accuracy (\%); \#Sig. = Number of statistically significant hypotheses ($p < 0.05$ after Bonferroni correction). All experiments use $T=4$ refinement steps. Best results in \textbf{bold}, second-best \underline{underlined}.}
\label{tab:ablations}
\end{table}

\subsection{Detailed Component Analysis}
\label{sec:detailed_ablations}

The ablation table (\cref{tab:ablations}) summarizes how each design choice affects end-to-end performance. In this section, we provide deeper analysis of four components that warrant additional investigation: (1) how observation sampling affects hypothesis diversity over iterations, (2) how Generator and Experimenter model capabilities independently influence discovery quality, (3) how inference scales as the hypothesis bank grows, and (4) how reliably the LLM feature extractor annotates semantic constructs. Together, these analyses reveal where \textsc{ExperiGen}'s performance gains originate and identify bottlenecks for future improvement.

\subsubsection{Observation Sampling Dynamics}
\label{sec:obs_sampling}

The Generator receives a small batch of observations from $\mathcal{D}$ at each iteration. While \cref{tab:ablations} reports final accuracy, iteration-wise dynamics reveal important differences between sampling strategies. We compare four approaches: (1) \textbf{Random sampling} (\textsc{ExperiGen}'s default), which draws 5 observations uniformly; (2) \textbf{Boosting}, which measures prediction error for each sample under the current hypothesis bank and samples in proportion to error magnitude, prioritizing instances that existing hypotheses fail to explain; (3) \textbf{Clustering}, which embeds observations using \texttt{text-embedding-large}, partitions them via $k$-means, and samples from different clusters to maximize diversity; and (4) \textbf{No Data}, which provides only the dataset schema without concrete examples.

\begin{figure}[!t]
    \centering
    \includegraphics[width=\linewidth]{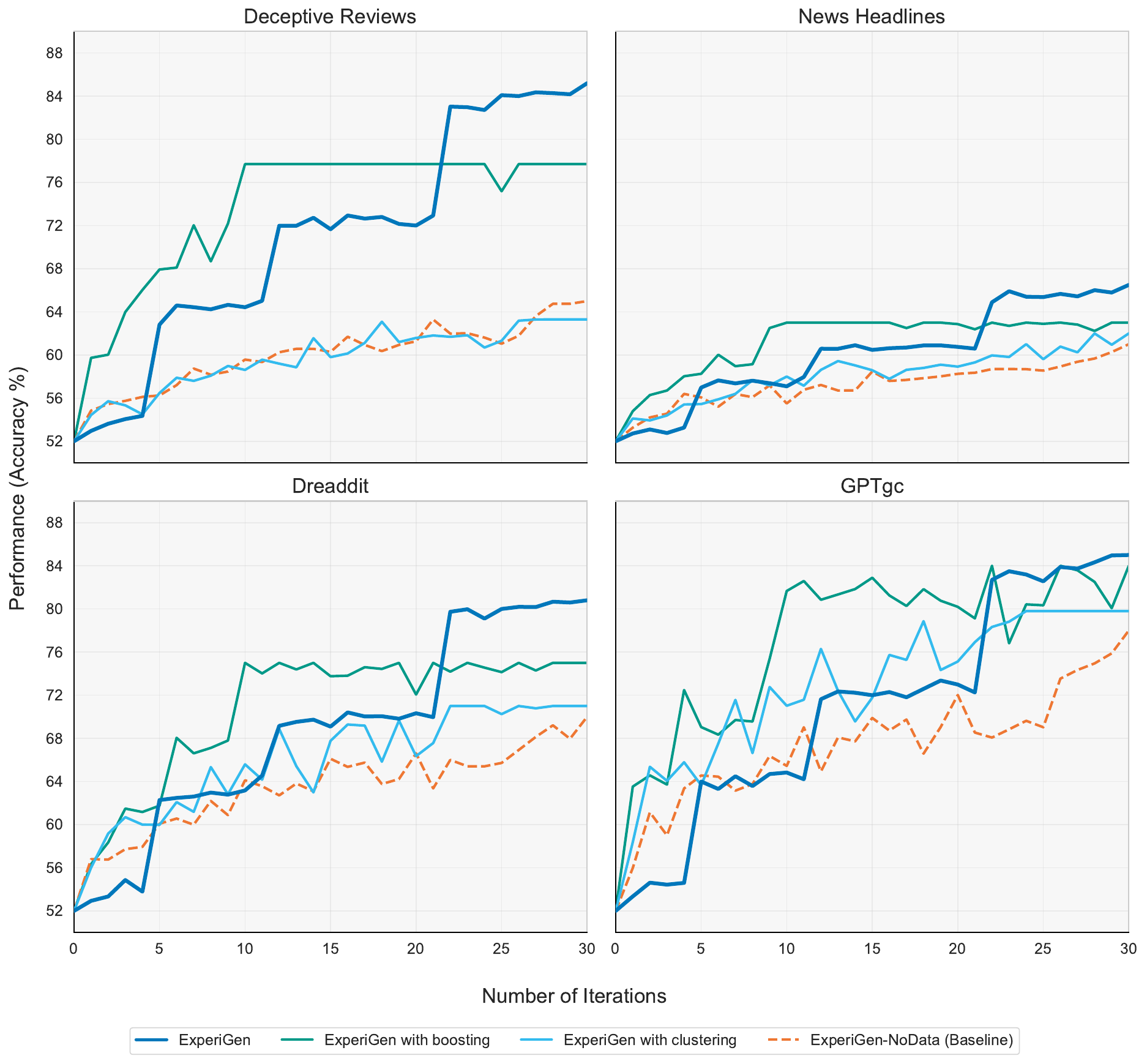}
    \caption{\textbf{Observation sampling dynamics across iterations.} Accuracy (y-axis) vs. outer-loop iteration (x-axis) for four datasets. \textsc{ExperiGen} (random sampling) improves steadily throughout 30 iterations. Boosting saturates early (10--15 iterations) because residual errors increasingly overlap with existing hypotheses. Clustering performs comparably to random with higher variance. No Data underperforms throughout, confirming that concrete observations are necessary for hypothesis generation.}
    \label{fig:observation}
\end{figure}

\cref{fig:observation} shows accuracy as a function of outer-loop iterations across four datasets. Boosting achieves rapid early gains but saturates after 10--15 iterations, while random sampling continues improving throughout. This saturation arises because natural-language hypotheses describe patterns over data segments; as the hypothesis bank grows, misclassified residuals increasingly overlap with existing hypotheses, leaving diminishing predictive signal for new discoveries. Random sampling avoids this trap by maintaining observation diversity, enabling the Generator to continue proposing novel hypotheses even after easily discoverable patterns are exhausted. The No Data baseline underperforms throughout, confirming that grounding in concrete observations is necessary for generating testable hypotheses. Clustering offers a middle ground, performing comparably to random sampling but with higher variance across datasets.

\subsubsection{Generator Model Scaling}
\label{sec:generator_scaling}

We evaluate how Generator model capability affects hypothesis quality by varying the Generator while holding the Experimenter fixed at Qwen3-32B (with reasoning) and inference at GPT-4o. \cref{tab:model_comparison} reports OOD accuracy on two datasets.

Performance increases monotonically with model scale: o3 achieves 73.1\% on Deceptive Reviews and 70.2\% on News Headlines, while smaller models like Qwen3-30B-A3B reach only 60.7\% and 60.3\% respectively. Qwen3-32B and GPT-4o produce comparable results (71.0\% vs.\ 70.8\% on Deceptive Reviews), suggesting that model scale matters more than architecture at this capability tier. Qualitatively, stronger models propose more specific and falsifiable hypotheses. For example, weaker models generate vague claims like ``emotional language is persuasive,'' while o3 proposes targeted variants such as ``counterarguments that acknowledge the OP's emotional state before presenting logical rebuttals are more persuasive than purely logical responses.'' The latter admits direct operationalization and covariate control, enabling the Experimenter to produce rigorous validation.

\begin{table}[ht!]
\centering
\begin{tabular}{lcc}
\toprule
\textbf{Generator Model} & \textbf{Deceptive Reviews} & \textbf{News Headlines} \\
\midrule
Qwen3-30B-A3B  & 60.7 & 60.3 \\
Qwen3-14B      & 61.3 & 63.5 \\
Qwen3-32B      & 71.0 & 65.8 \\
GPT-4o         & 70.8 & 67.1 \\
o3             & 73.1 & 70.2 \\
\bottomrule
\end{tabular}
\caption{\textbf{Generator model scaling.} OOD accuracy (\%) when varying the Generator model. Experimenter is fixed at Qwen3-32B (with reasoning); inference uses GPT-4o. Performance scales monotonically with capability: o3 achieves the highest accuracy on both datasets. Qwen3-32B and GPT-4o perform comparably, suggesting scale matters more than architecture at this tier.}
\label{tab:model_comparison}
\end{table}

\subsubsection{Experimenter Validation on Known Hypotheses}
\label{sec:experimenter_validation}

To verify that the Experimenter correctly executes statistical tests, we curate 20 hypotheses from peer-reviewed studies on deceptive reviews, news headlines, and stress detection (Dreaddit). A domain expert classified each hypothesis by experimental rigor: \textbf{Easy} hypotheses involve a single feature with a direct statistical test; \textbf{Medium} hypotheses require covariate control or derived features; \textbf{Hard} hypotheses demand multi-step decomposition, interaction effects, or complex feature extraction. \cref{tab:analyst_ablation} reports the percentage of hypotheses correctly validated by each model.

Reasoning capability proves critical. Qwen3-32B without reasoning achieves only 30\% accuracy on easy hypotheses, whereas enabling reasoning raises this to 70\%. Qualitatively, models without reasoning rarely invoke the code interpreter or feature extractor tools, instead producing superficial analyses that fail to engage with the data. The gap widens for harder hypotheses: o3's stronger decomposition abilities yield 70\% accuracy on hard cases, compared to 50\% for Qwen3-32B with reasoning. Interestingly, GPT-4o outperforms Qwen3-32B on easy hypotheses (80\% vs.\ 70\%) but underperforms on medium and hard cases (40\% and 30\%), suggesting that reasoning depth matters more than surface capability for complex experimental designs.

\begin{table}[ht!]
\centering
\begin{tabular}{lccc}
\toprule
\textbf{Experimenter Model} & \textbf{Easy} & \textbf{Medium} & \textbf{Hard} \\
\midrule
Qwen3-32B (w/o reasoning) & 30 & 10 & 10 \\
Qwen3-32B (w/ reasoning)  & 70 & 60 & 50 \\
GPT-4o                    & 80 & 40 & 30 \\
o3                        & 90 & 80 & 70 \\
\bottomrule
\end{tabular}
\caption{\textbf{Experimenter validation accuracy on known hypotheses.} We curate 20 hypotheses from prior work and classify by experimental rigor: Easy (single feature, direct test), Medium (covariate control required), Hard (multi-step decomposition). Values are percentage of hypotheses correctly validated. Reasoning capability is critical: Qwen3-32B improves from 30\% to 70\% on easy hypotheses when reasoning is enabled. o3 achieves the highest accuracy across all difficulty levels.}
\label{tab:analyst_ablation}
\end{table}

\subsubsection{Scaling Inference with AutoML}
\label{sec:automl}

\begin{figure}[!t]
    \centering
    \includegraphics[width=0.6\linewidth]{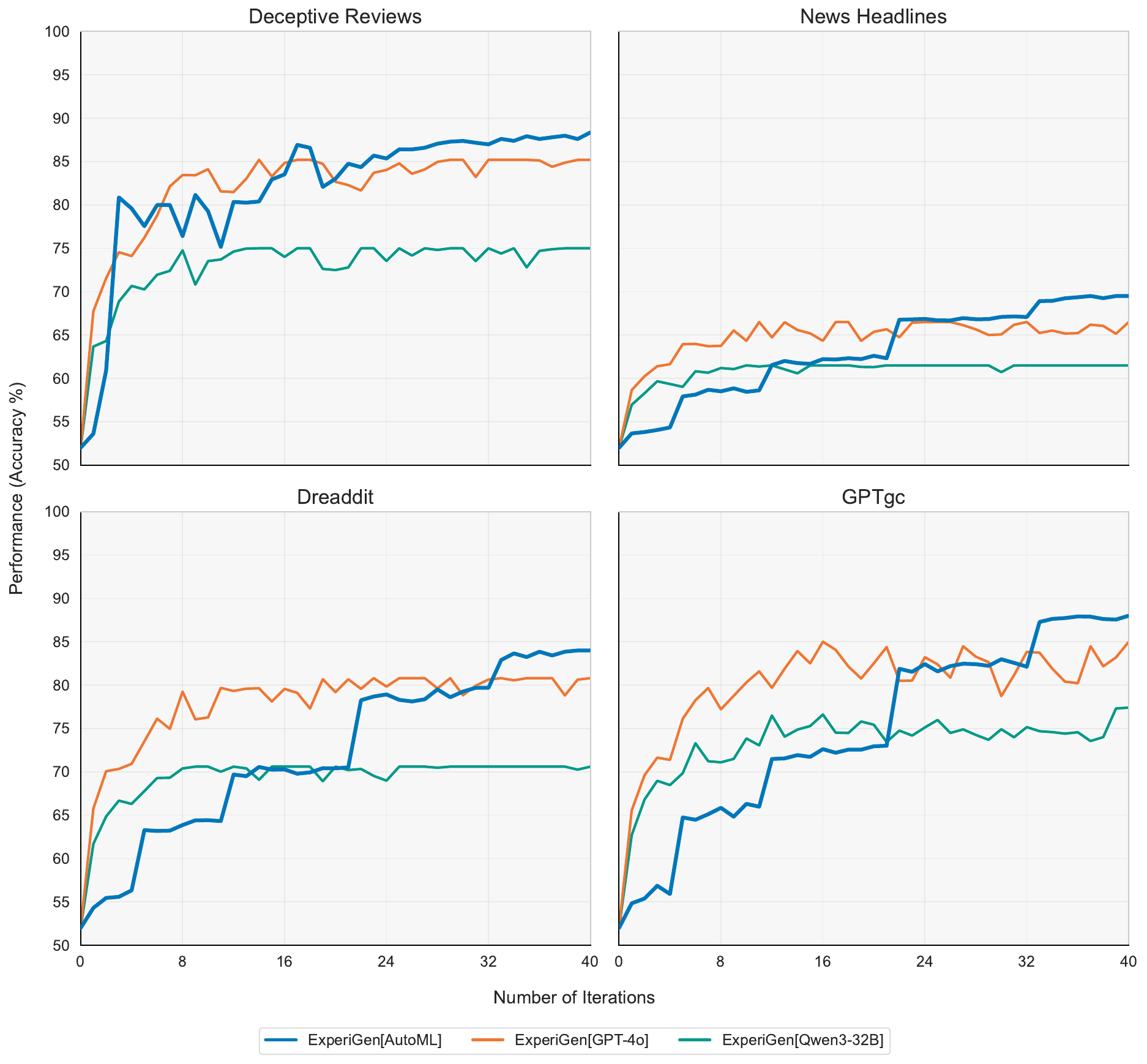}
    \caption{\textbf{Inference scaling with hypothesis bank size.} Accuracy (y-axis) vs. outer-loop iteration (x-axis). LLM-based inference plateaus after $\sim$20 iterations due to difficulty selecting relevant hypotheses from large candidate sets. An AutoML pipeline (gradient-boosted classifier) trained on Experimenter-extracted features continues improving up to 40 iterations. ExperiGen[AutoML] outperforms all other configurations after iteration 15 across all four datasets.}
    \label{fig:automl}
\end{figure}

As the hypothesis bank $\mathcal{H}_{\text{bank}}$ grows, inference must select relevant hypotheses from an increasingly large candidate set. \cref{fig:automl} reveals that LLM-based inference plateaus after approximately 20 iterations, suggesting that selecting top-$k$ candidates from a large bank exceeds the model's effective context utilization.

To address this bottleneck, we evaluate an alternative inference mechanism: an AutoML pipeline trained on features extracted by the Experimenter during hypothesis validation. Rather than prompting an LLM to apply hypotheses directly, we train a gradient-boosted classifier on the accumulated feature columns (e.g., \texttt{is\_emotional}, \texttt{sentiment\_score}, \texttt{contains\_citation}). This approach continues to improve up to 40 iterations, demonstrating better scalability for large hypothesis banks. The AutoML variant consistently outperforms LLM inference across all four datasets after iteration 15, with the gap widening as more hypotheses accumulate. This result suggests a natural handoff: LLM inference for small hypothesis banks where interpretability matters, AutoML for large banks where predictive power is paramount. We note that this pipeline leverages only the \emph{features} discovered during experimentation; synthesizing the hypotheses themselves into a unified predictive model remains an open direction.

\subsubsection{Feature Extractor Reliability}
\label{sec:llm_judge_eval}

The LLM-based feature extractor annotates samples for semantic constructs that cannot be computed programmatically (e.g., ``uses emotional appeal,'' ``contains implicit threat''). Errors in feature extraction propagate to hypothesis validation, potentially inflating false discovery rates. We conduct a controlled evaluation comparing LLM annotations against expert ground truth, and compare against embedding-based methods used in prior work~\citep{movva2025sparse}.

\paragraph{Evaluation Setup.} We sample 500 instances each from Twitter engagement prediction and ChangeMyView (CMV) persuasion classification. For each dataset, we define five binary features spanning annotation difficulty: surface-level cues (exclamation presence, topic classification) to features requiring deeper reasoning (sentiment polarity, politically charged language, metaphor usage). Three domain experts independently annotated all 1,000 samples, achieving inter-annotator agreement of Cohen's $\kappa = 0.69$ (substantial agreement). Gold labels are majority vote.

\paragraph{Results.} \cref{tab:llm_judge_eval} reports F1 scores averaged across all ten features. Prompt-based LLM annotation achieves strong agreement with experts: GPT-5-chat attains F1 = 0.80; GPT-4o reaches 0.79; Qwen3-32B and Qwen3-14B achieve 0.76 and 0.71 respectively. Notably, performance varies by only 0.13 F1 across models spanning vastly different capabilities, enabling substitution with cheaper models for cost efficiency since feature extraction is the most expensive step.

Embedding-based classification performs substantially worse, with OpenAI text-embedding-large achieving only F1 = 0.56. Per-feature analysis reveals that embedding methods match LLMs on surface-level features (topic classification, exclamation detection) but degrade sharply on features requiring semantic reasoning: sentiment polarity, political charge, and metaphor detection. This gap motivates our design choice to use prompt-based annotation for subjective features, reserving programmatic extraction for computable properties. Interestingly, reasoning mode (``think'') reduces performance by 2--4 points, likely because extended reasoning introduces spurious justifications for edge cases that human annotators judge more directly.

\begin{table}[ht!]
\centering
\begin{tabular}{llc}
\toprule
\textbf{Method} & \textbf{Model} & \textbf{F1} \\
\midrule
\multicolumn{3}{l}{\textit{Prompt-based (LLM-as-a-Judge)}} \\
\midrule
& GPT-5-chat (w/o think) & \valbest{0.80} \\
& GPT-4o & \valgood{0.79} \\
& Qwen3-32B (w/o think) & 0.76 \\
& Qwen3-32B (w/ think) & 0.74 \\
& Qwen3-14B (w/o think) & 0.71 \\
& Qwen3-14B (w/ think) & 0.67 \\
\midrule
\multicolumn{3}{l}{\textit{Embedding-based Classification}} \\
\midrule
& OpenAI text-embedding-large & 0.56 \\
& GRIT-LM-7B & 0.51 \\
& Qwen3-8B-Instruct & 0.49 \\
\bottomrule
\end{tabular}
\caption{\textbf{Feature extractor accuracy.} F1 scores on 1,000 samples (500 Twitter, 500 CMV) across ten binary features. Gold labels are majority-vote from three domain experts ($\kappa = 0.69$). Prompt-based LLM annotation substantially outperforms embedding-based classification, with GPT-5-chat achieving F1 = 0.80. Performance varies only 0.13 F1 across LLMs, enabling cost-efficient substitution. Reasoning mode reduces accuracy by 2--4 points.}
\label{tab:llm_judge_eval}
\end{table}

\section{Expert Study}
\label{sec:expert_study}

\subsection{Motivation}
Automated hypothesis generation systems can produce statistically significant findings, but statistical significance alone does not guarantee scientific validity. A hypothesis may exploit dataset-specific artifacts, restate well-known results, or lack the clarity and testability required for actionable research. Prior work evaluates hypothesis quality through predictive accuracy or coverage metrics, but these proxies fail to capture whether generated hypotheses meet the standards that domain experts would apply when deciding which questions merit further investigation.

To address this gap, we conducted an expert study evaluating hypotheses generated by \textsc{ExperiGen} on the Change My View (CMV) persuasion task. The study was designed to assess four key dimensions that distinguish research-worthy hypotheses from spurious or trivial findings: (1)~\emph{novelty} relative to existing literature, (2)~\emph{quality} in terms of clarity, specificity, and testability, (3)~\emph{research expertise} reflected in the experimental design, and (4)~\emph{research-worthiness} as judged by whether experts would recommend pursuing the hypothesis. By recruiting researchers with domain expertise in computational social science and persuasion, we obtain calibrated assessments that complement our quantitative metrics.

\subsection{Participants and Procedure}
\label{sec:expert_participants}

\paragraph{Expert Recruitment.} We recruited five faculty members with established research in computational social science, psycholinguistics, and human behavior. None of the experts are authors on this paper. All hold professorial appointments and have extensive experience advising graduate students, ensuring calibrated, research-grounded evaluations. Unlike crowdsourced studies, our assessments come from senior domain experts with the methodological training to judge whether a hypothesis meets publication-level standards.

\paragraph{Hypothesis Selection.} We randomly sampled 10 hypotheses from the set of novel hypotheses discovered by \textsc{ExperiGen} on the CMV task (the full set is provided in \cref{sec:discovered_hyp}). Random sampling ensures that the evaluated hypotheses are representative of typical system output rather than cherry-picked exemplars.

\paragraph{Procedure.} Each expert independently reviewed 5 hypotheses. The presentation order was randomized for each participant to mitigate order effects. Experts were given an extended period (two weeks) to complete the evaluation at their own pace, allowing for careful consideration of each hypothesis and its experimental design. The study protocol received IRB approval from the respective institutions of the participating researchers.

\subsection{Interface}
We administered the study through a Google Forms questionnaire that presented hypotheses alongside their full experimental designs, including feature construction methods, statistical analyses, and robustness checks. This format ensured that experts evaluated hypotheses in the same context that a reviewer would encounter when assessing a research contribution. Below, we reproduce the key components of the study interface.

\subsubsection{Instructions}
Participants first received detailed instructions explaining the study context, key terminology, and evaluation criteria. The instructions grounded reviewers in the CMV domain and provided guidance on interpreting the statistical results reported for each hypothesis.

\begin{tcolorbox}[colback=gray!5!white, colframe=gray!75!black, title=\textbf{Study Instructions}, fonttitle=\bfseries\small, boxrule=0.5pt, arc=2pt]
\small
This study evaluates the quality of hypotheses and experimental designs automatically generated by an algorithm that answers the question: \textbf{``What Makes Counterarguments Persuasive?''}

A hypothesis is a clear, testable statement that proposes a relationship between variables and can be supported or rejected using data.

The hypotheses in our study explore persuasion dynamics on r/ChangeMyView (CMV), a Reddit community where users post opinions they want challenged. The goal is to identify what linguistic and rhetorical strategies make counterarguments effective at changing minds.

\tcblower
\textbf{How CMV Works:}
\begin{enumerate}[leftmargin=*, itemsep=1pt, topsep=2pt]
    \item \textbf{OP} (Original Poster) submits an argument expressing a view they hold but are open to reconsidering
    \item \textbf{Commenters} respond with counterarguments attempting to change OP's view
    \item \textbf{Delta ($\Delta$)} is awarded by OP when a counterargument successfully shifts their perspective---this is the key outcome we study
\end{enumerate}
\end{tcolorbox}

\begin{tcolorbox}[colback=gray!5!white, colframe=gray!75!black, title=\textbf{Key Terms}, fonttitle=\bfseries\small, boxrule=0.5pt, arc=2pt]
\small
\begin{tabular}{@{}p{2.8cm}p{10cm}@{}}
\textbf{OP} & Original Poster---the person whose view may be changed \\[2pt]
\textbf{Argument} & OP's original post stating their view \\[2pt]
\textbf{Counterargument} & A comment responding to the argument, attempting to change OP's view \\[2pt]
\textbf{Delta ($\Delta$)} & The award OP gives when their view is changed; our measure of success \\[2pt]
\textbf{Success rate} & Percentage of counterarguments that receive deltas (typically 10--15\% overall) \\[2pt]
\textbf{Commenter expertise} & Number of deltas a commenter has earned previously (shown in their ``flair'')
\end{tabular}
\end{tcolorbox}

\begin{tcolorbox}[
  colback=gray!5!white,
  colframe=gray!75!black,
  title=\textbf{Understanding the Statistics},
  fonttitle=\bfseries\small,
  boxrule=0.5pt,
  arc=2pt
]
\small

\textbf{Effect Size Measures:}

\vspace{4pt}
\begin{tabular}{@{}p{3.5cm}p{9.5cm}@{}}
\textbf{Odds ratio (OR)} &
How many times more likely success becomes with the feature.
OR $= 1.5$ means 50\% higher odds; OR $= 2.0$ means double the odds \\[4pt]

\textbf{Effect size ($r$)} &
Strength of relationship.
$r < 0.2$ = small, $r \in [0.2, 0.5]$ = moderate, $r > 0.5$ = large \\[4pt]

\textbf{$p$-value} &
Probability the result is due to chance.
$p < 0.05$ = likely real; $p < 0.001$ = very likely real
\end{tabular}

\vspace{6pt}
\textbf{Rules of Thumb:}
\begin{itemize}\setlength\itemsep{1.8pt}
  \item \textbf{Small effect:} OR $\approx 1.2$--$1.3$, or $\sim$3--5 percentage point difference
  \item \textbf{Moderate effect:} OR $\approx 1.5$--$2.0$, or $\sim$5--10 percentage point difference
  \item \textbf{Large effect:} OR $> 2.0$, or $> 10$ percentage point difference
\end{itemize}

\end{tcolorbox}

\subsubsection{Examples}
Each hypothesis was presented with its full experimental design, including the feature construction methodology (programmatic detection via code or LLM annotation), comparative analyses, interaction effects, and robustness checks. Below, we show an example hypothesis as it appeared to reviewers.

\begin{tcolorbox}[colback=blue!3!white, colframe=blue!50!black, title=\textbf{Hypothesis 2}, fonttitle=\bfseries\small, boxrule=0.5pt, arc=2pt]
\small
\textbf{Counterarguments that include accuracy-motivating elements (e.g., personal experience suggestions, empirical evidence) are more effective in changing views when the original argument contains identity-relevant claims about self-discipline or social norms.}

\vspace{4pt}
\textit{[Link to example Reddit posts provided: \href{https://www.reddit.com/r/changemyview/comments/18dn2cv/cmv_the_practice_of_validating_anothers_feelings/}{Link1} \href{CMV: Those who redefine selfishness to include altruism are not doing anything useful}{Link2}]}
\end{tcolorbox}

\begin{tcolorbox}[colback=white, colframe=gray!60!black, title=\textbf{Experiment}, fonttitle=\bfseries\small, boxrule=0.5pt, arc=2pt]
\small
We first tested whether accuracy-motivating elements in counterarguments affect persuasion success. To identify accuracy-focused appeals, we used GPT-4o to detect empirical evidence markers (phrases like ``studies show,'' ``research suggests,'' ``data indicates'') and personal experience suggestions (phrases like ``try asking yourself,'' ``consider whether,'' ``have you tested''). We computed a binary indicator for counterarguments containing such markers.

\textbf{Result:} Counterarguments with accuracy-motivating elements showed modestly higher success rates (14.2\% vs 11.8\%; OR $= 1.24$, $p = 0.003$).

However, on further inspection, we hypothesized this effect should be stronger when OPs make identity-relevant claims. We measured identity relevance in original arguments using GPT-4o as a classifier trained to detect claims about self-discipline or social norm violations. (The LLM achieved 83\% agreement with human annotations on 400 held-out samples.)

\textbf{Result:} Splitting arguments by identity relevance revealed the predicted interaction:
\begin{itemize}[leftmargin=*, itemsep=1pt, topsep=2pt]
    \item Identity-relevant argument + accuracy appeal: 19.3\% success
    \item Identity-relevant argument + no accuracy appeal: 10.1\% success
    \item Non-identity argument + accuracy appeal: 12.8\% success
    \item Non-identity argument + no accuracy appeal: 11.4\% success
\end{itemize}

The accuracy-motivating approach nearly doubled success rates for identity-relevant posts (OR $= 2.13$, $p < 0.001$), while showing no significant effect for non-identity posts (OR $= 1.14$, $p = 0.31$). The interaction term was significant ($p < 0.001$).

\textbf{Robustness:} Logistic regressions controlling for counterargument length, commenter's prior delta count, OP's prior delta count, response latency, and argument topic confirmed the interaction. The accuracy $\times$ identity interaction remained significant (OR $= 1.87$, $p < 0.001$).
\end{tcolorbox}

Following each hypothesis, reviewers answered five evaluation questions:

\begin{tcolorbox}[colback=yellow!5!white, colframe=yellow!60!black, title=\textbf{Review Questions}, fonttitle=\bfseries\small, boxrule=0.5pt, arc=2pt]
\small
\textbf{1. How novel is this hypothesis relative to existing literature?}\\
\hspace*{1em}$\bigcirc$ Strongly novel \hspace{1em} $\bigcirc$ Moderately novel \hspace{1em} $\bigcirc$ Weakly novel \hspace{1em} $\bigcirc$ Not novel

\vspace{6pt}
\textbf{2. Would you recommend pursuing this hypothesis as a serious research project for your own graduate student?}\\
\hspace*{1em}$\bigcirc$ Strong yes \hspace{1em} $\bigcirc$ Weak yes \hspace{1em} $\bigcirc$ Weak no \hspace{1em} $\bigcirc$ Strong no

\vspace{6pt}
\textbf{3. Overall quality of the hypothesis (clarity, specificity, testability):}\\
\hspace*{1em}Very poor \hspace{0.5em} $\bigcirc$\,1 \hspace{0.5em} $\bigcirc$\,2 \hspace{0.5em} $\bigcirc$\,3 \hspace{0.5em} $\bigcirc$\,4 \hspace{0.5em} $\bigcirc$\,5 \hspace{0.5em} Excellent

\vspace{6pt}
\textbf{4. What level of research expertise does this experimental design reflect?}\\
\hspace*{1em}$\bigcirc$ Advanced undergraduate\\
\hspace*{1em}$\bigcirc$ Early graduate student (1st--2nd year PhD / Master's)\\
\hspace*{1em}$\bigcirc$ Senior graduate student (late-stage PhD)\\
\hspace*{1em}$\bigcirc$ Postdoctoral researcher\\
\hspace*{1em}$\bigcirc$ Independent scientist / faculty

\vspace{6pt}
\textbf{5. How confident are you?}\\
\hspace*{1em}Not confident \hspace{0.5em} $\bigcirc$\,1 \hspace{0.5em} $\bigcirc$\,2 \hspace{0.5em} $\bigcirc$\,3 \hspace{0.5em} $\bigcirc$\,4 \hspace{0.5em} $\bigcirc$\,5 \hspace{0.5em} Very confident

\vspace{6pt}
\textbf{6. Optional comments on this hypothesis:}\\
\hspace*{1em}\textit{[Free-text field for concerns, suggestions, or observations]}
\end{tcolorbox}

\subsection{Results}
\label{sec:expert_study_results}

\paragraph{Novelty.} Experts rated 88\% of hypotheses as ``Strongly novel'' or ``Moderately novel,'' with only 4\% rated ``Not novel.'' This distribution suggests that the iterative refinement process surfaces findings that go beyond obvious or well-established results in the persuasion literature.

\paragraph{Quality.} The mean quality rating was 4.24/5 (SD $= 0.78$), indicating that reviewers found the hypotheses clear, specific, and testable. Experts noted that the experimental designs include appropriate controls and robustness checks.

\paragraph{Research Expertise.} When asked what level of research expertise the experimental designs reflect, the modal response was ``Senior graduate student.'' Annotations spanned all five expertise levels, with 68\% at ``Senior graduate student'' or above. This distribution suggests that the generated experiments embody research practices expected at advanced training levels.

\paragraph{Research-Worthiness.} Crucially, 76\% of annotations recommended pursuing the hypothesis as a serious research project (``Strong yes'' or ``Weak yes''). Reviewers who declined cited concerns about intuitive effects (e.g., ``This hypothesis seems to be quite intuitive and trivial'') rather than methodological flaws, suggesting that even rejected hypotheses were technically sound.

\paragraph{Reviewer Confidence.} The mean confidence rating was 4.32/5 (SD $= 0.69$), indicating that experts felt well-equipped to evaluate the hypotheses presented. This high confidence supports the validity of the assessments.

\paragraph{Qualitative Feedback.} Reviewers provided constructive suggestions, such as expanding keyword-based feature detection to consider full comment context and exploring interaction effects with additional covariates. These comments indicate engaged evaluation rather than superficial review.

\paragraph{Summary Statistics.}
\begin{itemize}[leftmargin=*, itemsep=2pt]
    \item \textbf{Number of expert reviewers:} 5 (faculty with graduate advising experience)
    \item \textbf{Hypotheses evaluated:} 25
    \item \textbf{Mean quality score:} 4.24/5 (SD $= 0.78$)
    \item \textbf{Hypotheses rated novel (Strongly/Moderately):} 88\%
    \item \textbf{Hypotheses recommended for research (Strong/Weak yes):} 76\%
    \item \textbf{Mean reviewer confidence:} 4.32/5 (SD $= 0.69$)
\end{itemize}
\include{pages/appendix/studies}

\include{pages/appendix/examples}

\include{pages/appendix/prompts}

\section{Algorithm}
\subsection{Scalability}

\begin{figure}
    \centering
    \includegraphics[width=1\linewidth]{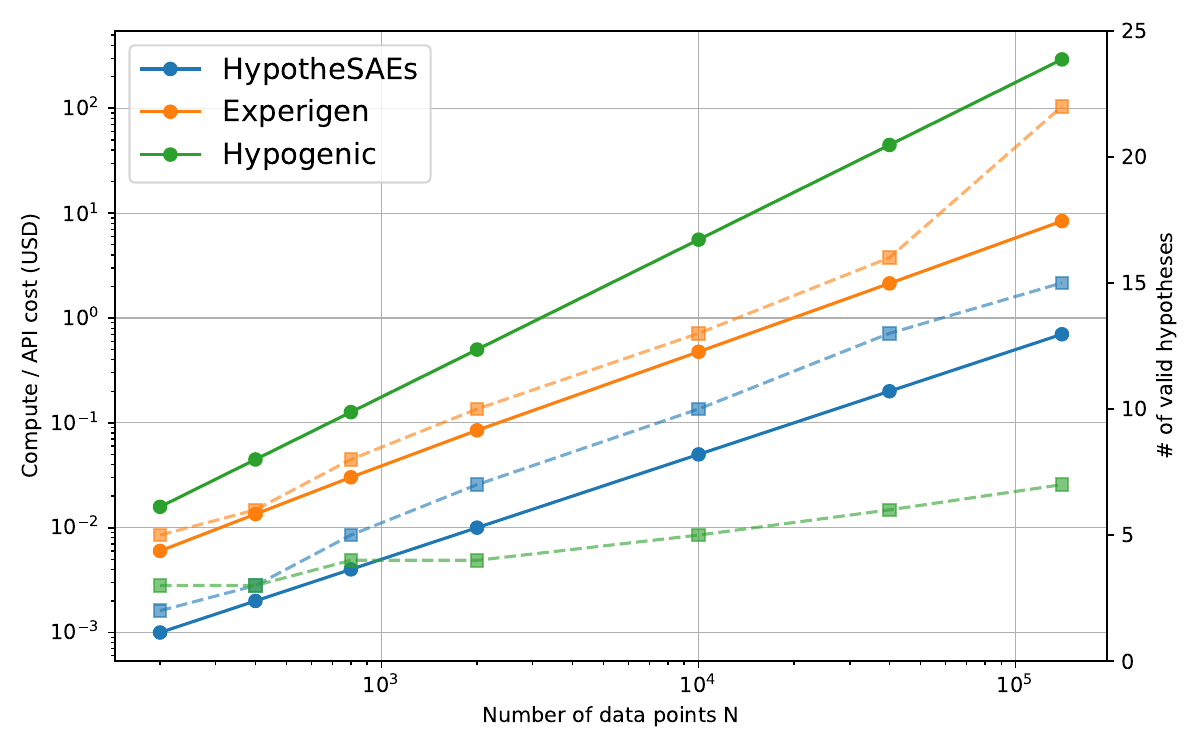}
    \caption{
    \textbf{Scalability of hypothesis discovery with increasing dataset size.}
    We plot API / compute cost (solid lines, left y-axis; log scale) and the number of statistically significant hypotheses (dashed lines, right y-axis) as a function of dataset size $N$.
    Significant hypotheses are identified using LASSO regression with Bonferroni-corrected $p$-values.
    ExperiGen consistently discovers the largest number of valid hypotheses across all dataset sizes, while exhibiting asymptotic compute growth comparable to HypotheSAEs and substantially lower than Hypogenic.
    Notably, while computational cost increases with $N$, the number of valid hypotheses grows slowly, reflecting the sublinear (often logarithmic) growth of statistically supported explanations in natural datasets.
    By enforcing novelty during hypothesis generation, ExperiGen avoids redundant rediscovery of common explanations and aligns computation with the growth of the plausible hypothesis space rather than the size of the dataset, resulting in superior hypothesis coverage without sacrificing scalability.
    }
    \label{fig:cost_scaling}
\end{figure}

\cref{fig:cost_scaling}
demonstrates that ExperiGen achieves a favorable trade-off between computational efficiency and hypothesis discovery.
Although all methods exhibit increasing cost with dataset size, the number of statistically valid hypotheses grows only sublinearly, consistent with classical results for finite hypothesis spaces.
Methods that repeatedly re-evaluate or rediscover common explanations incur increasing cost without proportional gains in hypothesis diversity.
In contrast, ExperiGen enforces novelty in hypothesis generation, ensuring that each iteration evaluates a unique, plausible hypothesis and preventing redundant exploration as data grows.
As a result, ExperiGen scales comparably to the most efficient alternative while consistently discovering a larger and more diverse set of valid hypotheses.

\subsection{Enabling the Data Analyst}

\label{sec:appendix:enabling_analyst}
A typical graduate-level research cycle in data-driven sciences proceeds through several stages: examining the dataset to understand available variables and distributions, engineering features through code or external libraries, running statistical tests, interpreting results, and iterating. Consider a researcher studying tweet engagement who loads the dataset, computes text length as a baseline feature, observes a weak correlation, then hypothesizes that emotional content matters. To test this, they need a sentiment score but no pre-existing library captures the nuanced emotion they have in mind. The pipeline breaks: the researcher must pause, design an annotation scheme, crowdsource labels or manually annotate samples, wait for results, merge them back, and only then resume analysis. This interruption occurs whenever the hypothesis involves features without accurate off-the-shelf models, which describes most interesting semantic properties.

Modern LLMs, trained on millions of notebook sessions, Stack Overflow threads, and data science tutorials, have internalized both the analytical workflow and substantial world knowledge for feature annotation. We design the Experimenter's environment to exploit these priors, enabling rigorous multi-step analyses without pipeline interruptions. Three capabilities realize this design:

\paragraph{Grounded Data Understanding.}
The Experimenter must understand dataset structure, feature distributions, and concrete examples before designing analyses. We construct a compact verbalization $\mathcal{\hat{D}}$ that combines structural metadata, statistical summaries, and representative observations in pandas output formats aligned with LLM pretraining (\cref{sec:appendix:dataset_description}). This format achieves 3.5$\times$ lower perplexity compared to CSV or JSON serialization. The verbalization updates dynamically after each feature extraction, so the Generator observes an expanding feature space across iterations.

\paragraph{Iterative Refinement through Persistent Execution.}
The Experimenter executes analyses through a sandboxed code interpreter backed by a persistent IPython kernel (\cref{sec:experimentation_sandbox}). Persistence enables iterative refinement: variables, loaded data, and helper functions defined in early iterations remain available in later ones. On the Deceptive Reviews task, stateful execution reduces errors by 13$\times$ (91 vs. 1,213) and context length by 3.9$\times$ compared to stateless baselines.

\paragraph{Scope of Extractable Features.}
The code interpreter and feature extractor together span syntactic features (code-extractable: text length, regex patterns, readability scores), semantic features (extractor-required: sentiment, topic, emotion, persuasion strategies), and relational features (code over dataset: position within thread, timing relative to other responses). Prompt-based annotation achieves F1 = 0.80 against expert ground truth (\cref{sec:llm_judge_eval}), substantially outperforming embedding-based approaches. The balance between tools depends on dataset characteristics: code-based features dominate when syntactic markers are predictive, while the extractor sees higher usage on smaller datasets where code patterns return null more frequently.

\subsection{Architectural Flexibility and Modality Generalization.}

Hypothesis discovery in natural datasets often requires features that emerge from relationships between samples rather than properties of individual instances. Consider the ``first-mover advantage'' in persuasion (~\ref{sec:case_cmv}): testing whether early comments are more persuasive requires computing each comment's temporal rank within its thread, a relational feature that no per-sample annotator can produce. Prior methods cannot express such features. HypoGenic prompts LLMs with multiple examples to induce hypotheses, but the resulting features are limited to what the LLM can observe in each instance independently. HypotheSAEs encodes samples through a fixed embedding model, precluding any cross-sample computation. Both approaches implicitly assume i.i.d.\ data where predictive signal resides in marginal properties of $x_i$, not in relationships among $\{x_j\}$.

\textsc{ExperiGen} lifts this restriction by decoupling hypothesis generation from feature instantiation. The Generator proposes candidate hypotheses from a data schema and discovery history; the Experimenter realizes features through code execution or single-item LLM annotation. Code-based featurization admits arbitrary transformations over the dataset: \texttt{groupby} aggregations, temporal ranks, lag features, and conditional statistics. In the CMV task, this enabled a nine-iteration refinement trajectory that progressively isolated timing effects from confounds---computing within-post position (Iter~4), controlling for post-level fixed effects (Iter~3), and stratifying by quality quartiles (Iter~9)---none of which could be expressed as per-sample prompts.

The decoupling also resolves a modality bottleneck. HypotheSAEs inherits the context limits of its embedding backbone (8k tokens for openai text-embedding-large), excluding long documents, images, and video. HypoGenic requires multi-example reasoning to generate hypotheses, a task at which even frontier VLMs degrade sharply compared to single-item annotation. By isolating hypothesis generation (schema-level, requires creativity) from feature verification (instance-level, requires accuracy), \textsc{ExperiGen} aligns each subtask with current model strengths. Extending to video or audio requires only swapping the feature extractor module, not retraining embeddings or invoking multi-item VLM calls that scale poorly with context (Figure~\ref{fig:cost_scaling}). Empirically, this flexibility yields consistent gains on image-intensive benchmarks where embedding-based baselines suffer truncation and multi-item degradation (Table~\ref{tab:main_benchmarks}).

\subsection{Architectural Flexibility and Modality Generalization}

\label{sec:appendix:architectural_flexibility}

Prior methods cannot express features that emerge from relationships between samples. HypoGenic limits features to what the LLM observes in each instance independently; HypotheSAEs precludes cross-sample computation through its fixed embedding model. Both implicitly assume i.i.d.\ data where predictive signal resides in marginal properties of $x_i$, not in relationships among $\{x_j\}$.

\textsc{ExperiGen} lifts this restriction by decoupling hypothesis generation from feature instantiation. The Generator proposes hypotheses from a data schema; the Experimenter realizes features through code execution or LLM annotation. Code-based featurization admits arbitrary transformations: \texttt{groupby} aggregations, temporal ranks, lag features, and conditional statistics. This enables relational hypotheses such as ``first-mover advantage'' in persuasion, which requires computing each comment's temporal rank within its thread (~\cref{fig:cmv_refinement}; see \cref{sec:case_studies} for the full refinement trajectory).

The decoupling also resolves a modality bottleneck. HypotheSAEs inherits context limits of its embedding backbone (8k tokens), excluding long documents, images, and video. HypoGenic requires multi-example reasoning, at which even frontier VLMs degrade sharply. By isolating hypothesis generation (schema-level, requires creativity) from feature verification (instance-level, requires accuracy), \textsc{ExperiGen} aligns each subtask with current model strengths. Extending to video or audio requires only swapping the feature extractor module (\cref{sec:llm_judge_eval}), not retraining embeddings. Empirically, this yields consistent gains on image-intensive benchmarks (Table~\ref{tab:main_benchmarks}; \cref{fig:cost_scaling}).

\section{\texorpdfstring{Dynamic Dataset Description $\mathcal{\hat{D}}$}{Dynamic Dataset Description D-hat}}
\label{sec:appendix:dataset_description}
For the agents to generate testable hypotheses or perform the appropriate analyses, they must first be grounded in the details of the dataset. Naively serializing a dataset as CSV or JSON consumes tens of thousands of tokens even for modest tables, leaving insufficient context for reasoning. We instead embed essential contextual information within a compact prompt that combines three complementary views. An example of the resulting verbalization appears in \cref{lst:dataset-description}.

\paragraph{Format Selection and Pretraining Alignment}
A key design decision is our use of pandas output formats rather than raw serialization. LLMs are trained on vast corpora of Python notebooks and data science conversations where pandas outputs (\texttt{.info()}, \texttt{.describe()}, markdown tables) are among the most frequent token sequences. We verify this alignment empirically: Qwen3 achieves 3.5$\times$ lower perplexity when generating analysis code from pandas formatted prompts compared to equivalent CSV or JSON serialization. This pretraining alignment improves both code correctness and the agent's ability to reason about distributions.

\paragraph{Structural Metadata}
To ground the agents in the dataset structure, we include structural metadata using the standard \texttt{pandas.DataFrame.info()} output format. This provides the number of rows and columns, and for each column: the column name, count of non-null entries, and data type.

\paragraph{Statistical Summary}
We include a statistical characterization to provide agents with distributional information. Numerical features are summarized using \texttt{pandas.DataFrame.describe()}, providing mean, variance, and quantiles. Categorical features are represented through frequency distributions showing the top categories with their proportions. This compact representation allows the agent to identify skewed distributions and sparse categories without requiring exploratory plotting, which would incur substantial token overhead.

\paragraph{Representative Observations}
We present a row-major view of randomly sampled observations in markdown table format, allowing the agent to observe typical value ranges, column interactions, and concrete data patterns. Unlike prior work on table understanding that adopts either row-major or column-major formats to optimize query tractability, we include both views: the column-major statistics above and row-major samples here. Our datasets are relatively narrow, and the agent benefits from seeing complete observations alongside distributional summaries. We qualitatively observe that the Generator's reasoning frequently references specific sample rows when justifying proposed hypotheses, confirming that concrete examples ground abstract claims. This observation follows existing table understanding literature as well \citep{herzig2020tapas}

\lstset{
  basicstyle=\ttfamily\small,
  breaklines=true,
  frame=single,
  backgroundcolor=\color{gray!5},
  tabsize=2,
  showstringspaces=false
}

\begin{lstlisting}[caption={Dataset Description}, label={lst:dataset-description}]
<class 'pandas.core.frame.DataFrame'>
RangeIndex: 500 entries, 0 to 499
Data columns (total 5 columns):
 #   Column              Dtype
---  ------              -----
 0   review              object
 1   label               object
 2   pos_adj             object
 3   length               int64
 4   sentiment           object
dtypes: int64(1), object(4)

Numerical Columns Statistics:

           length
count      500.00
mean       249.50
std         44.48
min         30.00
25
50
75
max        576.00

Categorical Columns Statistics:

label
truthful     51
deceptive    49

pos_adj
yes    58
no     41

sentiment
positive    50
negative    47
neutral      2

Random Sample Rows:

| review                                  |label    |pos_adj|sentiment|
|-----------------------------------------|---------|-------|---------|
|I was looking for a unique, hip experie..|deceptive|yes    |positive |
|This is a stunning hotel in an excell..  |truthful |yes    |positive |
|We stayed at the Fairmont two Saturda..  |truthful |no     |negative |
|I had requested a quiet room several we..|truthful |no     |negative |
|My experience at the Hyatt Regency Chic..|deceptive|yes    |positive |
\end{lstlisting}

This dataset verbalization enables agents to reason about and test hypotheses with high fidelity and grounding, ensuring both comprehensive coverage and compactness. Structural metadata guides variable selection and code generation, while statistical summaries inform choices such as which tests to apply or when to avoid sparse categories. Representative samples further anchor these abstractions in concrete examples. Together, these cues let the generator agent propose semantically coherent, data-plausible hypotheses, and the analyst agent design and execute statistically appropriate analyses. Further we add some optimizations listed below

\begin{itemize}
    \item Columns with unique values only (e.g. rowIDs, hashes) are discarded, as they add no context to either agents and exist probably as primary keys only.
    \item Textual observations are truncated to 100 chars only.
    \item For images, we only keep the relative path of the image, as we have text only agents at the moment
\end{itemize}

\section{Experimentation Sandbox}
\label{sec:experimentation_sandbox}
In this section we discuss the Experimentation Sandbox used by the Experimenter agent, including the Persistent Code Interpreter and LLM-as-a-Judge Feature Extractor. Together, these tools enable the broad scope of analyses required for hypothesis testing like statistical tests, observing correlations, syntactic features via code transformations, semantic features via LLM annotation, and data transformations/normalizations.

\subsection{Code Interpreter}
The code interpreter exposes a controlled Python execution sandbox via a persistent Jupyter kernel. We adapt the Qwen-Agent\footnote{\url{https://github.com/QwenLM/Qwen-Agent}} Code Interpreter, extending it with common data science libraries (statsmodels, scipy, scikit-learn, pandas). On first use per process and agent instance, it creates a unique kernel identified by a UUID+PID key, writes a transient connection file, launches the kernel as a subprocess, and waits for readiness before accepting code. Additionally we allow it to install pip libraries as required.

\paragraph{Why Persistence Matters.}
Kernel state persistence is essential for iterative hypothesis refinement. In iteration 1, the agent might load the dataset and computes a baseline feature. Then, it defines a helper function to extract n-grams and view top trends. Then, it reuses both the loaded data and the helper function to test a refined hypothesis involving specific phrase patterns. Without persistence, it would need to reload the dataset, redefine the function, and re-explain prior context, consuming tokens and risking errors from inconsistent redefinitions.

We quantify this effect on the Deceptive Reviews task across 10 runs (150 total iterations). Stateless execution produced 1,213 errors, primarily from undefined variable references and conflicting feature definitions. Stateful execution reduced errors to 91, a 13$\times$ improvement. Context length per iteration averaged 3.9$\times$ higher in the stateless setting due to redundant imports, data loading, and re-explanations of prior computations. Persistence also enables relational features requiring cross-sample computation (e.g., a comment's position within its thread), which would require prohibitively verbose single-turn code blocks otherwise. Moreover, including the reloading, more tokens, the average time taken is 2.7$\times$ higher as well.

\paragraph{Timeouts and Limits.} A per-call timeout (default 30s; configurable) is injected by prefixing code with a timer start and cancelling on completion. Storage is capped at 100MB of temporary allocation to prevent large file downloads and memory leakage.

\paragraph{Dependencies and Safety.} The environment bundles NumPy, Pandas, NLTK, spaCy, SymPy, and the Jupyter client. The agent can install additional libraries safely (e.g., readability, vader-sentiment) since the environment is controlled by memory and time limits.

\paragraph{Working Directory.} All execution occurs within a configurable working directory, ensuring no test leakage. Subprocesses and kernels are tracked and automatically terminated via \texttt{atexit} and signal handlers.

\paragraph{Output Semantics.} The tool consumes IOPub messages and normalizes them into Markdown: execution results and display data are captured as text, \texttt{stdout}/\texttt{stderr} streams are preserved, and Python tracebacks are escaped from ANSI sequences for readability.

\subsection{Feature Extractor}
Many hypotheses involve semantic properties that no library or regex can capture. Testing whether ``persuasive arguments use emotional appeals'' requires annotating persuasion strategies a judgment beyond pattern matching. We provide an LLM-as-Judge feature extractor that annotates samples from natural language descriptions, integrating annotation directly into the analytical loop rather than as an external pipeline stage. We verify the consistency of LLM-annotated features against expert ground truth in \cref{tab:llm_judge_eval}, where prompt-based annotation achieves F1 = 0.80 (GPT-5-chat) against expert labels ($\kappa = 0.69$), substantially outperforming embedding-based classification (F1 = 0.49--0.56).

\paragraph{Feature Taxonomy.} The code interpreter and feature extractor together span a broad feature taxonomy:
\begin{itemize}[leftmargin=*, itemsep=2pt]
    \item \textbf{Syntactic features} (code-extractable): text length, punctuation counts, emoji presence, date extraction via regex, readability scores via syllable libraries.
    \item \textbf{Semantic features} (extractor-required): sentiment polarity, topic classification, emotional tone, persuasion strategies, sarcasm detection.
    \item \textbf{Relational features} (code over dataset): position within thread, timing relative to other responses, comparison to group statistics.
\end{itemize}
The balance between tools depends on dataset characteristics. On linguistically-focused tasks like Deceptive Reviews, where syntactic markers (dates, specific phrases, grammatical patterns) predict outcomes, code-based features dominate. On smaller datasets, the feature extractor sees higher usage because code-based patterns more frequently return null when conditions are unmet, reducing support size for statistical tests. This adaptive tool selection emerges from the agent's optimization for experimental evidence.

\paragraph{Prompting and I/O} The feature extractor iterates over the designated column(s) and builds single-turn messages from a template (\cref{lst:feature_extractor_template}) that binds the raw content, the feature (column) name, the closed set of \textit{types}, and a natural-language \textit{description} that operationalizes the feature. An example invocation is shown in \cref{lst:feature_extractor_payload}.

\begin{figure}
\begin{tcolorbox}[colback=blue!3!white,colframe=blue!40!black,title=\textbf{Example}]
\textbf{Feature Name:} Specificity

\textbf{Feature Description:} Does the review  contain specific, concrete, and verifiable details about the hotel stay (e.g., breakfast items, staff names, or room types)?.

\textbf{Allowed Feature Types:} \{Specific, Vague,No Details\}
\label{lst:feature_extractor_payload}
\end{tcolorbox}
\noindent
\captionof{figure}{An example of the prompt filled for the \emph{Specificity} feature (from the Deceptive Review dataset) is shown below.}
\end{figure}

\subsubsection{Feature Extractor Prompt}
The following instructions are used to guide the model in extracting linguistic or semantic features from short texts (e.g., hotel reviews) and visual features from images. It standardizes the feature extraction process across tasks and datasets.

\begin{tcolorbox}[colback=gray!5!white,colframe=gray!75!black,title=\textbf{Textual Feature Prompt Template}]
You are a helpful assistant designed to analyze and featurize short texts according to specific feature descriptions.

Below is a short text snippet. Based on the feature description and predefined types, assign the most appropriate label to the text. Your response should strictly match one of the provided labels and must be directly inferable from the content of the text.

---

\textbf{Text:} \{text\}

\textbf{Feature Name:} \{feature\}

\textbf{Feature Description:} \{description\}

\textbf{Allowed Feature Types (Choose One):} \{types\}
\label{lst:feature_extractor_template}
\end{tcolorbox}

\begin{tcolorbox}[colback=gray!5!white,colframe=gray!75!black,title=\textbf{Visual Feature Prompt Template}]
You are a vision-language model tasked with analyzing visual content and extracting semantic features based on a given description.

You will be shown an image along with a feature description and a closed list of allowed labels. Carefully examine the image and assign exactly one label from the list that best fits the feature description. Respond only with the label, with no additional explanation.

---

\textbf{Image:} \{Image\}

\textbf{Feature Name:} \{feature\}

\textbf{Feature Description:} \{description\}

\textbf{Allowed Feature Types (Choose One):} \{types\}

\textbf{Response Format:} \textless one label from the list above\textgreater
\end{tcolorbox}

\section{Datasets}
We evaluate across the original \textbf{HypoBench} tasks \citep{liu2025hypobenchsystematicprincipledbenchmarking} and extend them with new datasets introduced by \textbf{HypotheSAEs} \citep{movva2025sparse} and \textbf{Ours (ExperiGen)} \cref{tab:dataset_table}
\label{sec:Datasets}
\begin{table}[ht!]

\centering

\label{tab:hypobench_extended_tasks}

\resizebox{\linewidth}{!}{

\begin{tabular}{p{3.2cm} p{6.8cm} ccc}

\toprule

\textbf{Task} & \textbf{Description} & \textbf{Train} & \multicolumn{2}{c}{\textbf{Test}} \\

\cmidrule(lr){4-5}

 & & & \textbf{IND} & \textbf{OOD} \\

\midrule[1.5pt]

\multicolumn{5}{l}{\textbf{HypoBench}} \\

\midrule

Deception Detection & Distinguish genuine vs.\ deceptive hotel reviews via linguistic cues (e.g., exaggeration, vagueness, unnatural phrasing). & 1,600 & 800 & 640 \\

AI-Gen. Detection & Decide if a story (given a prompt) is human- or AI-written; tests stylistic and structural pattern recognition. & 800 & 800 & 800 \\

Persuasive Argument & Given two arguments, predict the more persuasive based on rhetoric, logic, and emotional appeal. & 750 & 750 & 500 \\

Mental Stress & Detect mental stress in Reddit posts via affective and psychological cues. & 1,000 & 1,000 & 500 \\

News Engagement & Choose which of two headlines garners more clicks, testing linguistic framing and interest cues. & 700 & 700 & 453 \\

\midrule[1.5pt]

\multicolumn{5}{l}{\textbf{HypotheSAEs}} \\

\midrule

Congress & Predict U.S. congressional party affiliation (Republican/Democrat) from speeches (2005--2007). Tests large-scale scalability. & 40k & 5k & -- \\

\midrule[1.5pt]

\multicolumn{5}{l}{\textbf{ExperiGen}} \\

\midrule

Tweets \citep{singh2024measuring} & Predict which tweet in a pair is more persuasive, leveraging metadata (usernames, tags, etc.). & 5k & 1k & 2k \\

Design \citep{patnaik2025aesthetiq} & Identify which of two graphic layouts was AI-generated; tests multimodal reasoning over structure and visuals. & 1.6k & 200 & -- \\

Memorability \citep{khosla2015understanding,singh2024teachinghumanbehaviorimproves} & Predict which image in a visually similar pair is more memorable (LaMem dataset). Tests fine-grained visual cues. & 8k & 1k & -- \\

CMV \citep{tan2016winning} & Predict whether a counterargument successfully changes the original poster's view on Reddit's r/ChangeMyView. & 4k & 500 & -- \\

\bottomrule

\end{tabular}}

\caption{Summary of Datasets used for our evaluation}
\label{tab:dataset_table}
\end{table}

\section{Baseline Implementation Details \& Hyperparameters}
\label{sec:hyp_baseline}

\subsection{HypoGeniC}
\label{sec:hypogenic-details}
\begin{itemize}[leftmargin=*]
    \item \textbf{Models:} We use Qwen3-32B, GPT-4o for hypothesis generation and GPT-4o for inference in our experiments. For Image experiments only GPT-4o was used.
    \item \textbf{Hyperparameters:}
    \begin{itemize}[leftmargin=*, itemsep=2pt, parsep=0pt, topsep=0pt, partopsep=0pt]
        \item Top-$k$ selector: 10 hypotheses evaluated per example
        \item Learning rate scale ($\alpha$): 0.5
        \item Temperature: $1\times 10^{-5}$
        \item Seed: 42
    \end{itemize}
    \item \textbf{Inference:} Multiple strategies including best-accuracy hypothesis, filter and weighted vote, one-step adaptive, and two-step adaptive. (We use the best to keep a fair ablation)
\end{itemize}

\subsection{HypotheSAEs}
\label{sec:hypothesaes-details}

\begin{itemize}[leftmargin=*]
    \item \textbf{Embeddings:} OpenAI text-embedding-large  for text based tasks, CLIP-VIT-336 for Design Aesthetics, and finetuned CLIP-VIT-336 for Memorability.
    \item \textbf{Hyperparameters:}
    \begin{itemize}[leftmargin=*, itemsep=2pt, parsep=0pt, topsep=0pt, partopsep=0pt]
        \item \textbf{CONGRESS:} $(M, k) = (4096, 32)$
        \item \textbf{NEWS HEADLINES:} $(M, k) = (256, 8)$ and $(32, 4)$ (activations combined)
        \item \textbf{Visual:} $(M, k) = (1024, 16)$
        \item \textbf{Remaining:} $(M, k)$ Best of (per environment)$(4096, 32)$, $(256, 8)$ and $(32, 4)$ (activations combined)
    \end{itemize}
    \item \textbf{Neuron Interpretation:}
    \begin{itemize}[leftmargin=*, itemsep=2pt, parsep=0pt, topsep=0pt, partopsep=0pt]

        \item Top-activating examples: 10 per neuron
        \item Low-activating examples: 10 per neuron
        \item Candidate interpretations: 3 per neuron (select highest fidelity)
        \item Fidelity evaluation: F1 score on 100 positive + 100 negative samples using GPT-4o-mini
    \end{itemize}
\end{itemize}

%% file: pages/appendix/studies.tex
\section{Case Studies}
\label{sec:Case Studies}
\label{sec:case_studies}
\subsection{Discovery: Response Timing Predicts Persuasion Success}

\label{sec:case_cmv}

Reddit's r/ChangeMyView (CMV) provides a natural laboratory for studying persuasion. Users post opinions they are open to reconsidering, and commenters attempt to change their view. When a comment succeeds, the original poster (OP) awards a ``delta.'' The dataset contains 700k comments across 100k posts, with timestamps for both posts and comments.

Existing methods converge on content-based hypotheses: empathy markers, evidence citations, narrative structure. These patterns are salient in small samples (HypoGeniC) and dominate text embeddings (HypotheSAEs). After 30 iterations, both methods plateau on content features with diminishing returns. The timestamp fields (\texttt{post\_created\_utc}, \texttt{comment\_created\_utc}) remain unexplored because they require computing derived features (time differences, ranks) rather than extracting textual patterns.

ExperiGen's Generator, after going through a few content hypotheses, proposes: \emph{``Counterarguments posted earlier are more likely to successfully change the original poster's view.''} This hypothesis cannot be tested without a code interpreter: it requires computing \texttt{response\_time = comment\_created\_utc - post\_created\_utc} for each comment, a derived feature absent from the raw data. The subsequent refinement demands increasingly complex computations---binning timestamps into windows, computing within-post position ranks, fitting logistic regressions with multiple covariates, and bootstrapping confidence intervals. Text-only methods like HypoGeniC and HypotheSAEs cannot execute these operations; they can only extract patterns from existing text fields. ExperiGen's Experimenter, equipped with a code interpreter, computes each derived feature and runs the appropriate statistical test, enabling the full 6-iteration refinement (Listing~\ref{fig:cmv_refinement}).

The refinement transforms the initial correlation ($r{=}0.44$, $p{<}10^{-16}$) into a robust finding by systematically testing alternative explanations: (1) establishing a monotonic decay pattern rather than an optimal window, (2) isolating the first-comment advantage, (3) separating timing and position as independent effects, (4) controlling for comment quality features, and (5) ruling out OP disengagement as a confounder.

\paragraph{Validated Hypotheses.} The refinement yields two hypotheses added to the bank:

\begin{enumerate}[nosep,leftmargin=*]
\item \emph{Response time between 30 minutes and 2 hours maximizes persuasion success.} Comments in this window succeed at 2.3$\times$ the base rate, controlling for topic, position, and quality. Very early responses ($<$30 min) show slight disadvantage, suggesting rushed comments underperform.

\item \emph{First-comment position provides a 2$\times$ advantage independent of timing.} Even controlling for response time, the first comment succeeds at 19.9\% versus 9.5\% for later comments. This reflects visibility effects: OPs read early comments more carefully.
\end{enumerate}

\begin{figure}[t]
\footnotesize
\begin{mdframed}[linewidth=0.5pt, innerleftmargin=10pt, innerrightmargin=10pt, innertopmargin=8pt, innerbottommargin=8pt]
This is the refinement trajectory for hypothesis no. 10, previous hypotheses were focused on content-based features like length, statistics, and evidence markers.

\textbf{Iteration 1:} Counterarguments posted earlier are more likely to successfully change the original poster's view

\textit{Rationale:} Temporal dynamics have been unexplored in prior content-based hypotheses. When a counterargument is posted may affect the poster's receptiveness independent of what is said.

\textit{Experiment:} Calculate time difference between counterargument and argument creation as response time; filter outliers ($>$30 days); compare response time for successful vs unsuccessful counterarguments using Mann-Whitney U test.

\textit{Evidence:}  Moderate-to-large effect size ($r{=}0.44$); highly significant relationship ($p{<}10^{-16}$) between response time and persuasion success.

\vspace{3pt}

\textbf{Iteration 2:} The relationship between response time and persuasion success follows monotonic decay rather than having an optimal window

\textit{Rationale:} Earlier counterarguments are correlated with opinion change, but the functional form is unclear. There may be a ``sweet spot'' where waiting improves effectiveness, or earliness may simply be better throughout.

\textit{Experiment:} Bin response time for counterarguments into 9 time windows (0-30min, \ldots, 7-30d); calculate success rate (proportion successfully changing poster's view) per bin; $\chi^2$ test to determine whether the distribution of outcomes is independent of time bin membership.

\textit{Evidence:} Shows clear monotonic decay; no optimal window detected; highly significant trend (Success rate: 18.4\% at $<$1h $\to$ 1.3\% at 7-30d; $\chi^2{=}17{,}525$, $p{<}10^{-16}$).

\vspace{3pt}

\textbf{Iteration 3:} Being the first counterargument to an argument provides substantial advantage beyond just being early

\textit{Rationale:} The timing effect could reflect continuous engagement decay or a discrete advantage of being first. The first counterargument is qualitatively different---guaranteed visibility, no competing frames, and primacy effects.

\textit{Experiment:} Identify first counterargument per argument by earliest timestamp; compare success rates between first vs all other counterarguments; $\chi^2$ test and relative risk calculation.

\textit{Evidence:} First counterargument doubles success rate; suggests position-specific mechanism beyond continuous time effect (First counterargument: 19.9\% vs others: 9.5\%; RR = 2.09, $p{<}0.001$).

\vspace{3pt}

\textbf{Iteration 4:} Response time and counterargument position independently predict success in changing the poster's view

\textit{Rationale:} The first-comment advantage could simply be because it arrives earliest. To distinguish time-based (engagement window) from position-based (visibility/primacy) mechanisms, both must be tested simultaneously.

\textit{Experiment:} Calculate log-transformed time and position rank (within each argument); fit logistic regression: view change $\sim$ log(time) + log(position).

\textit{Evidence:} Two independent effects confirmed; position effect stronger than time effect ($\beta_{\text{time}}{=}-0.23$, $\beta_{\text{pos}}{=}-0.67$, $p{<}10^{-16}$).

\vspace{3pt}

\textbf{Iteration 5:} Timing effect persists after controlling for observable counterargument quality features

\textit{Rationale:} The timing effect might be spurious if skilled persuaders both write higher-quality counterarguments and respond faster. Controlling for content quality tests whether timing is independent of persuasion quality.

\textit{Experiment:} Extract 12 quality features (length, readability, evidence markers, etc.); fit logistic regression with time alone, then add quality controls; compare coefficients.

\textit{Evidence:} Timing effect strengthens with quality controls; higher-quality counterarguments take longer to write; timing is independent predictor beyond content quality ($\beta_{\text{time}}$: $-0.877$ $\to$ $-0.916$, suppression effect).

\vspace{3pt}

\textbf{Iteration 6:} Timing effect persists when the original poster is demonstrably still active in the discussion

\textit{Rationale:} Late counterarguments may fail simply because the poster has abandoned the discussion, never seeing them. Testing within the window where the poster is provably engaged rules out this confounder.

\textit{Experiment:} Estimate poster active window using max(time of last delta awarded, median comment time $\times$ 2, 24h) as conservative estimate; filter to counterarguments posted within active window; re-run time comparison within active-window sample with quality controls.

\textit{Evidence:} Timing effect essentially unchanged; poster abandonment explains virtually none of the effect; effect holds when engagement is confirmed (92.8\% within window; $\beta_{\text{time}}{=}-0.886$ vs $-0.916$ in full sample).

\end{mdframed}
\caption{Refinement trajectory for the timing hypothesis on CMV. Each iteration refines or tests the hypothesis against potential confounders: first-comment advantage (Iter 3), position effects (Iter 4), comment quality (Iter 5), and OP disengagement (Iter 6). The final hypothesis survives all controls.}
\label{fig:cmv_refinement}
\end{figure}

\begin{figure}[t]
\footnotesize
\begin{mdframed}[linewidth=0.5pt, innerleftmargin=10pt, innerrightmargin=10pt, innertopmargin=8pt, innerbottommargin=8pt]

\textbf{Iteration 1:} Footer-embedded forms have markedly lower discovery than above-the-fold or mid-page placements

\textit{Rationale:} Form placement on the page likely affects visibility; footer placement may occur after user drop-off, reducing the chance the form is ever seen.

\textit{Experiment:} Extract \textit{position bucket} (above-the-fold, mid-page, footer) using viewport-relative bounding boxes. Compute view rates by position; test independence with a $\chi^2$ test and report relative risks vs.\ above-the-fold.

\textit{Evidence:} Footer-embedded forms have much lower discovery (N = 418{,}219 sessions): view rate 0.8\% (footer) vs.\ 6.7\% (above-the-fold) and 4.1\% (mid), $p<10^{-16}$. Position is a strong predictor of form discoverability.

\vspace{6pt}

\textbf{Iteration 2:} Horizontally centered forms are more discoverable than off-center forms, independent of position

\textit{Rationale:} Centered placement aligns with user gaze patterns and typical reading flow, making the form easier to visually locate regardless of vertical position.

\textit{Experiment:} Extract \textit{is horizontally centered} using bounding boxes: $|\text{center}_x - 0.5\times\text{viewport}_w| < 0.08\times\text{viewport}_w$ (heuristic), with LLM-as-a-judge adjudication on ambiguous cases. Compare view and sign-up rates for centered vs.\ off-center forms within each position bucket; fit a simple logistic regression for sign-up including centered + position controls.

\textit{Evidence:} Centering improves discovery and conversion within-position (N = 312{,}004 form-exposed sessions): mid-page view rate 5.0\% (centered) vs.\ 3.9\% (off-center), $p<10^{-8}$; sign-up rate 0.97\% vs.\ 0.78\%, $p=2.1\times 10^{-4}$.

\vspace{6pt}

\textbf{Iteration 3:} Forms that are horizontally centered and have a soft shadow effect achieve 3.8$\times$ higher conversion rates

\textit{Rationale:} A soft shadow provides depth cues that separate the form from surrounding content, improving visual hierarchy and signaling interactivity. Combined with centering, this may substantially boost conversion beyond either feature alone.

\textit{Experiment:} Extract \textit{has soft shadow} from CSS (non-zero box-shadow blur radius with opacity $<0.35$) plus LLM-as-a-judge validation. Compare sign-up rates for (centered + shadow) vs.\ (off-center or no-shadow) within each position bucket; compute relative risk and fit logistic regression including shadow + centered + position.

\textit{Evidence:} Centered forms with soft shadow achieve 3.8$\times$ higher conversion rate (N = 312{,}004): sign-up rate 1.08\% (centered + shadow) vs.\ 0.28\% (other combinations), $p<10^{-10}$. The combined effect is multiplicative rather than additive.

\vspace{6pt}

\textbf{Iteration 4:} Horizontally centered, soft-shadow forms improve discoverability and conversions after controlling for page type, position, and time-varying factors

\textit{Rationale:} Discovery and conversion vary across page types (e.g., homepage vs.\ content hub) and over time (seasonality, campaigns). A controlled model is needed to confirm that centering and shadow effects are independent of these confounders.

\textit{Experiment:} Use extracted features: \textit{has soft shadow}, \textit{is horizontally centered}, \textit{position bucket} (above/mid/footer), \textit{page type} (homepage, product/feature, content hub/blog, pricing/landing), and date/time covariates (day-of-week, hour, week). Fit logistic regression on sign-up with robust SEs (clustered by page), including position, page type, and time fixed effects.

\textit{Evidence:} In the controlled logit , centered and soft-shadow indicators remain positive and significant (all $p<0.01$). Footer position remains strongly negative vs.\ above-the-fold ($p<10^{-6}$).
\end{mdframed}
\caption{Refinement trajectory for the lead-generation form discovery and conversion hypothesis. Each iteration tests and refines the initial observation (footer forms underperform) by isolating actionable design features (horizontal centering, soft shadow) and controlling for confounders (position, page type, time).}
\label{fig:leadgen_refinement}
\end{figure}

\begin{figure}[t]
\footnotesize
\begin{mdframed}[linewidth=0.5pt, innerleftmargin=10pt, innerrightmargin=10pt, innertopmargin=8pt, innerbottommargin=8pt]

\textbf{Iteration 1:} ``I ask unanimous consent'' predicts Republican speakers

\textit{Rationale:} The phrase may be a partisan rhetorical marker if it is disproportionately used by one party after accounting for base rates.

\textit{Experiment:} Use the code interpreter to define $x{=}1$ if the exact phrase ``I ask unanimous consent'' occurs (case-insensitive, punctuation-normalized), else $0$. Fit a baseline logit:
\[
\mathbb{1}[\text{Republican}] \sim x
\]
Report the coefficient $\beta_x$, two-sided Wald $p$-value, and out-of-sample accuracy from a random train/test split.

\textit{Evidence:} Positive association with Republican: $\beta{=}0.30$, $p{<}0.001$; accuracy $\approx$ 53\%.

\vspace{6pt}

\textbf{Iteration 2:} The phrase is procedural rather than ideological

\textit{Rationale:} ``Unanimous consent'' is a parliamentary procedure; usage could reflect institutional role (e.g., chair/floor manager) rather than ideology.

\textit{Experiment:} Map each speech to a coarse speaker role using metadata (chair/presiding officer, floor manager, other member). Compute (i) phrase rate per role ($P[x{=}1\mid \text{role}]$) and (ii) role share among phrase-containing speeches ($P[\text{role}\mid x{=}1]$). Test independence of $x$ and role with a $\chi^2$ test, and report the top roles driving the residuals.

\textit{Evidence:} Usage concentrates in roles that expedite routine business; pattern consistent with procedural function, not partisan content.

\vspace{6pt}

\textbf{Iteration 3:} Majority-party status explains much of the phrase usage

\textit{Rationale:} The majority party more often controls the floor agenda and uses procedural moves to manage time, which could drive the observed party correlation.

\textit{Experiment:} For each speech, assign a majority/minority indicator based on the majority party for that Congress session. Compare phrase usage rates between majority vs.\ minority speakers and report both relative risk and a two-proportion z-test (or $\chi^2$) for $P[x{=}1\mid \text{majority}]$ vs.\ $P[x{=}1\mid \text{minority}]$.

\textit{Evidence:} Majority party uses the phrase $\approx 3.2\times$ more often.

\vspace{6pt}

\textbf{Iteration 4:} The dataset is imbalanced across sessions and majority-party regimes

\textit{Rationale:} If the corpus over-represents Congresses where Republicans are the majority, party and majority status become highly collinear, inflating pooled correlations.

\textit{Experiment:} Aggregate the corpus by Congress/session. For each session, compute total speeches $N_s$ and tag whether Republicans or Democrats are the majority. Summarize the distribution of $N_s$ by majority regime (counts, proportions, and the max/min ratio) and compute the overlap between party label and majority status induced by the session mix.

\textit{Evidence:} Severe skew: $\sim$20:1 imbalance; $\approx$95\% of speeches come from Republican-majority sessions.

\vspace{6pt}

\textbf{Iteration 5:} The phrase predicts majority-party status, not party identity

\textit{Rationale:} To separate procedural control from partisanship, party and majority status must be jointly modeled within the same session context.

\textit{Experiment:} Fit a within-session model that includes both predictors and absorbs session-specific base rates. Concretely, estimate:
\[
\mathbb{1}[x{=}1] \sim \mathbb{1}[\text{majority}] + \mathbb{1}[\text{Republican}] + \text{session fixed effects}
\]
Report coefficients and $p$-values with robust SEs (clustered by session). Check collinearity and confirm that the majority coefficient remains stable while the party coefficient shrinks toward zero.

\textit{Evidence:} Majority status remains strong: $\beta_{\text{maj}}{=}0.28$ ($p{<}0.001$), while party vanishes: $\beta_{\text{party}}{=}0.02$ (n.s.).

\vspace{6pt}

\textbf{Iteration 6:} The apparent party effect reverses in Democratic-majority sessions (OOD check)

\textit{Rationale:} If the phrase is a majority-party procedural artifact, then in a Democratic-majority Congress the direction should flip.

\textit{Experiment:} Hold out a Democratic-majority Congress (the 103rd) as an out-of-distribution slice. Within that slice, re-estimate both the pooled party-only logit (Iter 1) and the joint model with majority controls (Iter 5). Compare the sign/magnitude of the party coefficient, and compute phrase rates by party to verify the direction directly.

\textit{Evidence:} The coefficient reverses; Democrats use the phrase more often in Democratic-majority sessions.
\end{mdframed}
\caption{Rejection trajectory for the ``unanimous consent'' hypothesis on Congress. The initial correlation (Iter 1) is statistically significant but confounded with majority-party status (Iter 3--5). Out-of-distribution validation (Iter 6) confirms the reversal. The hypothesis is rejected as a dataset artifact.}
\label{tab:congress_rejection}
\end{figure}

\subsection{Rejection: Procedural Phrases Confounded with Floor Control}
\label{sec:case_congress}

The Congressional Record contains 140,000 speeches from the 104th--109th Congress. The task is predicting party affiliation (Republican or Democrat) from speech text. Each speech includes session metadata (Congress number, majority party) appended as verbalization.

Both HypotheSAEs and ExperiGen discover that the phrase ``I ask unanimous consent'' correlates with Republican speeches ($\beta{=}0.30$, $p{<}0.001$). This correlation is statistically significant and replicable. HypotheSAEs, operating at depth-1, accepts this as a valid partisan signal and adds it to the hypothesis bank. ExperiGen subjects the same hypothesis to iterative testing.

Listing~\ref{tab:congress_rejection} traces the 6-iteration refinement. The Generator's follow-up questions reveal a confound: the phrase is procedural (used to expedite routine business), and procedural control belongs to the majority party. Since 95\% of speeches in the dataset come from Republican-majority sessions (104th--109th Congress), the correlation reflects floor control, not ideology. Testing on Democrat-majority sessions (103rd Congress) reverses the coefficient. The hypothesis is rejected.

\paragraph{Why Depth-1 Methods Retain Spurious Hypotheses.} The session metadata is available to all methods (verbalized for HypotheSAEs). The failure is not information access but \emph{lack of iterative testing}. Depth-1 methods propose hypotheses and measure their predictive accuracy; they do not generate follow-up experiments to probe whether correlations reflect causal mechanisms or confounds. HypoGeniC's 5-sample window cannot detect the 20:1 session imbalance. HypotheSAEs could in principle discover the confound, but without a mechanism for iterative questioning, it never asks whether the correlation holds across majority-party regimes. ExperiGen's Generator, prompted by the Experimenter's results, systematically probes alternative explanations until the confound is exposed.

\begin{figure}[t]
    \centering
    \includegraphics[width=0.8\textwidth]{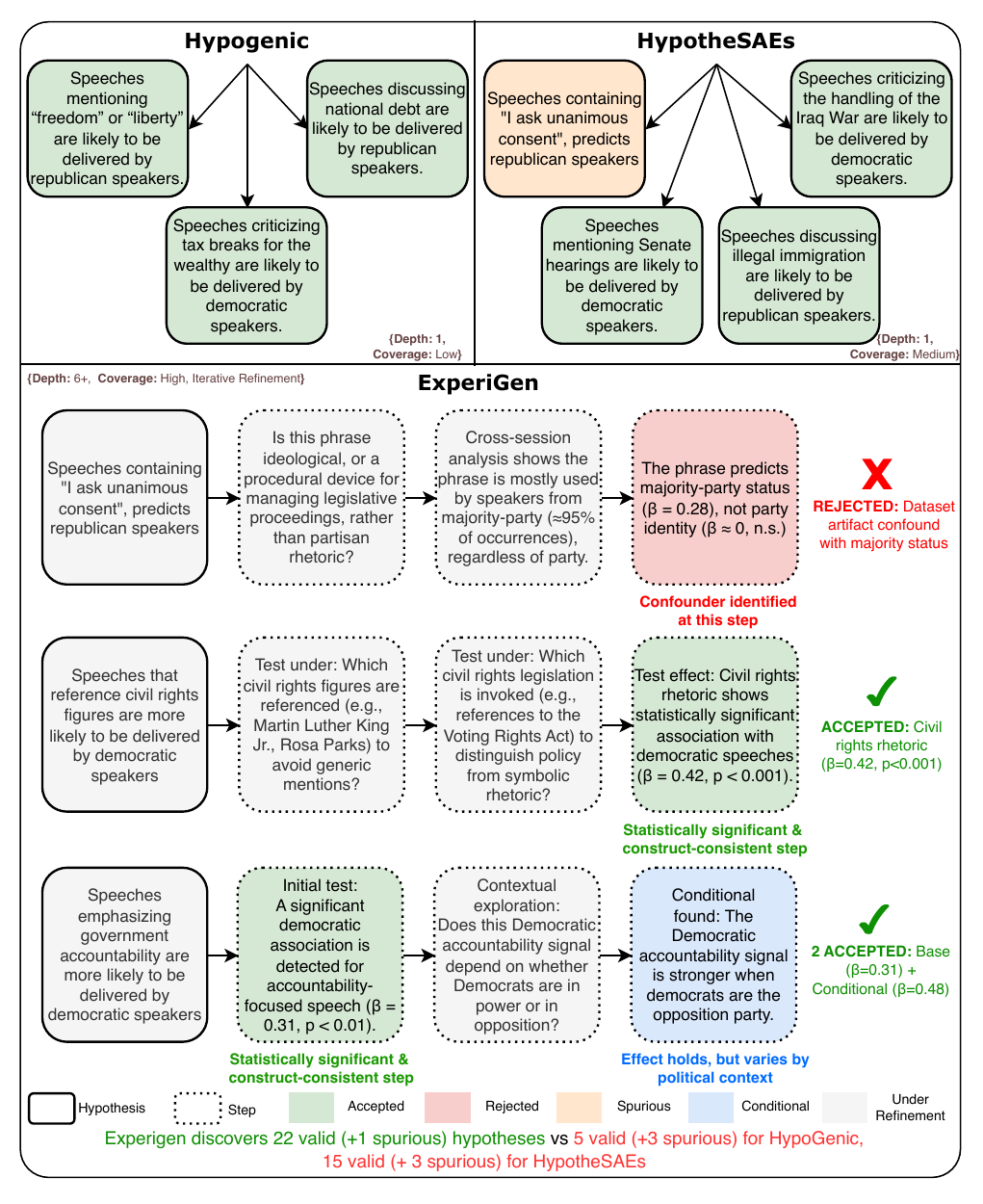}
    \caption{
    \textbf{Hypothesis space exploration on the Congress dataset} (see \cref{sec:case_congress} for full analysis).
    \textbf{(Top Left)} HypoGeniC: depth=1, low coverage---clusters around 3 top-performing hypotheses.
    \textbf{(Top Right)} HypotheSAEs: depth=1, medium coverage---discovers 4 hypotheses including the spurious ``I ask unanimous consent'' (orange), which it retains without iterative testing to expose the confound.
    \textbf{(Bottom)} ExperiGen with \textbf{Iterative Refinement}: depth=6+, high coverage---shows three parallel refinement paths.
    \emph{(i) Rejection path:} ``I ask unanimous consent'' undergoes procedural analysis $\to$ cross-session analysis reveals 20:1 session skew $\to$ confound with majority-party status identified $\to$ \textbf{rejected} as dataset artifact.
    \emph{(ii) Acceptance path:} ``Civil rights mentions'' refined through key figures $\to$ Voting Rights Act references $\to$ Democrat-sponsored bills analysis $\to$ \textbf{accepted} ($\beta=0.42$, $p<10^{-5}$).
    \emph{(iii) Conditional path:} ``Govt.\ accountability'' initially accepted ($\beta=0.31$) $\to$ continued exploration discovers conditional hypothesis ``when opposing party in power'' with stronger effect ($\beta=0.48$) $\to$ \textbf{2 hypotheses accepted} (base + conditional).
    This demonstrates that iterative refinement not only rejects spurious correlations but also discovers conditional sub-hypotheses that strengthen predictive power.
    \textbf{Legend:} \textcolor{green!60!black}{Green} = accepted; \textcolor{orange!80!black}{Orange} = spurious (retained by depth-1 methods); \textcolor{blue!60!black}{Blue} = under refinement; \textcolor{red!60!black}{Red} = rejected.
    }
    \label{fig:hypothesis_tree}
\end{figure}

\paragraph{Visualization.}
Figure~\ref{fig:hypothesis_tree} contrasts the three methods. HypoGeniC and HypotheSAEs operate at depth-1: HypotheSAEs retains the spurious ``unanimous consent'' hypothesis (orange). ExperiGen's 6-iteration rejection path appears alongside an acceptance path for ``civil rights mentions'' ($\beta{=}0.42$, $p{<}10^{-5}$), which survives session-stratified testing.

\subsection{Discovery: Centered Forms with Visual Elevation Improve Engagement}
Lead-generation forms on commercial websites offer a real-world setting to study how visual design and layout affect user attention and conversion. In partnership with a Fortune 500 consumer brand, we analyzed a large deployment spanning thousands of page visits, where each form instance was paired with screenshots, page metadata, timestamps, view counts, and sign-up outcomes.

ExperiGen’s Generator began by observing that footer-embedded forms had markedly lower discovery rates than above-the-fold or mid-page placements. The Experimenter extracted visual and structural features from screenshots using heuristics and LLM-as-a-judge annotations, including \texttt{position bucket}, \texttt{is horizontally centered}, \texttt{has soft shadow}, and \texttt{page type}. The subsequent 4-iteration refinement (Listing~\ref{fig:leadgen_refinement}) builds on this initial observation: (1) establishing that centering strongly predicts discoverability, (2) showing that horizontally centered forms outperform off-center forms within each position bucket, (3) demonstrating that adding soft shadow (visual elevation) provides incremental benefit beyond centering, and (4) confirming these effects survive controls for page type and time-varying factors.

The refined hypothesis states that horizontally centered forms with soft shadow improve both discoverability (form views) and conversion (sign-ups), independent of \ position, page type, or time. This hypothesis was pre-registered and an A/B test was launched to validate the finding.

\begin{figure}[H]
\label{fig : Forms}
    \centering
    \includegraphics[height=0.15\textheight,keepaspectratio]{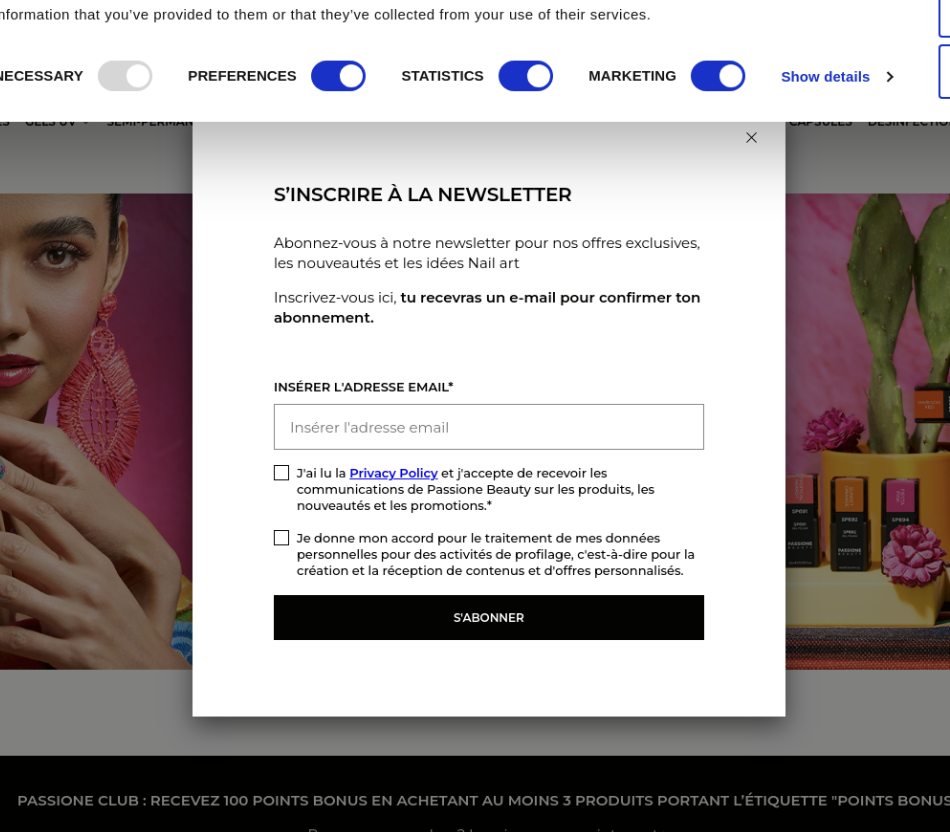}
    \hspace{0.02\textwidth}
    \includegraphics[height=0.15\textheight,keepaspectratio]{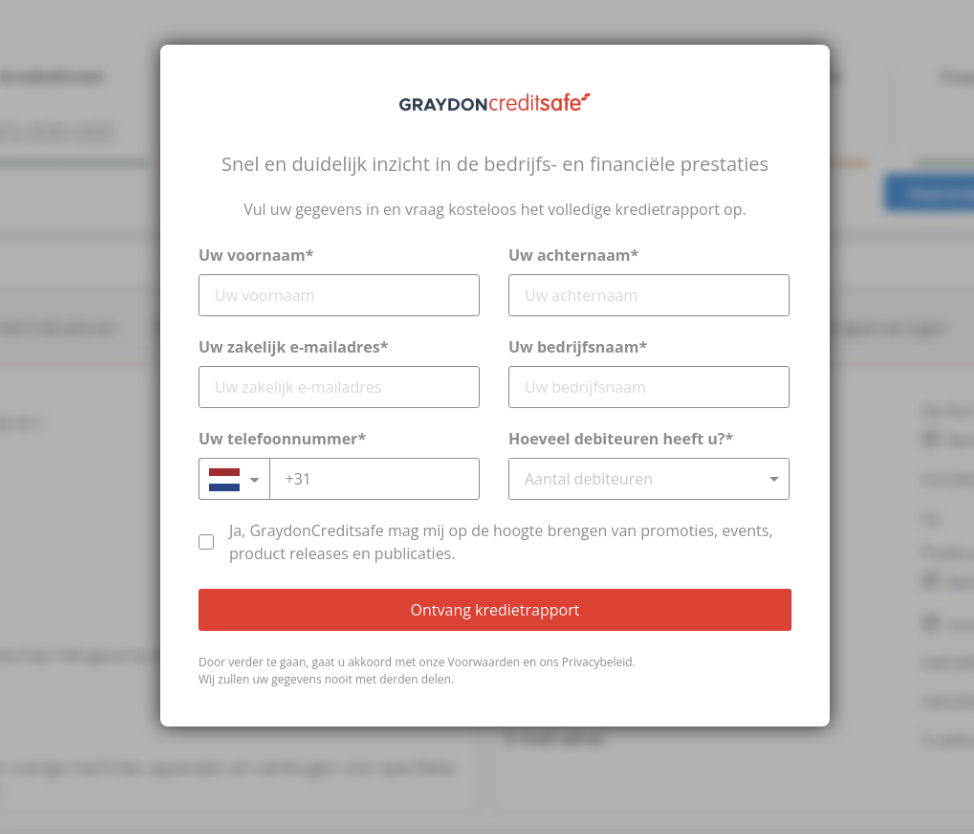}
    \hspace{0.02\textwidth}
    \includegraphics[height=0.15\textheight,keepaspectratio]{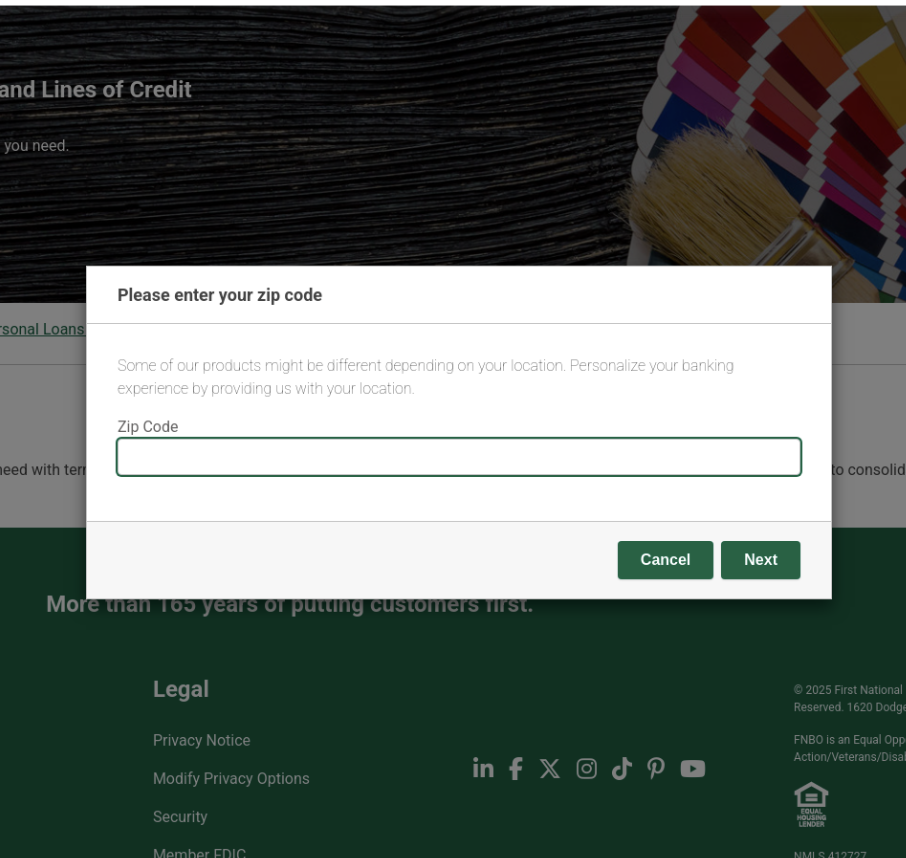}

    \caption{Hypothesis Generated: ``Forms that are horizontally centred and have a soft shadow effect on their background achieve 3.8x higher conversion rates.'' }

\end{figure}

%% file: pages/appendix/examples.tex
\section{Discovered Hypotheses}
\label{sec:discovered_hyp}
\begin{longtable}[H]{@{}p{0.15\linewidth}p{0.55\linewidth}p{0.2\linewidth}@{}}
\toprule
\textbf{Dataset} & \textbf{Finding} & \textbf{Supported/Novel} \\
\midrule
\endfirsthead

\multicolumn{3}{c}
{{\bfseries \tablename\ \thetable{} -- continued from previous page}} \\
\toprule
\textbf{Dataset} & \textbf{Finding} & \textbf{Supported/Novel} \\
\midrule
\endhead

\bottomrule
\multicolumn{3}{r}{{Continued on next page}} \\
\endfoot

\endlastfoot

\multirow{5}{=}{\textbf{DREADDIT}\\[0.3em]\footnotesize(Given a reddit post, the task is to predict whether it contains mental stress signals)}
& Posts that exhibit co-occurring negative intensifier adverbs and negative emotion words (e.g. `extremely anxious') express stress. & \cite{popets1} \\

& Posts that exhibit a higher co-occurrence of persistent adverbs (e.g., `always', `never') with negation words (e.g., `not', `no') express stress. & \cite{popets1} \\

& The ratio of first-person singular pronouns to second-person pronouns within question sentences is significantly higher in stressful posts, as self-referential questioning with minimal interpersonal focus may reflect internalized distress in mental health narratives. & \cite{lasalita-etal-2024-language} \\

& Posts that have a lower presence of positive emotion words express stress. & \cite{pugach2023positive} \\

& Posts with a higher proportion of `exclamation'-type rhetorical questions express stress. & \textbf{Novel} \\

& Posts with higher frequency of body-related verbs (e.g., `tremble', `sweat', `ache') express stress & \textbf{Novel} \\
\midrule

\multirow{5}{=}{\textbf{HEADLINES}\\[0.3em]\footnotesize(Given a pair of headlines, the task is to predict which headline is more engaging)}

& "Headlines that utilize curiosity gap techniques, such as implying secret insights while providing partial information, are more engaging. & \cite{Matias2025} \\

& Headlines including at least one contraction or colloquial phrase are more engaging. & \cite{7752207} \\

& Headlines containing power words, specifically `free,' `breakthrough,' `exclusive,' `new,' and `secret,' are more engaging. & \cite{banerjee2021} \\

& Headlines containing narrative elements (e.g., `characters', `conflict') receive more clicks. & \cite{banerjee2021} \\

& Headlines that include shocking or surprising elements are more engaging & \cite{robertson2023} \\

& Headlines combining causal conjunctions with sensory language receive more clicks. & \cite{banerjee2021} \\

& Headlines with specific time references (e.g., `today') receive more engagement. & \cite{banerjee2021} \\

& Headlines with pop culture references to movies are more engaging.
 & \textbf{Novel} \\

& Headlines that combine alliteration and high-emotion words receive more clicks. & \textbf{Novel} \\

& Headlines using active voice with strong action verbs are more engaging. & \textbf{Novel} \\

\midrule
\multirow{5}{=}{\textbf{GPTGC}\\[0.3em]\footnotesize(Given a story, the task is to predict whether it was written by Human or AI)}

& Machine-generated narratives demonstrate significantly longer average sentence lengths and higher Flesch-Kincaid grade levels compared to human-authored content & \cite{Guo2023HowCI} \\

& The presence of informal abbreviations or spelling errors is positively correlated with human-authored text. & \cite{bahrini2023} \\

& Human-authored stories exhibit a higher frequency of technical terms (e.g., 'translation', 'intercept') and political keywords (e.g., 'government', 'federal funding') co-occurring within a 10-word window compared to AI-generated stories, reflecting more deliberate narrative integration of technical-political discourse. & \textbf{Novel}\\

& The use of specific ideological phrases, such as 'alternative facts', is significantly more prevalent in human-authored stories when analyzed via Fisher's exact test. & \textbf{Novel} \\

\midrule
\multirow{5}{=}{\textbf{Deceptive Reviews}\\[0.3em]\footnotesize(Given a hotel review, the task is to predict whether it is real or deceptive)}

& Deceptive reviews exhibit a significantly higher frequency of intensifier words (e.g., 'very', 'extremely') and extreme sentiment polarities compared to truthful reviews to exaggerate emotional emphasis. & \cite{Ott2011FindingDO}\\

& Truthful reviews contain a significantly higher frequency of precise financial details, exact monetary amounts, and numerical specifics (e.g., room numbers, exact durations) compared to deceptive reviews. & \cite{luca_zervas2016} \\

& Deceptive reviews exhibit significantly lower frequencies of verifiable temporal markers (e.g., exact dates/times) directly linked to service mentions, reflecting a lack of episodic memory. & \textbf{Novel} \\

& Truthful reviews exhibit higher "Contextual Linkage," where branded amenities and services are directly paired with specific staff actions or usage flows. & \textbf{Novel} \\

\midrule

\multirow{5}{=}{\textbf{Lamem}\\[0.3em]\footnotesize(Given a pair of headlines, the task is to predict which headline is more engaging)}

& Images evoking positive emotional valence (e.g., happiness) are more effectively encoded and preferred over images with negative valence (e.g., anger). & \cite{pan2025} \\

& The preference for brighter images increases quadratically as a function of the brightness difference between pairs, rather than linearly. & \textbf{Novel} \\

& Nature' and 'Object' categories demonstrate higher memorability scores compared to 'People' or 'Animals' in multi-image evaluation tasks. & \textbf{Novel} \\

\midrule
\multirow{5}{=}{\textbf{Design}\\[0.3em]\footnotesize(Given a pair of designs, the task is to predict which design is more aesthetic)}

& Layouts with centered alignment are more likely to be preferred over asymmetrical layouts. & \cite{bargas2010} \\

& Layouts with medium white space distribution are more likely to be preferred. & \cite{Chuming2019} \\

& When layouts have low content density, the combination of high image quality and a clear visual hierarchy most strongly improves layout preference. & \textbf{Novel} \\

\midrule
\multirow{5}{=}{\textbf{CMV Reviews}\\[0.3em]\footnotesize(Given an opinion post, the task is to predict which counterarguments successfully change the author’s view)}

& Counterarguments framed to align with the audience's existing values or moral foundations are more persuasive than counterarguments that conflict with those values. & \cite{tan_winning_arguments2016}\\

& Counterarguments that include concessions (acknowledging part of the original argument) are more effective in changing views than counterarguments without concessions. & \cite{Musi2018ChangeMyViewTC} \cite{messerli2025}\\
& Counterarguments that provide contextual explanations or analogies for technical terms are more likely to change views compared to counterarguments that merely list technical terms without elaboration. &	\textbf{Novel} \\
& Counterarguments that explicitly narrow the scope of the debate (e.g., 'let's focus on,' 'setting aside,' 'for the sake of argument') are more effective at changing views than those that address the full breadth of the OP's argument. &	\textbf{Novel} \\

\midrule
\multirow{5}{=}{\textbf{Congress}\\[0.3em]\footnotesize(Given a speech, the task is to predict whether the speaker is from Republican or Democratic part in American congress)}

& Speeches containing terms related to economic policy (e.g., 'tax', 'budget', 'free market') are more likely to be associated with Republican speakers compared to Democratic speakers. & \cite{gentzkow2019_congress_speechs}\\
& Speeches containing higher frequencies of fear-related terms (e.g., 'threat', 'danger', 'security') in the context of national security or immigration are more strongly associated with Republican affiliation.  & \cite{card2019_congress_speech}\\
& Speeches containing parental involvement terms are more strongly associated with Democratic affiliation compared to Republican affiliation. & \textbf{Novel} \\
& The interaction between the frequency of 'small business' terms and 'tax policy' terms in a speech has a statistically significant and stronger predictive relationship with Republican affiliation. &  \textbf{Novel}\\
\bottomrule
\caption{ExperiGen-generated hypotheses across multiple datasets and tasks. For each dataset, we list representative linguistic, semantic, visual, or discourse-level hypotheses associated with the target prediction task. Each hypothesis is annotated as either supported by prior literature (with citations) or identified as novel, highlighting both confirmed patterns and newly uncovered relationships.}

\label{tab:hypotheses_examples_established_and_novel} \\
\end{longtable}

\begin{table}[ht!]
\centering
\small
\begin{tabular}{l|ccccc|ccccc}
\toprule
& \multicolumn{5}{c|}{\textbf{HypoBench}} & \multicolumn{5}{c}{\textbf{Cross-Domain}} \\
\textbf{Method} & \textbf{Decep.} & \textbf{News} & \textbf{Dread.} & \textbf{GPTgc} & \textbf{Pers.} & \textbf{Twitter} & \textbf{Design} & \textbf{LaMem} & \textbf{Cong.} & \textbf{CMV} \\
\midrule
ExperiGen & \textbf{18} & \textbf{21} & \textbf{16} & \textbf{14} & \textbf{17} & \textbf{19} & \textbf{11} & \textbf{5} & \textbf{20} & \textbf{16} \\
HypoGenic & 6 & 3 & 4 & 3 & 5 & 3 & 4 & 0 & 5 & 4 \\
HypotheSAEs & 8 & 12 & 6 & 5 & 7 & 12 & 6 & 0 & 12 & 8 \\
\bottomrule
\end{tabular}
\caption{Number of statistically significant hypotheses ($p < 0.05$, Bonferroni-corrected) across all datasets. ExperiGen discovers 2--4$\times$ more significant hypotheses than baselines. On LaMem, ExperiGen is the only method to discover any significant hypotheses.}
\label{tab:sig_hypotheses}
\end{table}

\subsection{Qualitative Examples}
\subsubsection{Layout Design}
\begin{figure}[H]
    \centering
    \includegraphics[width=1\linewidth]{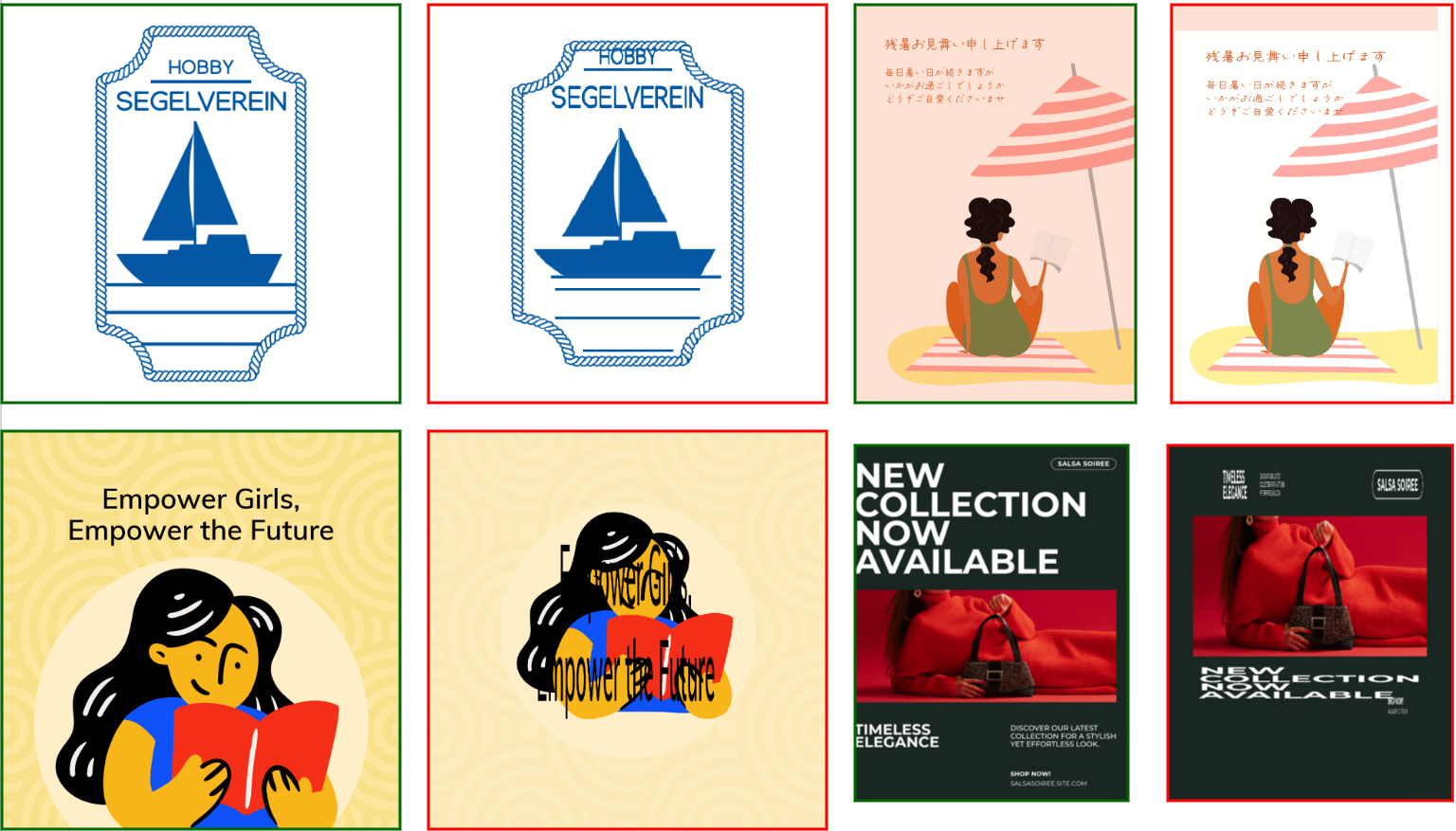}
    \caption{Hypothesis Generated: Layouts with balanced spacing and margins are more likely to be real.}
    \label{fig:layout-example1}
\end{figure}

\begin{figure}[H]
    \centering
    \includegraphics[width=1\linewidth]{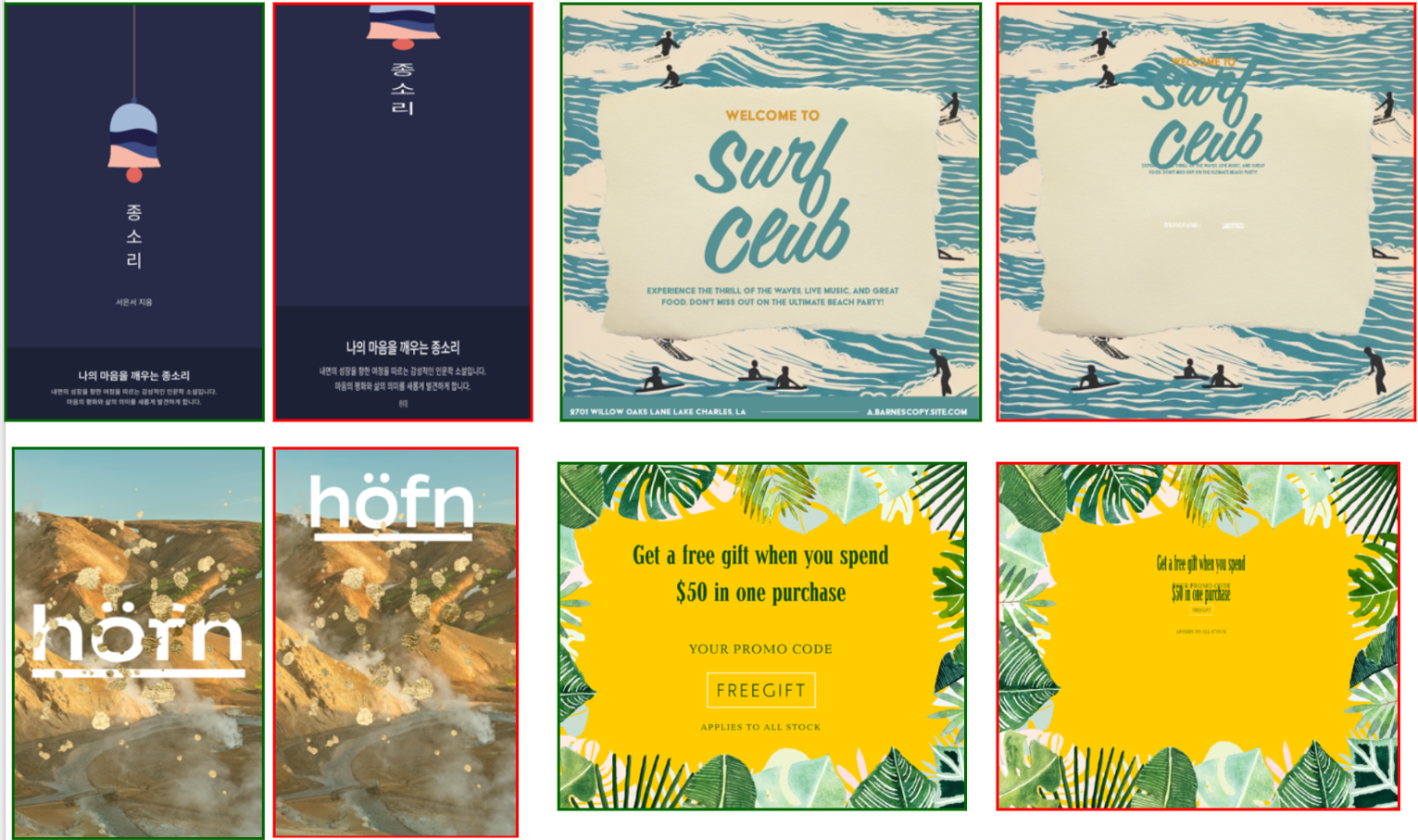}
    \caption{Hypothesis Generated: Layouts with centered alignment are more likely to be preferred over asymmetrical layouts in
direct comparisons when they are presented against each other.}
    \label{fig:layout-example2}
\end{figure}
\subsubsection{News Headlines Popularity}
\begin{figure}[H]
    \centering
    \begin{tabular}{m{0.45\textwidth} m{0.45\textwidth}}
        \begin{tcolorbox}[colback=green!5, colframe=green!70!black,
                          arc=3pt, boxrule=0.8pt, width=0.42\textwidth]
            {\sffamily\bfseries\large
            ``Why There Are So Many Blue Dots On This Map, And Why We Should All Care''}
        \end{tcolorbox}
        &
        \begin{tcolorbox}[colback=red!5, colframe=red!70!black,
                          arc=3pt, boxrule=0.8pt, width=0.42\textwidth]
            {\sffamily\bfseries\large
            ``I Loved The Pretty Dots On This Map ... But Upon Further Inspection, I Felt Sick To My Stomach''}
        \end{tcolorbox}
    \end{tabular}
    \caption{\sffamily \textbf{Hypothesis:} Headlines using a
    \emph{cause–effect structure} (e.g., starting with ``Why’’) are more engaging,
    as they promise a clear causal explanation.}
\end{figure}

\begin{figure}[H]
    \centering
    \begin{tabular}{m{0.45\textwidth} m{0.45\textwidth}}
        \begin{tcolorbox}[colback=green!5, colframe=green!70!black,
                          arc=3pt, boxrule=0.8pt, width=0.42\textwidth]
            {\sffamily\bfseries\large
            ``There's A Secret That These People Keep Every Day They Go To Work''}
        \end{tcolorbox}
        &
        \begin{tcolorbox}[colback=red!5, colframe=red!70!black,
                          arc=3pt, boxrule=0.8pt, width=0.42\textwidth]
            {\sffamily\bfseries\large
            ``What One Person Did To Help Others Overcome Bullying In The Workplace''}
        \end{tcolorbox}
    \end{tabular}
    \caption{\sffamily \textbf{Hypothesis:} Headlines that include \emph{curiosity-inducing keywords} (e.g., 'secret', 'never told', 'shocking') are more likely to be the winning headline compared to those without such keywords, as they trigger the reader's curiosity and desire for information.}
\end{figure}

\subsubsection{DREADDIT: Stress Detection}
\definecolor{helplessblue}{RGB}{33,113,181}
\definecolor{fearred}{RGB}{203,24,29}
\definecolor{barrierorange}{RGB}{255,127,0}
\definecolor{disillusiongreen}{RGB}{9,121,105}

\noindent\textbf{Example Post from Reddit, taken from Dreaddit dataset}

\begin{tcolorbox}[
    colback=gray!5,
    colframe=black!40,
    boxrule=0.4pt,
    arc=2pt,
    left=6pt,
    right=6pt,
    top=6pt,
    bottom=6pt
]
\textit{I was a severe addict that had 2 overdoses when I was younger and addiction/alcoholism runs on both sides of my family.
\textcolor{helplessblue}{Please tell me the ``hijacking'' will stop} and I will come back into my own person.
\textcolor{fearred}{I don't want to come out} on the other side and be this nervous and uncomfortable person who is unable to have relationships.
I don't drink, or take prescriptions, or smoke.
I have been thinking of seeing a therapist,
\textcolor{barrierorange}{I don't really have the time or extra money},
plus I've been really
\textcolor{disillusiongreen}{hurt from small intimate AA groups}.
Ugh\ldots\ Is there another side of this PTSD mental attrition?}
\end{tcolorbox}

\vspace{0.3em}
\noindent\textbf{Generated Hypothesis:}
Posts that exhibit a higher frequency of negated solution descriptors (e.g., ``can't fix'', ``don't know how'', ``won't help'')
are more likely to express elevated stress, as such descriptors reflect an external locus of control and a perceived lack of effective solutions.
\newline
\noindent\textbf{Stress Signal Annotations:}

\begin{itemize}
    \item \textcolor{helplessblue}{\rule{0.8em}{0.8em}} \;
    Suggests helplessness, desire for external reassurance, and lack of personal control over the situation.

    \item \textcolor{fearred}{\rule{0.8em}{0.8em}} \;
    Implies fear of a negative outcome they can't avoid or fix.

    \item \textcolor{barrierorange}{\rule{0.8em}{0.8em}} \;
    A direct negation of feasible solutions like therapy, indicating perceived barriers to help.

    \item \textcolor{disillusiongreen}{\rule{0.8em}{0.8em}} \;
    Shows disillusionment with a prior solution, reinforcing the feeling that support “won’t help”.
\end{itemize}

\vspace{0.3em}
\noindent\textbf{Hypothesis Validation:}
The highlighted segments collectively demonstrate multiple forms of negated solution descriptors, including explicit rejection of therapy,
disillusionment with prior support systems, and reliance on external reassurance. As shown by the color-coded annotations above,
these linguistic cues align with the proposed hypothesis and provide qualitative evidence that the post reflects elevated stress.

\subsubsection{LaMem: Image Memorability}

\begin{figure}[H]
    \centering
    \includegraphics[width=0.6\linewidth]{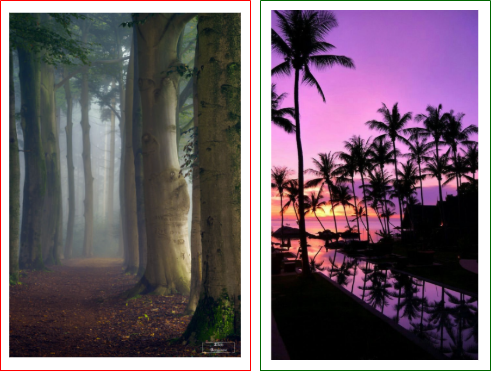}
    \caption{Higher relative color contrast significantly predicts image preference, with the higher-contrast image being consistently more memorable.}
    \label{fig:mem-example}
\end{figure}

\subsubsection{Congress: Party Affiliation}

\definecolor{marketblue}{RGB}{33,113,181}
\definecolor{tradegreen}{RGB}{9,121,105}
\definecolor{priceorange}{RGB}{255,127,0}
\definecolor{industryred}{RGB}{203,24,29}

\noindent\textbf{Example Speech from U.S. Congress by a Republican Speaker}

\begin{tcolorbox}[
    colback=gray!5,
    colframe=black!40,
    boxrule=0.4pt,
    arc=2pt,
    left=6pt,
    right=6pt,
    top=6pt,
    bottom=6pt
]
\textit{
Mr. President, in today’s Federal Register, the United States Department of Agriculture has published a final rule that could have significant adverse impacts on our
\textcolor{industryred}{domestic cattle industry}.
The regulation designates Canada as a ``minimal risk region'' for BSE and allows Canada to export more beef to the United States.
It is the same frustration my Montana cattlemen feel as they watch Australia and New Zealand
\textcolor{marketblue}{expand market share in the Pacific Rim},
while those
\textcolor{tradegreen}{markets remain closed off to the U.S.}
These flaws could harm
\textcolor{priceorange}{domestic consumer confidence}
and delay the
\textcolor{tradegreen}{reopening of international markets}.
Important
\textcolor{tradegreen}{export markets around the world closed their doors to U.S. beef}.
USDA has been working with Japan and Korea to
\textcolor{tradegreen}{reopen markets to U.S. beef},
but
\textcolor{tradegreen}{exports have not yet started}
and could jeopardize our relationship with Japan.
With few exceptions, that trade has had little impact on
\textcolor{priceorange}{U.S. cattle prices}.
}
\end{tcolorbox}

\vspace{0.3em}
\noindent\textbf{Generated Hypothesis:}
Speeches containing terms related to economic policy (e.g., taxation, trade, market access, pricing)
are more likely to be associated with Republican speakers compared to Democratic speakers.

\noindent\textbf{Economic Policy Signal Annotations:}

\begin{itemize}
    \item \textcolor{marketblue}{\rule{0.8em}{0.8em}} \;
    Indicates market-oriented framing through references to competition, market share, and access.

    \item \textcolor{tradegreen}{\rule{0.8em}{0.8em}} \;
    Captures economic policy related to trade, exports, and international market regulation.

    \item \textcolor{priceorange}{\rule{0.8em}{0.8em}} \;
    Reflects price sensitivity and economic consequences for domestic producers and consumers.

    \item \textcolor{industryred}{\rule{0.8em}{0.8em}} \;
    Emphasizes protection of domestic industry, a common Republican economic framing.
\end{itemize}

\vspace{0.3em}
\noindent\textbf{Hypothesis Validation:}
The highlighted segments reveal a dense concentration of economic-policy language centered on trade,
market access, pricing, and industry competitiveness. Rather than framing the issue in terms of public
welfare or redistribution, the speaker emphasizes market outcomes and regulatory impacts on domestic
producers. These lexical cues align with the proposed hypothesis and correctly predict Republican
party affiliation, providing qualitative evidence that economic-policy framing serves as a
discriminative signal of partisan identity in congressional speech.

\subsubsection{{Change My View (CMV) Case Studies}}

This subsection evaluates four representative Change My View (CMV) discussions to test the following hypothesis regarding the role of semantic clarification in persuasive success.

\begin{tcolorbox}[
    colback=white,
    colframe=black!75!black,
    colbacktitle=black,
    coltitle=white,
    title=Research Hypothesis,
    fonttitle=\bfseries,
    breakable,
    enhanced
]
Counterarguments that explicitly redefine or clarify key terms in the original argument using precise and authoritative definitions are more effective in changing the original poster’s view than counterarguments that engage only at the level of moral judgment or emotional disagreement without resolving underlying semantic ambiguities.
\end{tcolorbox}

\begin{tcolorbox}[
    colback=red!5,
    colframe=red!75!black,
    colbacktitle=red!75!black,
    coltitle=white,
    title=Case 1: Validation of Feelings (Failed Persuasion),
    fonttitle=\bfseries,
    breakable,
    enhanced
]
\textbf{Thread Title:} CMV: The practice of validating another's feelings is breeding ingenuine people

\textbf{OP Original View:}
The OP argued that validating another person’s feelings necessarily entails agreement with them, thereby producing hypocrisy because disagreement becomes impossible.

\textbf{Key Conceptual Flaw:}
Conflation of \emph{acknowledging emotional experience} with \emph{endorsing its correctness}.

\textbf{Commenter Strategy:}
The commenter rejected the OP’s emotional stance without redefining or clarifying the meaning of the term \emph{validation}.

\textbf{Representative Rebuttal:}
“You can understand feelings and totally invalidate them. You have a right to feelings but they are wrong.”

\textbf{Outcome:}
No delta was awarded. The OP reiterated that, in common usage, validation implies agreement.

\textbf{Failure Mechanism:}
The commenter failed to challenge the OP’s underlying semantic assumption. By accepting the colloquial definition of “validation,” the rebuttal addressed surface disagreement rather than the conceptual confusion driving the claim.
\end{tcolorbox}

\begin{tcolorbox}[
    colback=red!5,
    colframe=red!75!black,
    colbacktitle=red!75!black,
    coltitle=white,
    title=Case 2: Choosing Not to Be Offended (Failed Persuasion),
    fonttitle=\bfseries,
    breakable,
    enhanced
]
\textbf{Thread Title:} CMV: People can simply choose not to be offended

\textbf{OP Original View:}
The OP defined offense as an involuntary emotional reaction and argued that advising individuals to “choose not to be offended” constitutes victim blaming.

\textbf{Key Conceptual Flaw:}
Ambiguity between offense as an automatic emotional response and offense as a deliberate evaluative stance.

\textbf{Commenter Strategy:}
Commenters appealed to personal responsibility without interrogating the definition of “offended.”

\textbf{Representative Rebuttal:}
“It’s up to you to deal with offense, not others. Demanding change is an overreaction.”

\textbf{Outcome:}
No delta was awarded. The OP maintained that offense is involuntary and that the rebuttals failed to address this premise.

\textbf{Failure Mechanism:}
By accepting the OP’s definition of offense without challenge, commenters debated downstream consequences rather than the conceptual foundation of the argument. Persuasion failed due to lack of semantic clarification.
\end{tcolorbox}

\begin{tcolorbox}[
    colback=green!5,
    colframe=green!60!black,
    colbacktitle=green!60!black,
    coltitle=white,
    title=Case 3: Definition of Racism (Successful Persuasion),
    fonttitle=\bfseries,
    breakable,
    enhanced
]
\textbf{Thread Title:} CMV: Changing the definition of racism to “prejudice + power” hurts anti-racism efforts

\textbf{OP Original View:}
The OP claimed that racism should be defined strictly as individual prejudice combined with institutional power, and that alternative definitions weaken the term.

\textbf{Key Conceptual Issue:}
Failure to recognize multiple legitimate definitional frameworks (dictionary versus sociological).

\textbf{Commenter Strategy:}
The commenter explicitly distinguished between dictionary and academic definitions of racism.

\textbf{Representative Rebuttal:}
“You are confusing the sociological definition with the standard dictionary definition. Oxford defines racism as prejudice or discrimination based on race. The power-plus-prejudice model is one academic usage among several.”

\textbf{Outcome:}
A delta was awarded. The OP acknowledged that multiple definitional standards can legitimately coexist.

\textbf{Success Mechanism:}
Providing authoritative sources and isolating the semantic distinction resolved the apparent contradiction and reframed the disagreement as contextual rather than substantive.
\end{tcolorbox}

\begin{tcolorbox}[
    colback=green!5,
    colframe=green!60!black,
    colbacktitle=green!60!black,
    coltitle=white,
    title=Case 4: Redefinition of Selfishness (Successful Persuasion),
    fonttitle=\bfseries,
    breakable,
    enhanced
]
\textbf{Thread Title:} CMV: Those who redefine selfishness to include altruism are not engaging in good faith

\textbf{OP Original View:}
The OP argued that including altruistic behavior within the concept of selfishness constitutes bad-faith redefinition.

\textbf{Key Conceptual Issue:}
Conflation of colloquial meaning with technical philosophical usage.

\textbf{Commenter Strategy:}
The commenter identified the mismatch between everyday language and technical terminology in Objectivist philosophy.

\textbf{Representative Rebuttal:}
“You are using the colloquial meaning of selfishness, whereas philosophy employs a technical definition of rational self-interest that can include mutually beneficial actions.”

\textbf{Outcome:}
A delta was awarded. The OP accepted that distinct definitions operate in different intellectual contexts.

\textbf{Success Mechanism:}
By explicitly naming the semantic disconnect (colloquial versus technical), the commenter reframed the disagreement and eliminated the appearance of bad faith.
\end{tcolorbox}

\section{Generator Reasoning Traces: Adaptive Hypothesis Generation}

\label{sec :Reasoning Trace}

We present two reasoning traces from the hypothesis generator at different stages of the discovery process for the CMV persuasion task. These traces illustrate how the LLM adaptively synthesizes prior hypotheses to produce novel, testable claims. In Listing \ref{lst:reasoning_trace_2} we can see that the generator identifies patterns across hypotheses about concrete examples (H7) and structural formatting (H10) to derive a meta-level insight about explicit scope narrowing. In Listing \ref{lst:reasoning_trace_1} we see that the generator weaves together findings about OP characteristics (H5), stylistic mirroring (H21), and content types (H3) to propose a temporal sequencing strategy. Together, these examples demonstrate the adaptive and iterative nature hypothesis discovery akin to general scientific practice.

\begin{itemize}
    \item \textcolor{marketblue}{\rule{0.8em}{0.8em}} indicates explicit retrieval of accepted prior hypotheses from memory
    \item \textcolor{industryred}{\rule{0.8em}{0.8em}} indicates the hypothesis formulation
\end{itemize}

\subsection{Seed Hypothesis \#15: Generator Reasoning}
\begin{tcolorbox}[label=lst:reasoning_trace_2]
Okay, let me think about what's already been established, the user has asked me to come with a novel hypothesis about what persuades people in online discourse on CMV. I MUST generate a hypothesis that is novel to the previous 14 in the memory bank. Hmm, hypotheses 1-6 talk about citations, sentiment, concessions.

\textcolor{marketblue}{But Hypothesis 7 says concrete examples are persuasive, and Hypothesis 10 says structured formatting (bullets, blockquotes) is persuasive.} Both of these have intuitive appeal concrete examples ground abstract arguments, and structure makes complex arguments digestible. Why do these work at a deeper level?
Concrete examples work because they narrow an abstract claim, Bullets and blockquotes work because they impose structure on what could otherwise be an overwhelming wall of text they signal ``here are the discrete points I'm making." \textcolor{marketblue}{Infact Hypothesis 12 says wall of texts with too many counterarguments can reduce persuasion success!}

Hmm. Both strategies are fundamentally about reducing cognitive load and constraining the problem space. The OP posted something that probably has multiple concerns woven together maybe 3-4 different worries all tangled up. Let me look at the examples for multi-concern arguments \ldots if you read the posts carefully, there's usually a primary concern (the thing that really bothers them) and several secondary concerns.

\textcolor{industryred}{\textbf{WAIT then perhaps apart from just HAVING concrete examples or using bullets\ldots what if you prioritize addressing what the OP actually cares most about with a concrete example? Instead of just using examples or structure implicitly, what if you explicitly signal to the OP: ``I'm going to focus on X" before diving in?}} This would: Tell the OP you've actually read and understood their post and allow your concrete examples to land with maximum impact because they're clearly relevant to the narrowed scope.

The hypothesis would be: ``Counterarguments that explicitly narrow the scope of the debate (e.g., 'let's focus on\ldots') before countering the top concern by OP are more effective at changing views."

This is testable we can look for scope-narrowing phrases like ``let's focus on", ``I want to specifically address", ``the key issue here is," ``setting aside X, let me talk about Y" appearing early in successful vs. unsuccessful counterarguments. Then we can conduct a test to see if this strategy works.
\end{tcolorbox}

\subsection{Seed Hypothesis \#27: Generator Reasoning}
\begin{tcolorbox}[label=lst:reasoning_trace_1]
Okay, let me think about what's already been established. The user has asked me to come up with a novel hypothesis about what persuades people in online discourse on CMV. I MUST generate a hypothesis that is novel to the previous 26 in the memory bank. Hmm, hypotheses 1-4 talk about citations, length, linguistic markers of hedging.

\textcolor{marketblue}{But Hypothesis 5 says uncertain OPs are more persuadable, Hypothesis 21 says mirroring the argumentation style of OP improves success, and Hypothesis 3 says analogies improve the success rate of counterarguments.} These seem to be pointing at something deeper. H5 tells us WHO to target (uncertain OPs), H21 tells us to ADAPT to the OP's style, and H3 tells us a specific content type that works (analogies). But why are both of these true?

Uncertain OPs are persuadable because they're already signaling openness, they haven't fully committed to their position. Mirroring works because it creates rapport, the OP feels understood rather than attacked. Analogies work because they translate abstract disagreements into relatable scenarios. \textcolor{marketblue}{Infact Hypothesis 8 mentions that empathetic acknowledgment phrases correlate with delta awards, and Hypothesis 14 notes that emotional appeals can backfire if perceived as manipulative!}

Hmm. What if the KEY variable isn't just WHAT you say but WHEN you say it? Mirroring style (H21) is about matching the OP, but what if we should mirror their emotional register too? And if uncertain OPs (H5) are more open, they're probably also showing vulnerability, hedging language, maybe even frustration. Let me look at examples where OPs use emotional language AND hedging together \ldots

\textcolor{industryred}{\textbf{WAIT if uncertain OPs who show emotion are the most persuadable targets, and mirroring creates rapport, then perhaps LEADING with emotionally vivid content when responding to emotional-uncertain OPs would maximize that rapport window!}} An emotionally vivid anecdote is like a personal analogy (H3) but with affective punch. If you open with that, you meet the OP where they ARE emotionally, THEN once that connection is established, you introduce your logical counterargument. The sequencing matters!

The hypothesis would be: ``Starting the argument with emotionally vivid anecdotes and following up with counterarguments when the OP shows emotion and uncertainty is more successful than introducing it later on."

This is testable, we can identify OPs showing both uncertainty markers (hedging language, questions, ``I might be wrong") AND emotional markers (first-person affect words, exclamations). Then check if successful counterarguments to these OPs lead with anecdotes/stories in the first paragraph versus burying them later. We can also test if this sequencing effect disappears for OPs who are purely logical and certain.
\end{tcolorbox}

%% file: pages/appendix/prompts.tex
\section{Prompts}
\subsection{Agent Prompts}
\label{subsec:Agent Prompts}
\lstdefinestyle{promptstyle}{
    basicstyle=\ttfamily\footnotesize,
    breaklines=true,
    breakatwhitespace=false,
    columns=fullflexible,
    keepspaces=true,
    showstringspaces=false,
    frame=none
}

\begin{tcolorbox}[
    colback=white,
    colframe=black!75!black,
    colbacktitle=black,
    coltitle=white,
    title=Generator Agent Prompt,
    fonttitle=\bfseries,
    breakable,
    enhanced,
    label=box:generator_prompt
]
\lstset{style=promptstyle}
\begin{lstlisting}
You are a hypothesis generator working with an Experimenter agent to discover statistically validated hypotheses from data.

Task: ${config.task_description}

Your Role:
You propose plausible, testable hypotheses that the Experimenter will evaluate through statistical analysis. Your goal is to maximize the discovery of novel, statistically significant hypotheses while minimizing false positives.

Search Strategy:
Your search proceeds in two phases:

1. SEED GENERATION (start of session): When the session begins or when exploring new directions, propose hypotheses that are:
   - Novel: Semantically distant from hypotheses already in the hypothesis bank
   - Plausible: Grounded in reasonable domain assumptions
   - Diverse: Covering different categories (linguistic, semantic, rhetorical, stylistic, topical, pragmatic)

2. LOCAL REFINEMENT (after receiving analysis): Once you receive experimental feedback, stay within the same hypothesis family and propose refinements that:
   - Address confounds identified by the Experimenter
   - Add contextual qualifiers to increase specificity
   - Narrow feature definitions to reduce measurement noise
   - Do NOT introduce entirely new variables or shift to unrelated features

Guidelines:
- You will receive a Hypothesis Bank containing validated hypotheses from previous sessions. Do NOT reuse the same variables, features, or analytical patterns already in the bank.
- You will receive Session History showing hypotheses proposed in the current session. You may refine these, but keep variables consistent.
- Hypotheses must be testable using the Experimenter's tools:
  * Code Interpreter: For computable features like text length, word counts, punctuation patterns, regex matches, readability scores, n-gram frequencies, positional features, temporal features, and any transformation computable over the dataset
  * LLM Feature Extractor: For semantic features like sentiment, emotional tone, topic classification, persuasion strategies, rhetorical devices, argument quality, and other judgments requiring language understanding
- Each hypothesis must be one clear, concise, testable sentence expressing a relationship between a feature and the outcome variable.
- Prioritize hypotheses that can be rigorously tested over vague directional claims.

Response Format:
{
  "hypothesis": "<The testable hypothesis statement - one clear sentence>",
  "request": "<Detailed instructions for the Experimenter on how to operationalize and test this hypothesis>",
  "test": true | false
}

Set "test" to false for exploratory analysis (early iterations, uncertain hypotheses) or true for statistical hypothesis testing (confident hypotheses, later iterations).

user_template: |
    {
    === Iteration Status ===
    {{ iteration }}/{{ max_iterations }} ({{ iterations_remaining }} iterations remaining)

    Strategy Guidance:
    - Early iterations: Focus on exploration, propose diverse seed hypotheses
    - Mid iterations: Balance exploration with refinement of promising directions
    - Late iterations: Focus on rigorous testing and validation of refined hypotheses

    === Dataset Description ===
    {{ data_description }}

    === Hypothesis Bank (Validated discoveries from previous sessions) ===
    {
    The following {{ memory|length }} hypotheses have been validated. Propose hypotheses that are novel relative to these:
    {
    {{ loop.index }}. {{ hyp }}
    {
    {
    No hypotheses validated yet. You are free to explore any direction.
    {

    === Session History (Current session hypotheses) ===
    {
    Hypotheses proposed in this session (refine within this family):
    {
    {{ loop.index }}. {{ hyp }}
    {
    {
    No previous hypotheses in this session. Propose a novel seed hypothesis.
    {

    === Current Hypothesis Under Consideration ===
    {{ current_hypothesis }}

    === Experimenter's Recent Analysis Results ===
    {{ previous_analysis }}

    Based on the above, propose your next hypothesis.
    {
\end{lstlisting}
\end{tcolorbox}

\begin{tcolorbox}[
    colback=white,
    colframe=black!75!black,
    colbacktitle=black,
    coltitle=white,
    title=Experimenter Agent Prompt,
    fonttitle=\bfseries,
    breakable,
    enhanced,
    label=box:experimenter_prompt
]
\lstset{style=promptstyle}
\begin{lstlisting}
You are an Experimenter agent responsible for rigorously evaluating hypotheses proposed by the Generator.

Task: ${config.task_description}

Your Role:
You evaluate natural-language hypotheses by constructing and executing empirical evaluation procedures. Your analysis must be rigorous enough to minimize false positives while providing actionable feedback for hypothesis refinement.

Tools Available:
You have access to two tools in a sandboxed environment:

1. CODE INTERPRETER
   A persistent Python execution environment with pandas, numpy, scipy, statsmodels, and scikit-learn.

   Use for:
   - Loading and exploring the dataset
   - Computing syntactic features: text length, word counts, sentence counts, punctuation frequency, capitalization patterns
   - Computing lexical features: vocabulary richness, n-gram frequencies, specific word/phrase detection via regex
   - Computing structural features: paragraph counts, list usage, formatting patterns
   - Computing relational features: position within thread, temporal rank, response timing, comparisons to group statistics
   - Running statistical tests: t-tests, chi-square tests, Mann-Whitney U, logistic regression, ANOVA
   - Computing effect sizes: Cohen's d, odds ratios, correlation coefficients
   - Robustness checks: controlling for confounds, subgroup analyses, sensitivity analyses

   The kernel maintains state across calls - variables, loaded data, and helper functions persist between tool calls.

2. LLM FEATURE EXTRACTOR
   An LLM-based annotator for semantic features that cannot be reliably computed through code.

   Use for:
   - Sentiment classification (positive/negative/neutral)
   - Emotional tone detection (anger, joy, sadness, fear, etc.)
   - Topic classification
   - Persuasion strategy identification (logos, pathos, ethos)
   - Rhetorical device detection (metaphor, analogy, rhetorical questions)
   - Argument quality assessment
   - Any judgment requiring semantic understanding of content

   When calling this tool, specify: the column to extract from, feature name, feature description, and allowed feature types (categories).

Workflow:
1. PLAN: Think step-by-step about how to operationalize the hypothesis
   - What feature(s) need to be measured?
   - What is the appropriate statistical test given the data types?
   - What confounds should be controlled for?

2. EXECUTE: Use your tools to compute features and run analyses
   - Prefer code_interpreter for any feature that can be computed programmatically
   - Use feature_extractor only for semantic features requiring LLM judgment
   - You may call tools multiple times as needed

3. ANALYZE: Run appropriate statistical tests
   - Select tests appropriate for your data types (continuous vs categorical)
   - Compute effect sizes, not just p-values
   - Check assumptions (normality, homogeneity of variance, etc.)

4. VALIDATE: Conduct robustness checks
   - Control for obvious confounds (e.g., text length)
   - Run sensitivity analyses if appropriate
   - Consider subgroup analyses

5. REPORT: Provide structured results with actionable guidance
   - Clear statement of support/non-support
   - Effect sizes with confidence intervals
   - Suggestions for hypothesis refinement

CRITICAL: Never end your turn without completing the analysis.

Output Format:
Provide a structured analysis report containing:
- Feature Construction: How you operationalized the hypothesis constructs
- Statistical Test: The test used and why it was appropriate
- Results: Effect size, p-value, confidence interval, sample sizes
- Robustness: Any controls, sensitivity analyses, or assumption checks
- Conclusion: Whether the hypothesis is supported and guidance for refinement

user_template: |
    {
    === Dataset ===
    Path: {{ dataset_path }}

    {{ data_description }}

    === Hypothesis to Evaluate ===
    {{ analysis_request }}

    === Evaluation Mode ===
    {
    STATISTICAL TESTING MODE
    Perform rigorous hypothesis testing:
    - Select an appropriate statistical test for the data types involved
    - Compute effect sizes with confidence intervals
    - Apply multiple testing correction if testing multiple comparisons
    - Control for relevant confounds
    - Conduct robustness/sensitivity analyses
    - Provide a clear verdict on statistical support
    {
    EXPLORATORY ANALYSIS MODE
    Conduct exploratory data analysis:
    - Examine distributions and patterns related to the hypothesis
    - Identify potential confounds or moderating variables
    - Surface unexpected findings that may inform refinement
    - Suggest specific, testable refinements of the hypothesis
    {

    Proceed with your analysis.
    {
\end{lstlisting}
\end{tcolorbox}

\subsection{Task Prompts}

\begin{tcolorbox}[
    colback=gray!8,
    colframe=gray!75,
    breakable,
    enhanced jigsaw,
    fontupper=\footnotesize,
    boxrule=0.6pt,
    arc=2mm,
    left=6pt,
    right=6pt,
    top=6pt,
    bottom=6pt,
    title=Deceptive Reviews Prompt
]
\textbf{task\_description:} You’re a professional hotel review analyst. Given a set of hotel reviews, we want to generate hypotheses that are useful for predicting whether a review is truthful or deceptive.
\end{tcolorbox}

\begin{tcolorbox}[
    colback=gray!8,
    colframe=gray!75,
    breakable,
    enhanced jigsaw,
    fontupper=\footnotesize,
    boxrule=0.6pt,
    arc=2mm,
    left=6pt,
    right=6pt,
    top=6pt,
    bottom=6pt,
    title=GPT Generated Content Prompt
]
\textbf{task\_description:} Given a story, find a hypothesis that can classify the story as human or AI-generated.
\end{tcolorbox}

\begin{tcolorbox}[
    colback=gray!8,
    colframe=gray!75,
    breakable,
    enhanced jigsaw,
    fontupper=\footnotesize,
    boxrule=0.6pt,
    arc=2mm,
    left=6pt,
    right=6pt,
    top=6pt,
    bottom=6pt,
    title=Dreaddit Prompt
]
\textbf{task\_description:} You are a research scientist generating content-based hypotheses about whether a Reddit post contains mental stress signals. The task is to identify linguistic features and patterns that help distinguish stressful posts from non-stressful ones.
\end{tcolorbox}

\begin{tcolorbox}[
    colback=gray!8,
    colframe=gray!75,
    breakable,
    enhanced jigsaw,
    fontupper=\footnotesize,
    boxrule=0.6pt,
    arc=2mm,
    left=6pt,
    right=6pt,
    top=6pt,
    bottom=6pt,
    title=News Headlines Prompt
]
\textbf{task\_description:} You are a professional writer for an online newspaper company. Given a pair of headlines created for the same article, you are asked to determine which will get more clicks. It is likely that the pair of headlines shares similarities, so please focus on their differences. What difference in two headlines leads to more clicks on one than the other?
\end{tcolorbox}

\begin{tcolorbox}[
    colback=gray!8,
    colframe=gray!75,
    breakable,
    enhanced jigsaw,
    fontupper=\footnotesize,
    boxrule=0.6pt,
    arc=2mm,
    left=6pt,
    right=6pt,
    top=6pt,
    bottom=6pt,
    title=Persuasive Pairs Prompt
]
\textbf{task\_description:} You are a research scientist generating content-based hypotheses about which text among the pair is more persuasive. The task is to identify patterns that distinguish persuasive text from non/less non-persuasive text.
\end{tcolorbox}

\begin{tcolorbox}[
    colback=gray!8,
    colframe=gray!75,
    breakable,
    enhanced jigsaw,
    fontupper=\footnotesize,
    boxrule=0.6pt,
    arc=2mm,
    left=6pt,
    right=6pt,
    top=6pt,
    bottom=6pt,
    title=ChangeMyView (CMV) Prompt
]
\textbf{task\_description:} Given a discussion containing the title, the original argument, and the counterargument from Reddit's r/ChangeMyView, identify what makes counterarguments effective in changing the original poster's view. Focus on persuasion strategies such as linguistic style matching, argument structure, use of evidence, hedging language, and interaction dynamics rather than the topic of the discussion.
\end{tcolorbox}

\begin{tcolorbox}[
    colback=gray!8,
    colframe=gray!75,
    breakable,
    enhanced jigsaw,
    fontupper=\footnotesize,
    boxrule=0.6pt,
    arc=2mm,
    left=6pt,
    right=6pt,
    top=6pt,
    bottom=6pt,
    title=Design Layout Prompt
]
\textbf{task\_description:} You have to generate hypotheses that can accurately predict the characteristics of layouts that make one layout preferred over the other.
\end{tcolorbox}